\documentclass[journal]{IEEEtran}

\usepackage{cite}
\usepackage{graphicx,subfigure}
\usepackage{amsmath,amsthm,amssymb,bm}
\usepackage{pifont}
\usepackage{multirow}

\usepackage[hidelinks]{hyperref}
\usepackage{subfigure}
\usepackage{algorithm}
\usepackage{algorithmic}

\usepackage{color}
\usepackage{array}
\usepackage{booktabs}

\newtheorem{definition}{Definition}

\begin{document}

\title{Catastrophic Interference in Reinforcement Learning: A Solution Based on Context Division and Knowledge Distillation}

\author{Tiantian~Zhang, Xueqian~Wang,~\IEEEmembership{Member,~IEEE,}\\
        Bin~Liang,~\IEEEmembership{Senior Member,~IEEE,}
        and~Bo~Yuan$^\ast$,~\IEEEmembership{Senior Member,~IEEE}
\thanks{Tiantian Zhang, and Bo Yuan are with the Intelligent Computing Lab, Shenzhen International Graduate School, Tsinghua University, 518055 Shenzhen, P.R. China (e-mail: ztt19@mails.tsinghua.edu.cn; boyuan@ieee.org).}
\thanks{Xueqian Wang is with the Center for Artiﬁcial Intelligence and Robotics, Shenzhen International Graduate School, Tsinghua University, 518055 Shenzhen, P.R. China (e-mail: wang.xq@sz.tsinghua.edu.cn).}
\thanks{Bin Liang is with the Research Center for Navigation and Control, Department of Automation, Tsinghua University, 100084 Beijing, P.R. China (e-mail: liangbin@mail.tsinghua.edu.cn).}
\thanks{$^\ast$Corresponding Author.}}

\maketitle

\begin{abstract}
The powerful learning ability of deep neural networks enables reinforcement learning agents to learn competent control policies directly from continuous environments. In theory, to achieve stable performance, neural networks assume {\em i.i.d.} inputs, which unfortunately does no hold in the general reinforcement learning paradigm where the training data is temporally correlated and non-stationary. This issue may lead to the phenomenon of ``{\em catastrophic interference}" and the collapse in performance.
In this paper, we present IQ, {\em i.e.}, interference-aware deep Q-learning, to mitigate catastrophic interference in single-task deep reinforcement learning. 
Specifically, we resort to online clustering to achieve on-the-fly context division, together with a multi-head network and a knowledge distillation regularization term for preserving the policy of learned contexts.
Built upon deep Q networks, IQ consistently boosts the stability and performance when compared to existing methods, verified with extensive experiments on classic control and Atari tasks. The code is publicly available at: \url{https://github.com/Sweety-dm/Interference-aware-Deep-Q-learning}.
\end{abstract}


\begin{figure}[t]
  \centering
  \setlength{\abovecaptionskip}{3pt}
  {\includegraphics[width=0.98\linewidth]{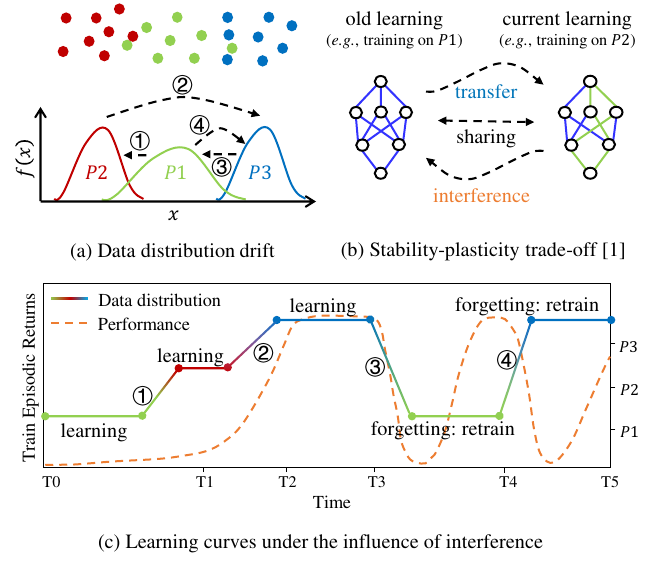}}
  \caption{An illustration of the catastrophic interference in the single-task RL. (a) The drift of data distributions during learning, where $P1\mbox{\textendash}P3$ are different data distributions and \ding{192}\textendash\ding{195} represent distribution transitions. The agent experiences the following data distribution transitions during learning:$P1\xrightarrow{\mbox{\ding{192}}}P2\xrightarrow{\mbox{\ding{193}}}P3\xrightarrow{\mbox{\ding{194}}}P1\xrightarrow{\mbox{\ding{195}}}P3$. (b) The stability-plasticity trade-off in DNNs\cite{riemer2019learning}. Sharing: both learning phases train the same model; Transfer: the current learning phase continues training on the model derived from the old learning phase; Interference: after the model is trained on $P2$, the weights in green are changed in the right network, affecting the model performance on $P1$. (c) The learning curves where the solid line corresponds to the data distribution transitions in (a) and the dashed line shows the training performance. Before $T3$, the data distribution is gradually drifted from $P1$ to $P2$ and to $P3$. When the model fits to $P3$, the learned policies on $P1$ and $P2$ are interfered, resulting in catastrophically degraded performance when the agent encounters states from $P1$ again. Therefore, the model needs to be retrained on $P1$ (in the time period $T3\rightarrow T4$). The same problem occurs in the time period $T4\rightarrow T5$.}
  \label{fig:introduction_example}
\end{figure}

\section{Introduction}
\IEEEPARstart{I}{n} recent years, the successful application of {\em deep neural networks} (DNNs) in {\em reinforcement learning} (RL) \cite{sutton2018reinforcement} has provided a new perspective to boost its performance on high-dimensional continuous problems. With the powerful function approximation and representation learning capabilities of DNNs, deep RL is regarded as a milestone towards constructing autonomous systems with a higher level of understanding of the physical world \cite{li2019deep}. Currently, deep RL has demonstrated great potential on complex tasks, from learning to play video games directly from pixels \cite{mnih2013playing,mnih2015human} to making immediate decisions on robot behavior from camera inputs \cite{faust2018prm,chiang2019learning,francis2020long}. However, these successes are limited and prone to catastrophic interference\footnote{A phenomenon observed in neural networks where later training is likely to overwrite and interfere with previously learned good policies and significantly degrades the performance on previous tasks.} due to the inherent issue of DNNs in face of the non-stationary data distributions, and they rely heavily on a combination of various subtle strategies, such as experience replay \cite{mnih2013playing} and fixed target networks \cite{mnih2015human}, or distributed training architecture \cite{mnih2016asynchronous,bellemare2017distributional,espeholt2018impala}.

Catastrophic interference is the primary challenge for many neural-network-based machine learning systems when learning over a non-stationary stream of data \cite{mccloskey1989catastrophic}. It is normally investigated in multi-task {\em continual learning} (CL), mainly including {\em supervised continual learning} (SCL) for classification tasks \cite{kirkpatrick2017overcoming,fernando2017pathnet,lopez2017gradient,rebuffi2017icarl,mallya2018packnet,delange2021continual} and {\em continual reinforcement learning} (CRL) \cite{kirkpatrick2017overcoming,fernando2017pathnet,riemer2019learning,kessler2020unclear,khetarpal2020towards} for decision tasks. In the multi-task CL, the agent continually faces new tasks and the neural network may quickly fit to the data distribution of the current task, while potentially overwriting the information related to learned tasks, leading to catastrophic forgetting of the solutions of old tasks. The underlying reason behind this phenomenon is the global generalization and overlapping representation of neural networks \cite{ghiassian2020improving,bengio2020interference}. Neural networks training normally assumes that the inputs are identically and independently distributed ({\em i.i.d.}) from a fixed data distribution and the output targets are sampled from a fixed conditional distribution. Only when this assumption is satisfied, can positive generalization be ensured among different batches of stochastic gradient descent. However, when the data distribution is drifted during training, the information learned from old tasks may be negatively interfered or even overwritten by the newly updated weights, resulting in catastrophic interference.

Deep RL is essentially a CL problem due to its learning mode of exploring while learning \cite{khetarpal2020towards}, and it is particularly vulnerable to catastrophic interference, even within many single-task settings where the environment is stationary (such as Atari 2600 games, or even simpler classic control tasks in OpenAI Gym) \cite{schaul2019ray,fedus2020catastrophic,lo2019overcoming,liu2019utility}. The non-stationarity of data distributions in the single-task RL is mainly attributed to the following properties of RL. Firstly, the inputs of RL are sequential observations received from the environment, which are temporally correlated. Secondly, in the progress of learning, the agent's decision making policy changes gradually, which makes the observations non-stationary. Thirdly, RL methods rely heavily on bootstrapping, where the RL agent uses its own estimated value function as the target, making the target outputs also non-stationary. In addition, as noted in \cite{fedus2020catastrophic}, replay buffers with prioritized experience replay \cite{schaul2016prioritized} that preferentially sample experiences with higher {\em temporal-difference} (TD) errors will also exasperate the non-stationarity of training data. Once the distribution of training data encounters notable drift, catastrophic interference and a chain reaction are likely to occur, resulting in a sudden deterioration of the training performance, as shown in Fig. \ref{fig:introduction_example}. 

Currently, there are two major strategies for dealing with catastrophic interference in the single-task RL training: experience replay \cite{mnih2013playing,mnih2015human} and local optimization \cite{liu2019utility,lo2019overcoming}. The former usually exhibits high-level sensitivity to key parameters ({\em e.g.,} replay buffer capacity) and often requires maintaining a large experience storage memory. Furthermore, the sufficiently large memory may increase the degree of off-policyness of transitions in the buffer \cite{fedus2020revisiting}, violating the requirement of current state-of-the-art algorithms that the data should be close to the on-policy distribution, even for off-policy algorithms like DQN. The latter advocates local network updating for the data with a specific distribution instead of global generalization to reduce the representation overlap among different data distributions. The major issues are that some methods are limited in the capability of model transfer among differently distributed data \cite{liu2019utility,lo2019overcoming}, or require pretraining and may not be suitable for the online settings \cite{liu2019utility}.

In this paper, we focus on the catastrophic interference problem caused by state distribution drift in the single-task RL. We propose a interference-aware scheme with low buffer-size sensitivity called {\em Interference-aware Deep Q-learning} (IQ\footnote{So named because it uses Deep Q-learning and features interference awareness. IQ also implies a smarter agent in the sense that it is the abbreviation of intelligence quotient.}) that estimates the value function online for each state distribution by minimizing the weighted sum of the original loss function of RL algorithms and the regularization term regarding the interference among different groups of states. The schematic architecture is shown in Fig. \ref{fig:IQ_framework}.

In order to mitigate the interference among different state distributions during model training, we introduce the concept of ``{\em context}" into the single-task RL, and propose a context division strategy based on online clustering. We show that it is essential to decouple the correlations among different state distributions with this strategy to divide the state space into a series of independent contexts (each context is a set of states distributed close to each other, conceptually similar to ``{\em task}" in the multi-task CRL). To achieve efficient and adaptive partition, we employ {\em Sequential K-Means Clustering} \cite{dias2008skm} to process the states encountered during training in real time. Then, we parameterize the value function by a neural network with multiple output heads commonly used in multi-task learning \cite{zenke2017continual,golkar2019continual,kessler2020unclear} in which each output head specializes on a specific context, and the feature extractor is shared across all contexts. In addition, we apply knowledge distillation as a regularization term in the objective function for value function estimation, which can preserve the learned policies while the RL agent is trained on states guided by the current policy, to further avoid the interference caused by the shared low-level representation. Furthermore, to ease the curse of dimensionality in high-dimensional state spaces, we employ a random encoder as its low-dimensional representation space can effectively capture the information about the similarity among states without any representation learning \cite{seo2021state}. Clustering is then performed in the low-dimensional representation space of the randomly initialized convolutional encoder. 

The contributions of this paper are summarized as follows:

\begin{itemize}
\item[1)] A novel context division strategy is proposed for the single-task RL. It is essential as the widely studied multi-task CRL methods cannot be used directly to reduce interference in the single-task RL due to the lack of predefined task boundaries. This strategy can detect contexts adaptively online, so that each context can be regarded as a task in multi-task settings. In this way, the strategies designed for the multi-task CRL can be used in the single-task RL to mitigate catastrophic interference.

\item[2)] A novel RL training scheme called IQ based on multi-head neural networks is proposed following the context division strategy. By incorporating the knowledge distillation loss into the objective function, IQ can better alleviate the interference suffered in the single-task RL than existing methods in a fully online manner.

\item[3)] A fixed random encoder is introduced into the context division of high-dimensional state spaces, which further stabilizes the performance of IQ on complex RL tasks ({\em e.g.}, image-level inputs) compared with the underlying RL trained encoder.

\item[4)] Extensive experiments on a suite of OpenAI Gym standard benchmark environments ranging from classic control tasks to high-dimensional complex Arcade Learning Environments (ALE) \cite{bellemare2013arcade} under various replay buffer capacity settings are conducted to validate that the overall superiority of our method over baselines in terms of the stability and the maximum achieved cumulative reward.
\end{itemize}
 
The rest of this paper is organized as follows. Section II reviews the relevant strategies for alleviating catastrophic interference as well as context detection and identification. Section III introduces the nature of RL in terms of continual learning and gives an example analysis of catastrophic interference in the single-task RL. The details of IQ are shown in Section IV, and experimental results and analyses are presented in Section V. Finally, this paper is concluded in Section VI with some discussions and directions for future work.

\section{Related Work}
Catastrophic interference within the single-task RL is a special case of CRL, which involves not only the strategies to mitigate interference but also the context detection and identification techniques.

\subsection{Multi-task Continual Reinforcement Learning}
Multi-task CRL has been an active research area with the development of RL architectures \cite{lesort2020continual}. Existing methods mainly consist of three categories: experience replay-based, regularization-based, and parameter isolation-based methods. 

The core idea of experience replay is to store samples of previous tasks in raw format ({\em e.g.,} Selective Experience Replay (SER) \cite{isele2018selective}, Meta Experience Replay (MER) \cite{riemer2019learning}, Continual Learning with Experience And Replay (CLEAR) \cite{rolnick2019experience}) or generate pseudo-samples from a generative model ({\em e.g.,} Reinforcement Pseudo Rehearsal (RePR) \cite{atkinson2021pseudo}) to maintain the knowledge about the past in the model. These previous task samples are replayed while learning a new task to alleviate interference, in the form of either being reused as model inputs for rehearsal \cite{isele2018selective,atkinson2021pseudo} or constraining the optimization of the new task loss \cite{rolnick2019experience,riemer2019learning}. Experience replay has become a very successful approach to tackling interference in CRL. However, the raw format may result in significant storage requirements for complex CRL settings. Although the generative model can be exempted from a replay buffer, it is still difficult to capture the overall distribution of previous tasks.

Regularization-based methods avoid storing raw inputs by introducing an extra regularization term into the loss function to consolidate previous knowledge while learning on new tasks. The regularization term includes penalty computing and knowledge distillation. The former focuses on reducing the chance of weights being modified. For example, Elastic
Weight Consolidation (EWC) \cite{kirkpatrick2017overcoming} and UNcertainty guided Continual LEARning (UNCLEAR) \cite{kessler2020unclear} use Fisher matrix to measure the importance of weights and protect important weights on new tasks. The latter is a form of knowledge transfer \cite{hinton2015distilling}, which expects that the model trained on a new task can still perform well on the old ones. It is often used for policy transfer from one model to another ({\em e.g.,} Policy Distillation \cite{rusu2016policy}, Genetic Policy Optimization (GPO) \cite{gangwani2018policy}, Distillation for Continual Reinforcement learning (DisCoRL) \cite{traore2019discorl}). This family of solutions is easy to implement and tends to perform well on a small number of tasks, but still faces challenges as the number of tasks increases.

Parameter isolation-based methods dedicate different model parameters to each task, to prevent any possible interference among tasks. Without the constraints on the size of neural networks, one can grow new branches for new tasks, while freezing previous task parameters ({\em e.g.,} Progressive Natural Networks (PNN) \cite{rusu2016progressive}). Alternatively, the architecture remains static, with fixed parts being allocated to each task. For instance, PathNet \cite{fernando2017pathnet} uses a genetic algorithm to find a path from input to output for each task in the neural network and isolates the used network parts in parameter level from the new task training. These methods typically require networks with enormous capacity, especially when the number of tasks is large, and there is often unnecessary redundancy in the network structure, bringing a great challenge to model storage and efficiency. 

\subsection{Single-task Reinforcement Learning}
Compared with the multi-task CRL, catastrophic interference in the single-task RL remains an emerging research area, which has been relatively under-explored. There are two primary aspects of previous studies: one is finding supporting evidence to confirm that catastrophic interference is indeed prevalent within a specific RL task, and the other is proposing effective strategies for dealing with it. 

Researchers in DeepMind studied the learning dynamics of the single-task RL and developed a hypothesis that the characteristic coupling between learning and data generation is the main cause of interference and performance plateaus in deep RL systems \cite{schaul2019ray}. Recent studies further confirmed this hypothesis and its universality in the single-task RL through large-scale empirical studies (called {\em Memento experiments}) in Atari 2600 games \cite{fedus2020catastrophic}. However, none of these studies has suggested any practical solution for tackling the interference.

In order to mitigate interference, many deep RL algorithms such as DQN \cite{mnih2015human} and its variants ({\em e.g.,} Double DQN \cite{van2016deep}, Rainbow \cite{hessel2018rainbow}) employ experience replay and fixed target networks to produce approximately {\em i.i.d.} training data, which may quickly become intractable in terms of memory requirement as task complexity increases. Furthermore, even with sufficient memory, it is still possible to suffer from catastrophic interference due to the imbalanced distribution of experiences. 

In recent studies \cite{liu2019utility,lo2019overcoming,ghiassian2020improving}, researchers proposed some methods based on local representation and optimization of neural networks, which showed that interference can be reduced by promoting the local updating of weights while avoiding global generalization. Sparse Representation Neural Network (SRNN) \cite{liu2019utility} induces sparse representations in neural networks by introducing a distributional regularizer, which requires a large batch of data generated by a fixed policy that covers the space for pretraining and have not been extended to the online setting. Dynamic Self-Organizing Map (DSOM) \cite{lo2019overcoming} with neural networks introduces a DSOM module to induce such locality updates. These methods can reduce interference to some extent, but they may inevitably suffer from the lack of positive transfer in the representation layer and require larger network capacity, which is not desirable in complex tasks. Recently, discretizing (D-NN) and tile coding (TC-NN) were used to remap the input observations to a high-dimensional space to sparsify input features, reducing the activation overlap \cite{ghiassian2020improving}. However, tile coding increases the dimension of inputs to a neural network, which can lead to scalability issues for spaces with high dimensionality.  

\subsection{Context Detection and Identification}
Context detection and identification is a fundamental step for learning task relatedness in CL. Most multi-task CL methods aforementioned rely on well-defined task boundaries, and are usually trained on a sequence of tasks with known labels or boundaries. Existing context detection approaches commonly leverage statistics or Bayesian inference to identify task boundaries. 

On the one hand, some methods tend to be reactive to a changing distribution by finding change points in the pattern of state-reward tuples ({\em e.g.,} Context QL \cite{padakandla2020reinforcement}), or tracking the difference between the short-term and long-term moving average rewards ({\em e.g.,} CRL-Unsup \cite{lomonaco2020continual}), or splitting a game into contexts using the undiscounted accumulated game score as a task contextualization \cite{jain2020algorithmic}. These methods can be agile in responding to scenarios with abrupt changes among contexts or tasks, but are insensitive to smooth transitions from one context to another.

On the other hand, some more ambitious approaches try to learn a belief of the unobserved context state directly from the history of environment interactions, such as Forget-me-not Process (FMN) \cite{milan2016forget} for piecewise-repeating data generating sources, and Continual Unsupervised Representation Learning (CURL) \cite{rao2019continual} for task inference without any knowledge about task identity. However, they both need to be pretrained with the complete data before applied to CL problems, and CURL itself also needs additional techniques to deal with the interference.

Furthermore, Ghosh {\em et al.} \cite{ghosh2018divide} proposed to partition the initial state space into a finite set of contexts by performing a K-Means clustering procedure, which can decompose more complex tasks, but cannot completely decouple the correlations among different state distributions from the perspective of interference prevention.

\section{Preliminaries and Problem Statement}
To better characterize the problem studied in this paper, some key definitions and glossaries of CRL problems are introduced in this section.

\subsection{Definitions and Glossaries}
Some important definitions of RL relevant to this paper are presented as follows.

\begin{definition}[RL Paradigm \cite{sutton2018reinforcement}]
A RL problem is regarded as a Markov Decision Process (MDP), which is defined as a tuple $M=\langle \mathcal{S},\mathcal{A},\mathcal{P},\mathcal{R},\gamma \rangle$, 
where $\mathcal{S}$ is the set of states; $\mathcal{A}$ is the set of actions; $\mathcal{P}:\mathcal{S}\times\mathcal{A}\times\mathcal{S}\rightarrow[0,1]$ is the environment transition probability function; $\mathcal{R}:\mathcal{S}\times\mathcal{A}\times\mathcal{S}\rightarrow\mathbb{R}$ is the reward function, and $\gamma\in[0,1]$ is the discount factor.
\label{def: RL_Paradigm}
\end{definition}

According to Definition \ref{def: RL_Paradigm}, at each time step $t\in\mathbb{N}$, the agent moves from $S_t$ to $S_{t+1}$ with probability $P(S_{t+1}|S_t,A_t)$ after taking action $A_t$, and receives reward $R(S_t,A_t)$. Based on this definition, the optimization objective of value-based RL models is defined as follow:

\begin{definition}[RL Optimization Objective \cite{khetarpal2020towards}]
The optimization objective of the value-based RL is to learn a policy $\pi(a|s)$ with internal parameter $\theta\in \Theta$ that maximizes the expected long-term discounted returns for each $(s,a)$ in time, also known as the value function:
\begin{equation}
\setlength{\belowdisplayskip}{-3pt}
\resizebox{.91\linewidth}{!}{$
    \displaystyle
    J(\pi) = Q_{\pi}(s,a) = \mathbb{E}_{\mathcal{P},\pi}\bigg[\sum_{k=0}^\infty\gamma^kR(S_{t+k},A_{t+k})\bigg|S_t=s,A_t=a\bigg].$}
\end{equation}
\label{def: RL_Goal}
\end{definition}

Here, the expectation is over the process that generates a history using $\mathcal{P}$ and decides actions from $\pi$ until the end of the agent's lifetime.

The optimization objective in Definition \ref{def: RL_Goal} does not just concern itself with the current state, but also the full expected future distribution of states. As such, it is possible to overcome the catastrophic interference for RL over non-stationary data distributions. However, much of the recent work in RL has been in the so called {\em episodic environments}, which optimizes the episodic RL objective:

\begin{definition}[Episodic RL Optimization Objective\cite{khetarpal2020towards}]
Given some future horizon $H$, find a policy $\pi(a|s)$, optimizing the expected discounted returns:
\begin{equation}
\setlength{\belowdisplayskip}{-3pt}
\resizebox{.91\linewidth}{!}{$
    \displaystyle
    J_{episodic}(\pi) = Q_{\pi}(s,a) = \mathbb{E}_{\mathcal{P},\pi}\bigg[\sum_{k=0}^{H-1}\gamma^kR(S_{t+k},A_{t+k})\bigg|S_t=s,A_t=a\bigg].$}
\end{equation}
\label{def: ERL_Goal}
\end{definition}

Here, to ensure the feasibility and ease of implementation of optimization, the objective is only optimized over a future horizon $H$ until the current episode terminates.

It is clear that the episodic objective in Definition \ref{def: ERL_Goal} is biased towards the current episode distribution while ignoring the possibly far more important future episode distributions over the agent's lifetime. Plugging in such an objective directly into the non-stationary RL settings leads to biased optimization, which is likely to cause catastrophic interference effects. 

For large scale domains, the value function is often approximated with a member of the parametric function class, such as a neural network with parameter $\theta\in\Theta$, expressed as $Q(s,a;\theta)$, which is fit online using experience samples of the form $(s,a,r,s^\prime)$. This experience is typically collected into a buffer $\mathcal{B}$ from which batches are later drawn at random to form a stochastic estimate of the loss:
\begin{equation}
    \displaystyle
    \mathcal{L}(\theta)=\mathbb{E}_\mu\bigg[L\Big(r+\gamma \max_{a^{\prime}\in\mathcal{A}}Q(s^\prime,a^\prime;\theta^-)-Q(s,a;\theta)\Big)\bigg],
    \label{eq:loss}
\end{equation}
where $L:\mathbb{R}\rightarrow\mathbb{R}$ is the agent's loss function, and $\mu\in\mathcal{P}(\mathcal{B})$ is the distribution that defines its sampling strategy. In general, the parameter $\theta^-$ used to compute the target $Q(s^\prime,a^\prime;\theta^-)$ is a prior copy of that used for action selection (as the settings of DQN \cite{mnih2015human}). 

Additionally, it is necessary to clarify some important glossaries in relation to CL.

{\em 1)} {\em Non-stationary} \cite{hadsell2020embracing}: a process whose state or probability distribution changes with time.

{\em 2)} {\em Interference} \cite{bengio2020interference}: a type of {\em influence} between two gradient-based processes with objectives $J_1$, $J_2$, sharing parameter $\theta$. Interference is often characterized in the first order by the inner product of their gradients: 
\begin{equation}
    \displaystyle
    \rho_{1,2}=\triangledown_\theta J_1^T\triangledown_\theta J_2,
    \label{interference}
\end{equation}
and can be seen as being constructive ($\rho>0$, {\em transfer}) or destructive ($\rho<0$, {\em interference}), when applying a gradient update using $\triangledown_\theta J_1$ on the value of $J_2$.
    
{\em 3)} {\em Catastrophic Interference} \cite{mccloskey1989catastrophic, hadsell2020embracing}: a phenomenon observed in neural networks training where learning a new task significantly degrades the performance on previous tasks.
    
\subsection{Problem Statement}
\label{Problem Statement}
The interference within the single-task RL can be approximately measured by the difference in TD errors before and after model update under the current policy, referred to as {\em Approximate Expected Interference} (AEI) \cite{liu2020towards}:
\begin{equation}
\resizebox{.91\linewidth}{!}{$
    \displaystyle
    AEI=\mathbb{E}_{\hat{d}}\bigg[\delta(s,a,r,{s^\prime};\theta_t)^{2}-\delta(s,a,r,{s^\prime};\theta_{t-1})^{2}\bigg],
    $}
    \label{AEI}
\end{equation}
where $\hat{d}$ is the distribution of $(s,a,r,s^\prime)$ under the current policy and $\delta(s,a,r,{s^\prime};\theta)= r+\gamma \max_{a^{\prime}\in\mathcal{A}}Q({s^\prime},{a^\prime};\theta)-Q(s,a;\theta)$ is the TD error.

\begin{figure}[t]
  \centering
  \setlength{\abovecaptionskip}{5pt}
  \subfigure[Phase I\label{fig:Interference_Performance_I}]
    {\includegraphics[width=0.98\linewidth]{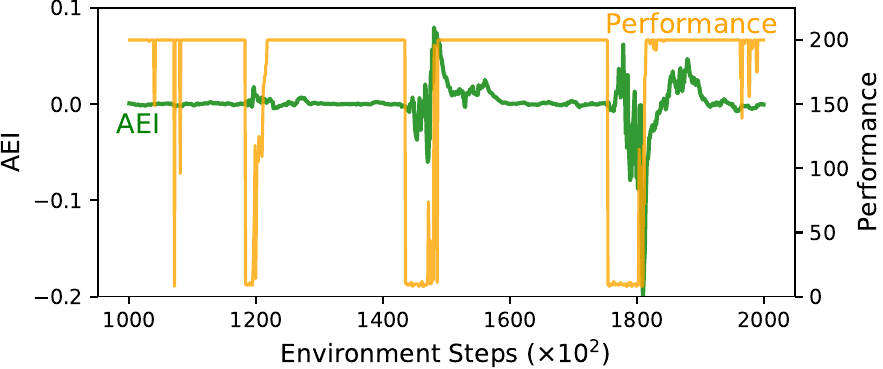}}
  \subfigure[Phase II\label{fig:Interference_Performance_II}]
    {\includegraphics[width=0.98\linewidth]{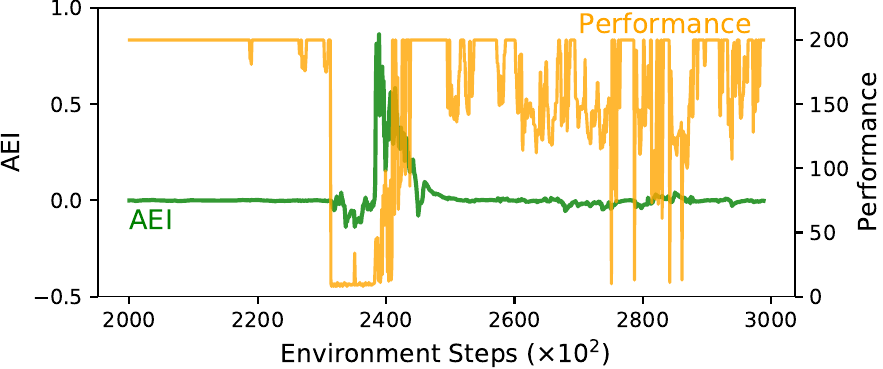}}
  \caption{The interference (green) and training performance (yellow) curve segments of a DQN agent on {\em CartPole} ($N=100$). The interference is measured as the expectation in Eq. \eqref{AEI} and the performance is evaluated by the sum of discounted reward per episode. (a) Phase I with $t=100K\sim200K$; (b) Phase II with $t=200K\sim300K$.}
  \label{fig:Interference_Performance}
\end{figure}

To illustrate the interaction between interference and the agent's performance during the single-task RL training, we run an experiment on {\em CartPole} using the DQN implemented in OpenAI Baselines\footnote{OpenAI Baselines is a set of high-quality implementations of RL algorithms implemented by OpenAI: https://github.com/openai/baselines.}, and set the replay buffer size $N$ to 100, a small capacity to trigger interference to highlight its effect. We trained the agent for 300K environment steps and approximated $\hat{d}$ with a buffer containing recent transitions of capacity 10K to evaluate the {\em AEI} value according to Eq. \eqref{AEI} after each update. Fig. \ref{fig:Interference_Performance} shows two segments of the interference and performance curves during training from which we can see that the performance started to oscillate when {\em AEI} started to increase ({\em e.g.}, $t\approx118K$, $t\approx143K$, and $t\approx175K$ in Fig. \ref{fig:Interference_Performance_I}, and $t\approx230K$ in Fig. \ref{fig:Interference_Performance_II}). In general, the performance of the agent tends to drop significantly in the presence of increasing interference. This result provides direct evidence that interference is correlated closely with the stability of the single-task RL model.

From the analysis above, we state the problem investigated in this paper as: proposing a novel and effective training scheme for the single-task RL, to reduce catastrophic interference and performance oscillation during training, improving the stability and overall performance simultaneously.

\section{The Proposed Method}
\begin{figure*}[t]
  \centering
  \setlength{\abovecaptionskip}{0pt}
  {\includegraphics[width=0.96\linewidth]{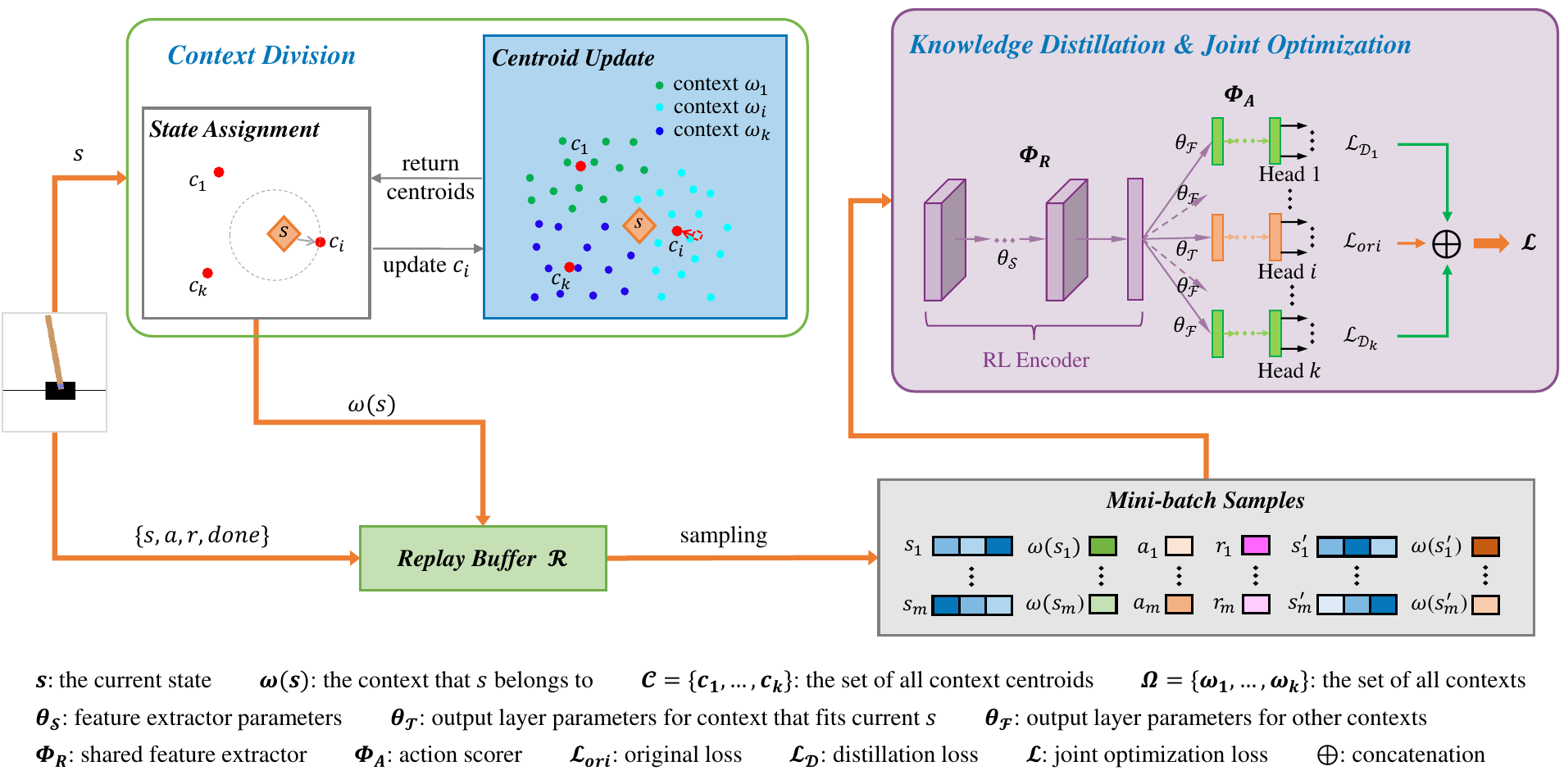}}
  \caption{An overview of the IQ scheme. This framework consists of three components: 1) Context division, including state assignment and centroid update. Adaptive context division is achieved using {\em Sequential K-Means Clustering} online; 2) Knowledge distillation. The knowledge distillation loss ($\mathcal{L_D}$) is incorporated into the objective function ($\mathcal{L}_{ori}$) to avoid interference among contexts due to the shared feature extractor; 3) Joint optimization with a multi-head neural network, which aims to estimate the value for each $(s,a,\omega(s))$ with the joint optimization loss ($\mathcal{L}=\mathcal{L}_{ori}+\lambda\mathcal{L_D}$). Here, to keep consistency with the random encoder introduced later, we also call the representation module of the neural network as "RL Encoder". In summary, our method can improve the performance by decoupling the correlations among differently distributed states and intentionally preserving the learned policies.}
  \label{fig:IQ_framework}
\end{figure*}

In this section, we give a detailed description of our IQ scheme whose architecture is shown in Fig. \ref{fig:IQ_framework}. IQ consists of three main components, which are jointly optimized to mitigate catastrophic interference in the single-task RL: context division, knowledge distillation, and the collaborative training of the multi-head neural network. On the basis of IQ, we further propose IQ-RE with a random encoder for the efficient contextualization of high-dimensional state spaces.

As mentioned before, catastrophic interference is an undesirable byproduct of global updates to the neural network weights on data whose distribution changes over time. A rational solution to this issue is to estimate an individual value function for each distribution, instead of using a single value function for all distributions. When an agent updates its value estimation of a state, the update should only affect the states within the same distribution. With this intuition in mind, we adopt a multi-head neural network with shared representation layers to parameterize the distribution specific value functions.

The IQ scheme proposed in this paper can be incorporated into any existing value-based RL methods to train a piecewise $Q$-function for the single-task RL. The neural network is parameterized by a shared feature extractor and a set of linear output heads, corresponding to each context. As shown in Fig. \ref{fig:IQ_framework}, the set of weights of the $Q$-function is denoted by $\theta=\{\theta_\mathcal{S},\theta_\mathcal{T},\theta_\mathcal{F}\}$, where $\theta_\mathcal{S}$ is a set of shared parameters while $\theta_\mathcal{T}$ and $\theta_\mathcal{F}$ are both context specific parameters: $\theta_\mathcal{T}$ is for the context that corresponds to the current input state $s$, and $\theta_\mathcal{F}$ is for others. In this section, we take the combination of IQ and the basic RL algorithm DQN as an illustrative example.

\subsection{Context Division}
In MDPs, states (or ``observations") represent the most comprehensive information regarding the environment. To better understand the states of different distributions, we define a variable $\omega$ for a set of states that are close to each other in the state space, referred to as “context”. Formally, 
\begin{equation}
    \displaystyle
    \begin{split}
        &\Omega=(\omega_i)_{i=1}^{k}, \\
        &\mathcal{S}=\cup_{i=1}^{k}\mathcal{S}_i,
    \end{split}
    \label{eq:context}
\end{equation}
where $\Omega$ is a finite set of contexts and $k$ is the number of contexts. For an arbitrary MDP, we partition its state space into $k$ contexts, and all states within each context follow approximately the same distribution, to decouple the correlations among states against distribution drift. More precisely, for a partition of $\mathcal{S}$ in Eq. \eqref{eq:context}, we associate a context $\omega_i$ with each set $\mathcal{S}_i$, so that for $s \in \mathcal{S}_i$, $\omega(s)=\omega_i$, where $\omega(s)$ can be thought of as a function of state $s$. 

The inherent learning-while-exploring feature of RL agents leads to the fact that the agent generally does not experience all possible states of the environment while searching for the optimal policy. Thus, it is unnecessary to process the entire state space. Based on this fact, in IQ, we only perform context division on states experienced during training. In this paper, we employ {\em Sequential K-Means Clustering} \cite{dias2008skm} (See Appendix \ref{Sequential K-Means Clustering}) to achieve context detection adaptively. 

In Fig. \ref{fig:IQ_framework}, $k$ centroids $\mathcal{C}=\{c_1,c_2,\dots,c_k\}$ are initialized at random in the entire state space. In each subsequent time step $t$, we execute {\em State Assignment} and {\em Centroid Update} steps for each incoming state received from the environment\footnote{Note that it is suggested to normalize the state in different dimensions before performing these two steps for more reasonable context division results.}, and store its corresponding transition $\{s_t,\omega(s_t),a_t,r_t,s_{t+1},\omega(s_{t+1})\}$ into the replay buffer $\mathcal{B}$. Accordingly, in the training phase, we randomly sample a batch of transitions from $\mathcal{B}$ and train the shared feature extractor $\Phi_R$ and the specific output head $\Phi_A$ corresponding to the input state simultaneously, while conducting fine-tuning of other output heads to avoid interference on learned policies. Since we store the context label of each state in the replay buffer, there are no additional state assignments required at every update step\footnote{In IQ, we only need to perform state assignment once for each state.}.

Note that it is also possible to conduct context division based on the initial state distribution \cite{ghosh2018divide}. By contrast, we show that the partition of all states experienced during training can produce more accurate and effective context division results, as the trajectories starting from the initial states within different contexts have a high likelihood of overlapping in subsequent time steps (See Appendix \ref{ISC_vs_ESC} for more details).

\begin{figure*}[t]
  \centering
  \setlength{\abovecaptionskip}{0pt}
    {\includegraphics[width=0.98\linewidth]{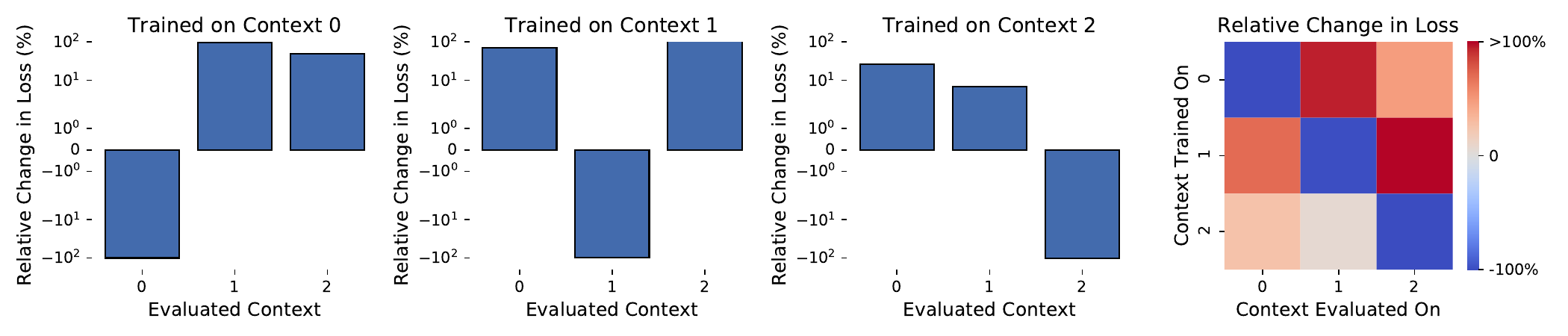}}
  \caption{Measuring the interference among contexts by clustering all experienced states when the agent is trained on {\em CartPole-v0} for 400K environment steps ($k=3$). We record the relative changes in Huber loss for all contexts when the agent is trained on a particular context. It is clear that training on a particular context generally reduces the loss on itself and increases the losses on all other contexts.}
  \label{fig:Interference_btw_Context_cartpole}
\end{figure*}
\textbf{Interference Among Contexts}: We investigate the interference among contexts obtained by our context division method in details. Specifically, we measure the Huber loss of TD errors in different contexts of the game as the agent learns in other contexts, and then record the relative changes in loss before and after the agent's learning, as shown in Fig. \ref{fig:Interference_btw_Context_cartpole}. The results show that, long-term training on any context may lead to negative generalization on all other contexts, even in such simple RL task {\em CartPole-v0}. The results on {\em Pendulum-v0} shown in Appendix \ref{Interference Among Contexts} also support the same conclusion.

\textbf{Computational Complexity}: Assuming a $d$-dimensional environment of $k$ contexts, the time and space complexities of our proposed context division module to process $T$ environment steps are $\mathcal{O}(Tkd)$ and $\mathcal{O}(kd)$, respectively. 

\subsection{Knowledge Distillation}
The shared low-level representation can cause the learning in new contexts to interfere with previous learning results, leading to catastrophic interference. A relevant technique to address this issue is knowledge distillation \cite{hinton2015distilling}, which works well for encouraging the outputs of one network to approximate the outputs of another. The concept of distillation was originally used to transfer knowledge from a complex ensemble of networks to a relatively simpler network to reduce model complexity and facilitate deployment. In IQ, we use it as a regularization term in value function estimation to preserve the previously learned information. 

When training the model on a specific context, we need to consider two aspects of the loss function: the general loss of the current training context (denoted by $\mathcal{L}_{ori}$), and the distillation loss of other contexts (denoted by $\mathcal{L_D}$). The former encourages the model to adapt to the current context to ensure plasticity, while the latter encourages the model to keep the memory of other contexts, preventing interference.

To incorporate IQ into the DQN framework, we rewrite the original loss function of DQN in Eq. \eqref{eq:loss} with the context variable $\omega$ as:
\begin{equation}
    \displaystyle
        \mathcal{L}_{ori}(\theta_\mathcal{S},\theta_\mathcal{T})=\mathbb{E}_\mu\bigg[L\Big(Q_\tau-Q(s,a,\omega(s);\theta_\mathcal{S},\theta_\mathcal{T})\Big)\bigg],
    \label{eq:dqn_loss}
\end{equation}
where
\begin{equation}
    \displaystyle
        Q_\tau=r+\gamma \max_{a^{\prime}\in\mathcal{A}}Q(s^\prime,a^\prime,\omega(s^\prime);\theta_\mathcal{S}^-,\theta_\mathcal{T}^-)
    \label{eq:dqn_target}
\end{equation}
is the estimated target value of $Q(s,a,\omega(s);\theta_\mathcal{S},\theta_\mathcal{T})$ and $\mu$ is the distribution of samples, {\em i.e.}, $\{s,\omega(s),a,r,s^\prime,\omega(s^\prime)\}\sim\mu$, and $L$ refers to the Huber loss.

For each of the other contexts that the environment contains, we expect the output value for each pair of $(s,a)$ to be close to the recorded output from the original network. In knowledge distillation, we regard the learned $Q$-function before the current update step as the {\em teacher network}, expressed as $Q_t^i=Q(s,a,\omega_i;\theta_\mathcal{S}^-,\theta_\mathcal{F}^-)$, and the current network to be trained as the {\em student network}, expressed as $Q_s^i=Q(s,a,\omega_i;\theta_\mathcal{S},\theta_\mathcal{F})$, where $\omega_i\in\Omega$ except the current context $\omega(s)$. Thus, the distillation loss is defined as:
\begin{equation}
    \displaystyle
        \mathcal{L}_\mathcal{D}(\theta_\mathcal{S},\theta_\mathcal{F})=\mathbb{E}_{\mu}\sum_{\omega_i\neq\omega(s),\omega_i\in\Omega}\mathcal{L}_{\omega_i}(\theta_\mathcal{S},\theta_\mathcal{F}),
    \label{eq:distillation_loss}
\end{equation}
where
\begin{equation}
    \displaystyle
        \mathcal{L}_{\omega_i}(\theta_\mathcal{S},\theta_\mathcal{F})=L(Q_t^i-Q_s^i).
    \label{eq:distillation_loss_function}
\end{equation}
is the distillation loss function of the output head corresponding to context $\omega_i$.

\begin{algorithm}[t]
    \caption{IQ: Interference-aware Deep Q-learning}
    \label{alg:IQ}
    \textbf{Input}: Initial replay buffer $\mathcal{B}$ with capacity $|\mathcal{B}|=N$;
    \newline \hspace*{0.95cm} Initial $Q$-function $f_\theta$ with random weights $\theta$;
    \newline \hspace*{0.95cm} Initial target $\hat{Q}$-function $f_{\theta^-}$ with weights $\theta^-=\theta$;
    \newline \hspace*{0.95cm} Initial context centroids $\mathcal{C}=\{c_1,c_2,\dots,c_k\}$;
    \newline \hspace*{0.95cm} Initial target context centroids $\hat{\mathcal{C}}=\mathcal{C}$.\\
    \textbf{Parameter}: Total training steps $T$, the number of contexts $k$,
    \newline \hspace*{1.88cm}target update period $C$, learning rate $\alpha$.\\
    \textbf{Output}: Updated $\mathcal{C}$ and $f_\theta$.
\begin{algorithmic}[1] 
    \STATE Initial state $s$;
    \FOR{$t = 1, T$}
        \STATE Interact with environment to obtain $\{s_t,a_t,r_t,s_{t+1}\}$.
        \STATE States assignment: $\omega(s_t)\xleftarrow{\hat{\mathcal{C}}}s_t$, $\omega(s_{t+1})\xleftarrow{\hat{\mathcal{C}}}s_{t+1}$.
        \STATE Store transition $\{s_t,\omega(s_t),a_t,r_t,s_{t+1},\omega(s_{t+1})\}$ in $\mathcal{B}$.
        \STATE Context centroids update: $\mathcal{C}\leftarrow SKM(s_t,\mathcal{C})$.
        \STATE Joint optimization: \\
        \hspace*{0.2cm} Sample mini-batch $\{s_i,\omega(s_i),a_i,r_i,s_i^\prime,\omega(s_i^\prime)\}_{i=1}^{m}$; \\
        \hspace*{0.2cm} Calculate $\mathcal{L}_{ori}$, $\mathcal{L_D}$ according to Eqs. \eqref{eq:dqn_loss} and \eqref{eq:distillation_loss}; \\
        \hspace*{0.2cm} Perform a gradient descent step on Eq. \eqref{eq:joint_optimization_loss} {\em w.r.t.} $\theta$: \\ 
        \hspace*{1.8cm} $\theta\leftarrow\theta-\alpha\nabla_{\theta}(\mathcal{L}_{ori}+\lambda\mathcal{L_D})$.
        \STATE Reset $\theta^-=\theta$ and $\hat{\mathcal{C}}=\mathcal{C}$ every $C$ training steps.
    \ENDFOR
\end{algorithmic}
\end{algorithm}

\subsection{Joint Optimization Procedure}
To optimize a $Q$-function that can guide the agent to make proper decisions on each context without being adversely affected by catastrophic interference, we combine Eqs. \eqref{eq:dqn_loss} and \eqref{eq:distillation_loss} to form a joint optimization framework. Namely, we solve the catastrophic interference problem by the following optimization objective:
\begin{equation}
    \displaystyle
    \min_{\theta_\mathcal{S},\theta_\mathcal{T},\theta_\mathcal{F}}\mathcal{L}_{ori}(\theta_\mathcal{S},\theta_\mathcal{T})+\lambda \mathcal{L}_\mathcal{D}(\theta_\mathcal{S},\theta_\mathcal{F}),
    \label{eq:joint_optimization_loss}
\end{equation}
where $\lambda\in[0,1]$ is a coefficient to control the trade-off between the stability and plasticity of the neural network. 

The complete procedure is described in Algorithm \ref{alg:IQ}. The proposed method performs the context division in parallel to the training process without requiring additional data. For network training, to reduce the correlations with the target and ensure the stability of model training, the target network parameter $\theta^-$ is only updated by the $Q$-network parameter $\theta$ every $C$ steps and is held fixed between individual updates, as in DQN \cite{mnih2015human}. Similarly, we also adopt fixed target context centroids ($\hat{\mathcal{C}}$) to avoid a small amount of instability of states assignment step introduced by constantly updated context centroids ($\mathcal{C}$). To simplify the model implementation, we set the updating frequency of the target context centroids to be consistent with the target network.

\subsection{Random Encoders for High-dimensional State Space}
\begin{figure}[t]
  \centering
  \setlength{\abovecaptionskip}{5pt}
  {\includegraphics[width=0.97\linewidth]{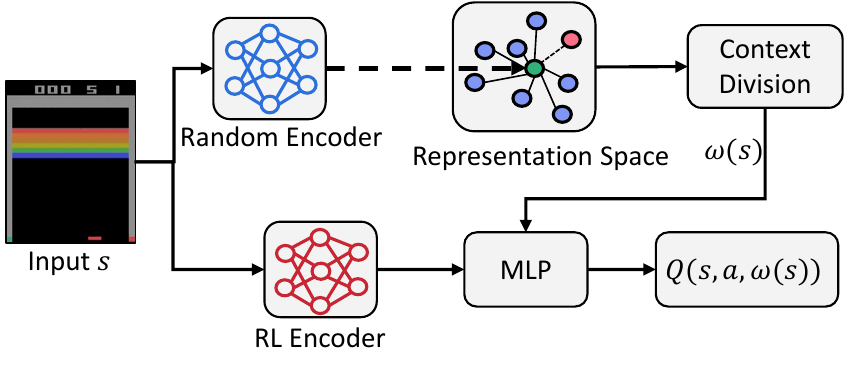}}
  \caption{Illustration of IQ-RE. The context division is performed in the low-dimensional representation space of a random encoder. A separate RL encoder is used to work with the MLP layers to estimate the value function.}
  \label{fig:IQ-RE}
\end{figure}

For high-dimensional state spaces, we propose to use random encoders for efficient context division, which can map high-dimensional inputs into low-dimensional representation spaces, overcoming the ``curse of dimensionality''. Although the original RL model already contains an encoder module, it is constantly updated and directly performing clustering in its representation space may introduce extra instability into context division. Therefore, on the basis of IQ, we exploit a dedicated random encoder module for dimension reduction. Fig. \ref{fig:IQ-RE} gives an illustration of this updated framework called IQ-RE in which the structure of the random encoder $f_{\theta_{re}}$ is consistent with the underlying RL encoder, but its parameter $\theta_{re}$ is randomly initialized and fixed throughout training. We provide the full procedure of IQ-RE in Appendix \ref{Random Encoder vs RL Trained Encoder}.

The main motivation of using random encoders arises from the observation that distances in the representation space of random encoder are adequate for finding similar states without any representation learning \cite{seo2021state}. That is, the representation space of a random encoder can effectively capture the information about the similarity among states without any representation learning (See Appendix \ref{Visualization Random Encoder}). Additional comparative experiments of IQ with the random encoder and the underlying RL trained encoder in Appendix \ref{Random Encoder vs RL Trained Encoder} further highlight the superiority of random encoders.

\section{Experiments and Evaluations}
In this section, we conduct comprehensive experiments on several standard benchmarks from OpenAI Gym\footnote{OpenAI Gym is a publicly available released implementation repository of RL environments: \href{https://github.com/openai/gym}{https://github.com/openai/gym}.} containing 4 classic control tasks and 6 high-dimensional complex Atari games to demonstrate the effectiveness of our method. 

\subsection{Datasets}
\textbf{Classic Control} \cite{brockman2016openai} contains 4 classic control tasks: {\em CartPole-v0}, {\em Pendulum-v0}, {\em CartPole-v1}, {\em Acrobot-v1}, where the dimensions of state spaces are in the range of 3 to 6. The maximum time steps are 200 for {\em CartPole-v0} and {\em Pendulum-v0}, and 500 for {\em CartPole-v1} and {\em Acrobot-v1}. Meanwhile, the reward thresholds used to determine tasks solved are $195.0$, $475.0$ and $-100.0$ for {\em CartPole-v0}, {\em CartPole-v1} and {\em Acrobot-v1}, respectively, while that for {\em Pendulum-v0} is not yet specified. We choose these commonly used domains as they are well-understood and relatively simple, suitable for highlighting the mechanism and verifying the effectiveness of our method in a straightforward manner.

\textbf{Atari Games} \cite{bellemare2013arcade} contain 6 image-level complex tasks: {\em Pong}, {\em Breakout}, {\em Carnival}, {\em Freeway}, {\em Tennis}, {\em FishingDerby}, where the observation is the screenshot represented by an RGB image of size $210\times160\times3$. We choose these domains to further demonstrate the scalability of our method on high-dimensional complex tasks that present significant challenges for existing baseline methods.

\subsection{Implementation}
\textbf{Network Structure}. For the 4 classic control tasks, we employ a fully-connected layer as the feature extractor and a fully-connected layer as the multi-head action scorer, following the network configuration for this type of tasks in OpenAI Baseline. For the 6 Atari games, we employ the similar convolution neural network as \cite{hessel2018rainbow}, \cite{castro18dopamine} for feature extracting and two fully-connected layers as the multi-head action scorer. More details can be found in Appendix \ref{Implementation Details}.

\textbf{Parameter Setting}. In IQ, there are two key parameters: $\lambda$ and $k$. To simplify parameter setting, we set $\lambda$ in accordance with the exploration proportion $\epsilon$ in all experiments: $\lambda=1-\epsilon$, due to the inverse relationship between them in training. In the early training, $\epsilon$ is close to 1, and the model is normally inaccurate with little interference, and a small $\lambda$ (close to 0) can promote plasticity construction of the model. Then, $\epsilon$ gradually approaches 0 during the subsequent training, and the model has learned more useful information, while interference is also likely to occur. Consequently, smoothly increasing $\lambda$ is needed to ensure plasticity while avoiding interference. Meanwhile, we set $k$ to 3 for all classic control tasks, and 4 for all Atari games. In IQ-RE, we set the extra parameter $d$ to 50 as in \cite{seo2021state}, which has been shown to be both efficient and effective. Other parameter settings can be found in Appendix \ref{Implementation Details}. For classic control tasks, we evaluate the training performance using the average episode returns every 10K time steps for {\em CartPole-v0}, {\em Pendulum-v0}, and 20K time steps for {\em CartPole-v1}, {\em Acrobot-v1}. For Atari games, the time step range for performance evaluation is 200K. All experiment results reported are the average episode returns over 5 independent runs.

\subsection{Baselines}
We evaluate our method in comparison to following state-of-the-art baseline methods for single-task RL:
\begin{itemize}
\item {\bf DQN} \cite{mnih2015human} is a representative algorithm of Deep RL, which reduces catastrophic interference using experience replay and fixed target networks. We use the DQN agent implemented in OpenAI Baselines.
\item {\bf Rainbow} \cite{hessel2018rainbow} is the upgraded version of DQN containing six extensions, including a prioritized replay buffer \cite{schaul2016prioritized}, n-step returns \cite{sutton2018reinforcement}, Adam optimizer \cite{kingma2014adam} and distributional learning \cite{bellemare2017distributional} for stable RL training. The Rainbow agent is implemented in Google's Dopamine framework\footnote{Dopamine is a research framework developed by Google for fast prototyping of reinforcement learning algorithms: https://github.com/google/dopamine.} \cite{castro18dopamine}.
\item {\bf SRNN} \cite{liu2019utility} employs a distributional regularizer to induce sparse representations in neural networks to avert catastrophic interference in the single-task RL. Here, we implement it in the form of fully online training.
\item {\bf DSOM} \cite{lo2019overcoming} introduces a DSOM module to control the activation of the representation output layer to achieve local optimization. We reproduce it with reference to the original DSOM implementation\footnote{Dynamic Self-Organized maps: https://github.com/rougier/dynamic-som.}.
\item {\bf TCNN} \cite{ghiassian2020improving} aims to remap the inputs to a higher-dimensional space using tile coding to sparsify the input features, reducing activation overlap. We adopt its implementation in \cite{pan2019fuzzy}.
\end{itemize}

In the experiments, we firstly use DQN as the underlying RL method to evaluate the effectiveness of IQ in comparison to all baselines on classic control tasks. We then extend them to high-dimensional Atari games to further validate the scalability of IQ. Note that TCNN suffers from scalability issues for benchmarks with high dimensionality as it increases the dimension of input to the neural network, and DSOM has not been applied to solve any high-dimensional RL tasks in \cite{lo2019overcoming}, whose implementation details are unclear. Therefore, we evaluate IQ-RE only in comparison to DQN and SRNN on the Atari games. Additionally, we also implement IQ-RE with Rainbow, to illustrate that our method is highly flexible and can be incorporated into various existing value-based RL models.

\subsection{Evaluation Metrics}
Following the convention in previous studies \cite{mnih2016asynchronous,bellemare2017distributional,espeholt2018impala,hessel2018rainbow,fedus2020catastrophic}, we employ the average training episode returns $\mathcal{R}_T$ to evaluate our method during training: 
\begin{equation}
    \displaystyle
        \mathcal{R}_T = \frac{1}{M}\sum_{i=1}^{M}\sum_{j=0}^{J_i} R_{ij}
    \label{eq:TER}
\end{equation}
where $M$ is the number of episodes experienced within each evaluation period; $J_i$ is the total time steps in episode $i$; $R_{ij}$ is the reward received at time step $j$ in episode $i$.

\begin{figure*}[t]
    \centering
    \setlength{\abovecaptionskip}{5pt}
    \subfigure[$N=50,000$ \label{fig:Classic_Control_a}]{
        \begin{minipage}[b]{0.97\textwidth}
			\includegraphics[width=0.24\textwidth]{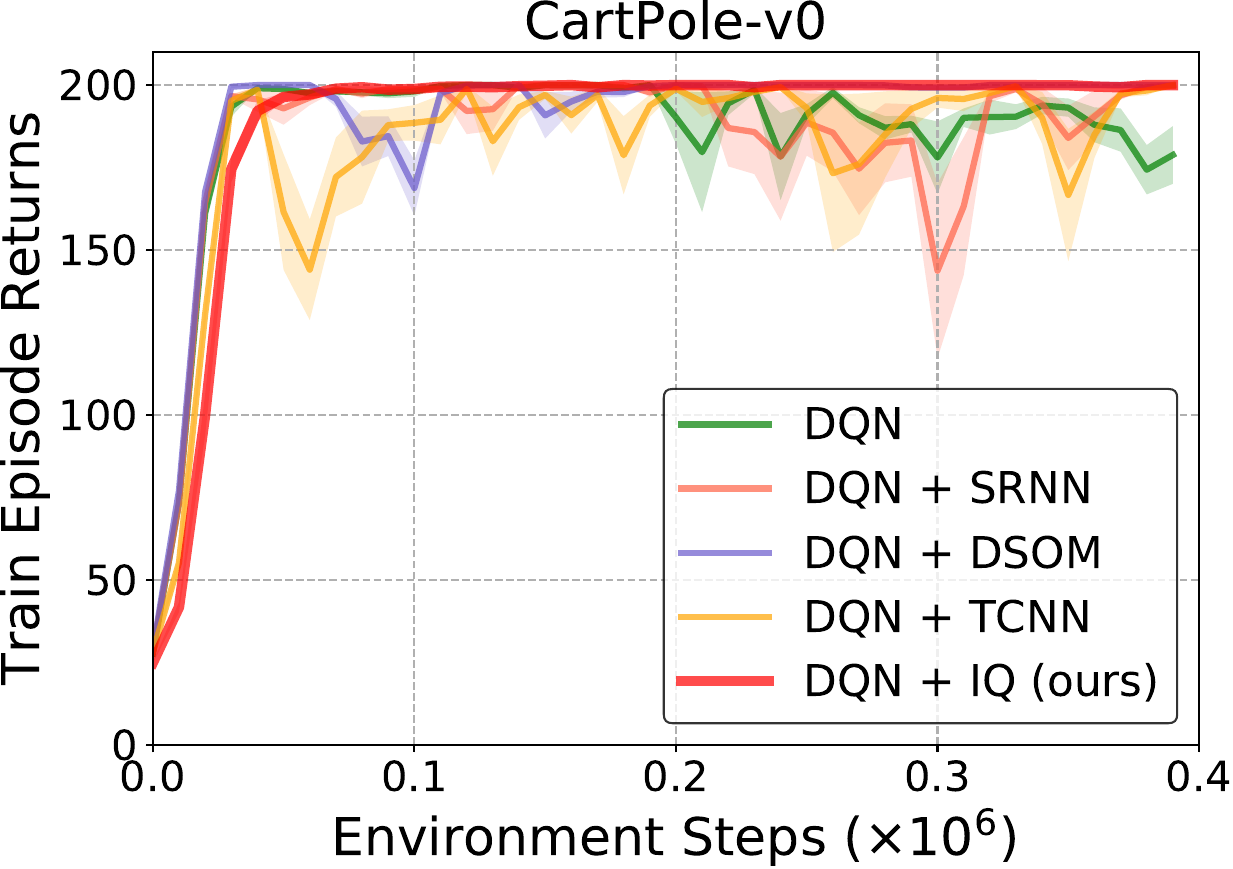}
			\includegraphics[width=0.252\textwidth]{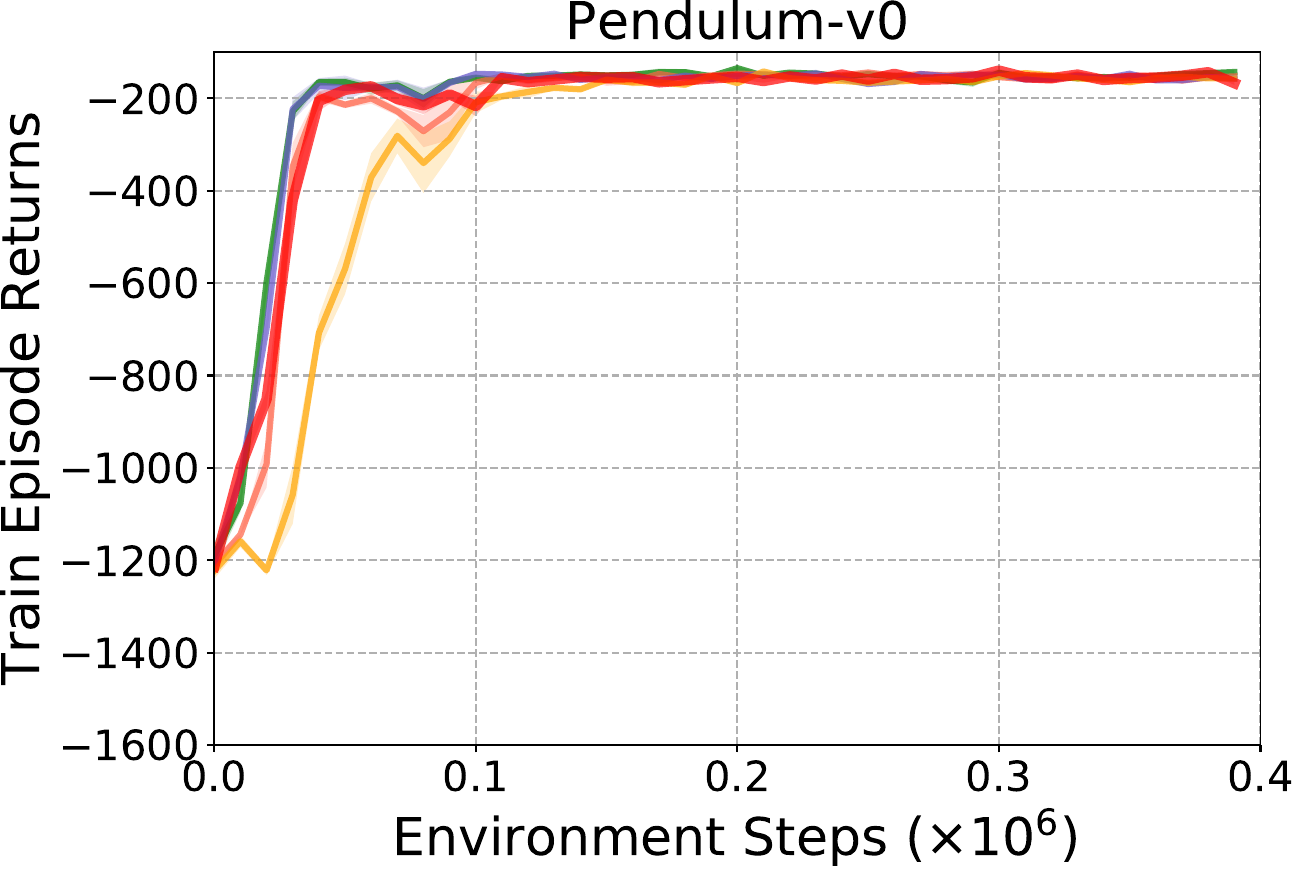}
			\includegraphics[width=0.24\textwidth]{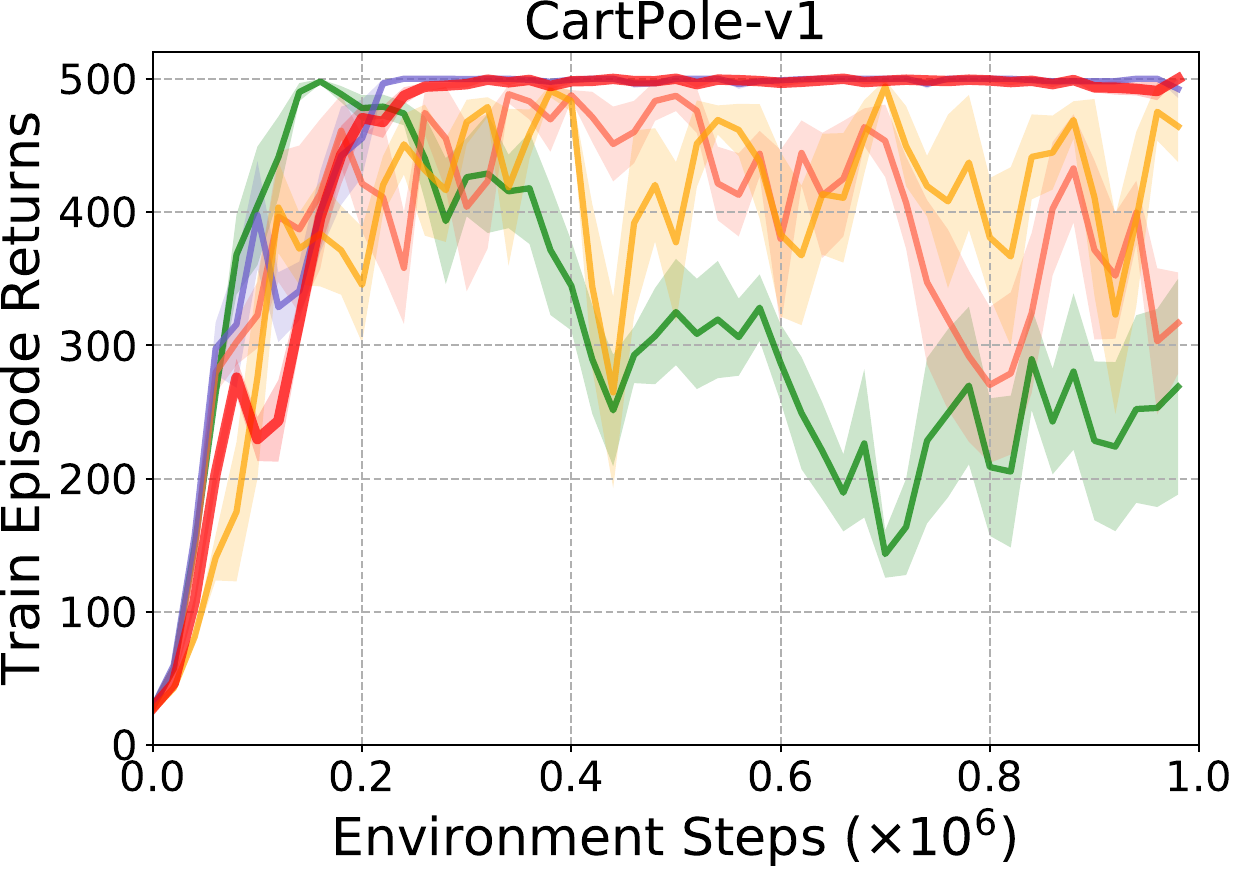}
			\includegraphics[width=0.247\textwidth]{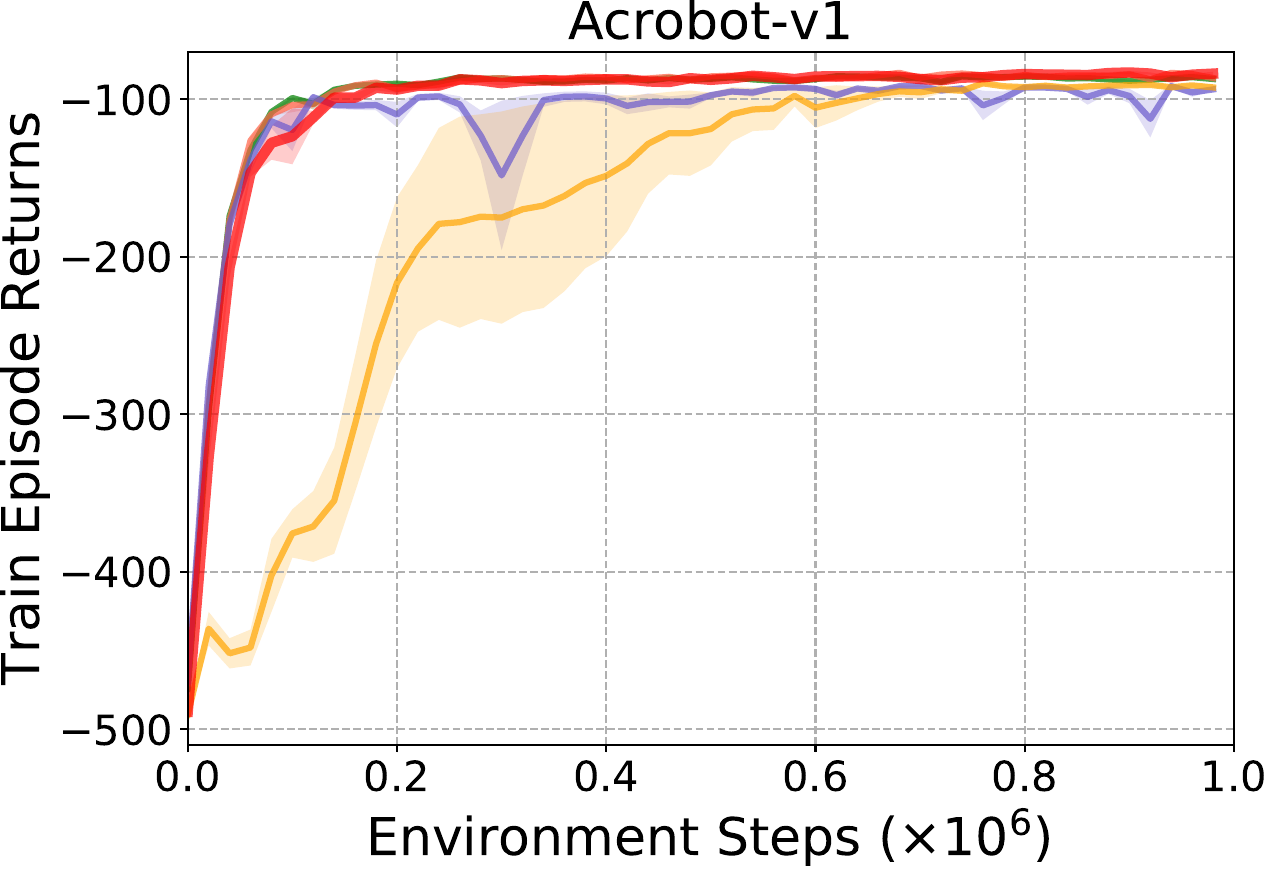}
		\end{minipage}
    }\\
    \subfigure[$N=100$ \label{fig:Classic_Control_b}]{
        \begin{minipage}[b]{0.97\textwidth}
			\includegraphics[width=0.24\textwidth]{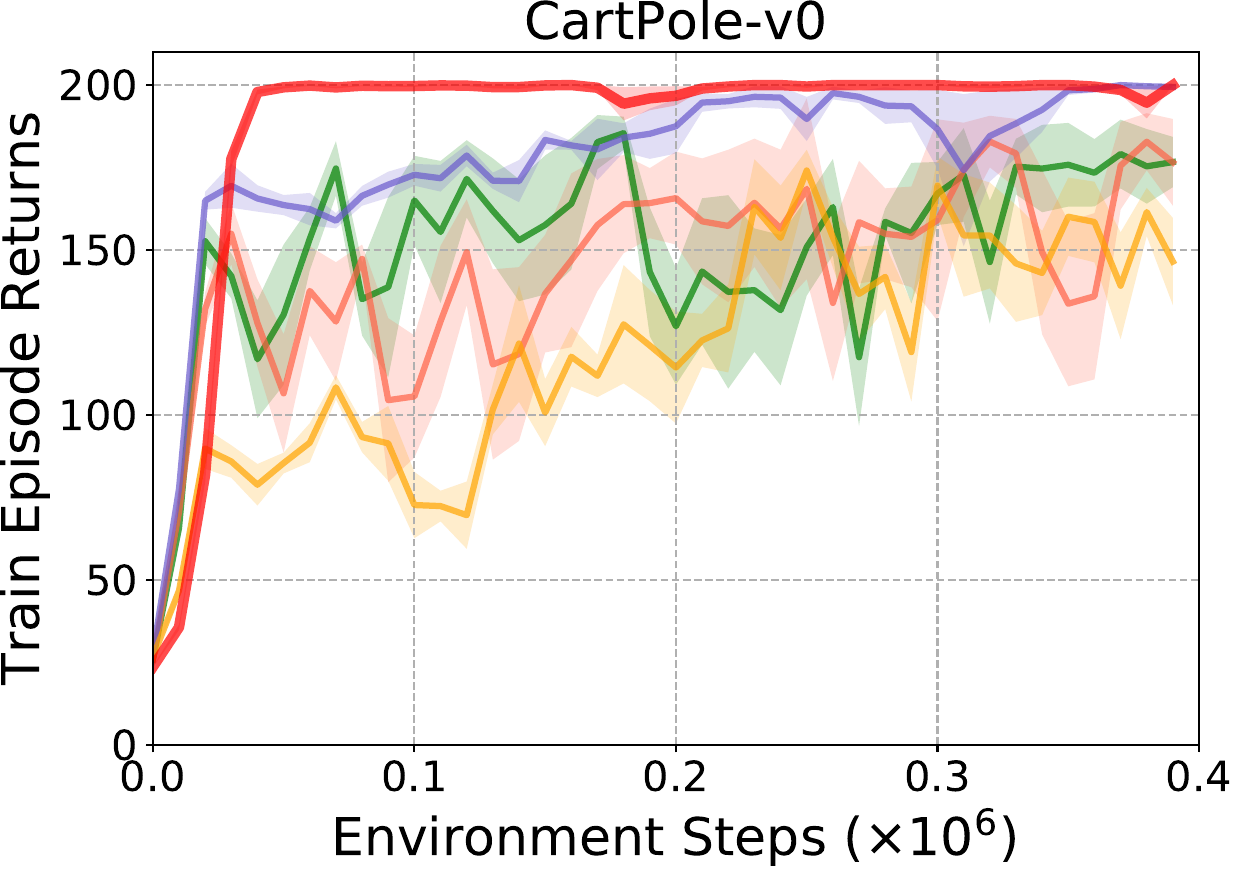}
			\includegraphics[width=0.252\textwidth]{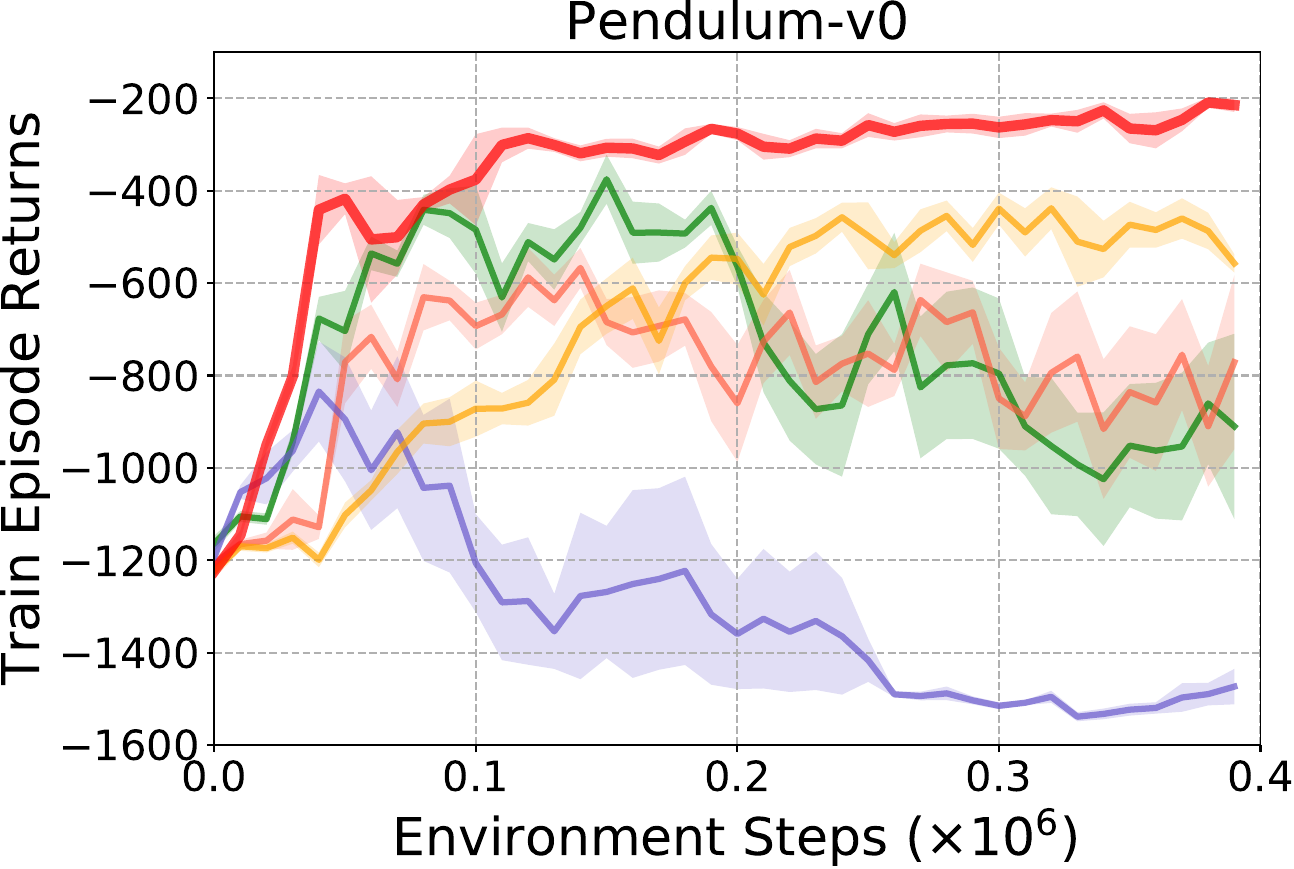}
			\includegraphics[width=0.24\textwidth]{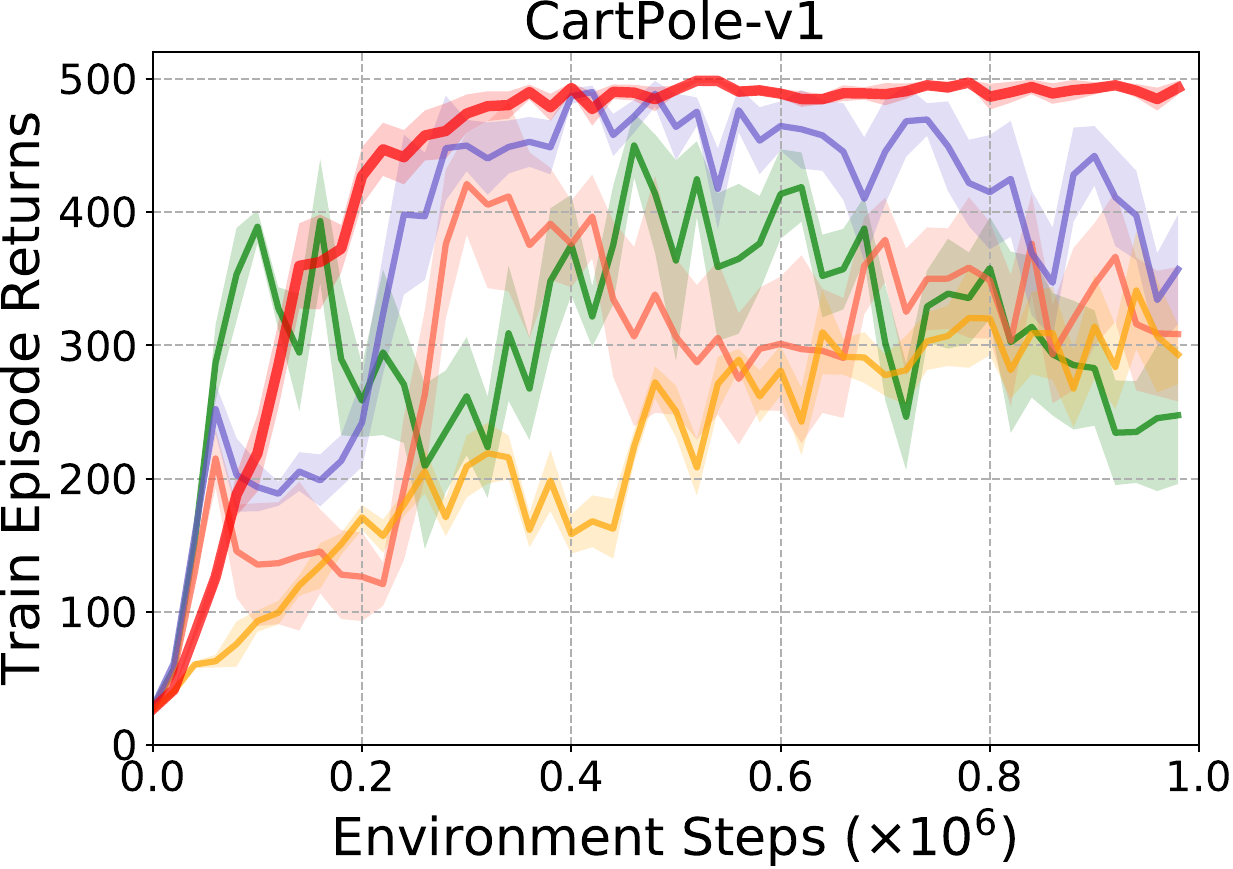}
			\includegraphics[width=0.247\textwidth]{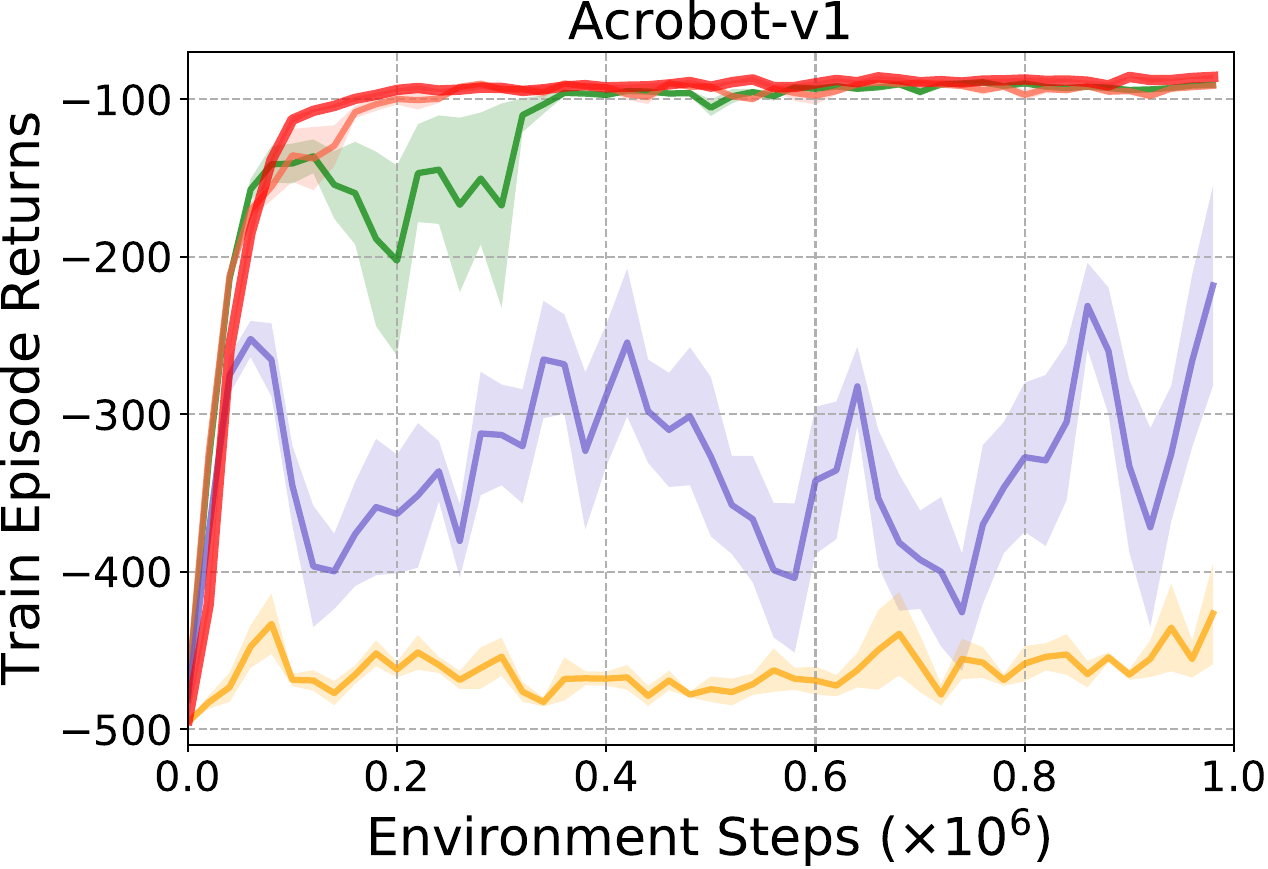}
		\end{minipage}
    }\\
    \subfigure[$N=1$ \label{fig:Classic_Control_c}]{
        \begin{minipage}[b]{0.97\textwidth}
			\includegraphics[width=0.24\textwidth]{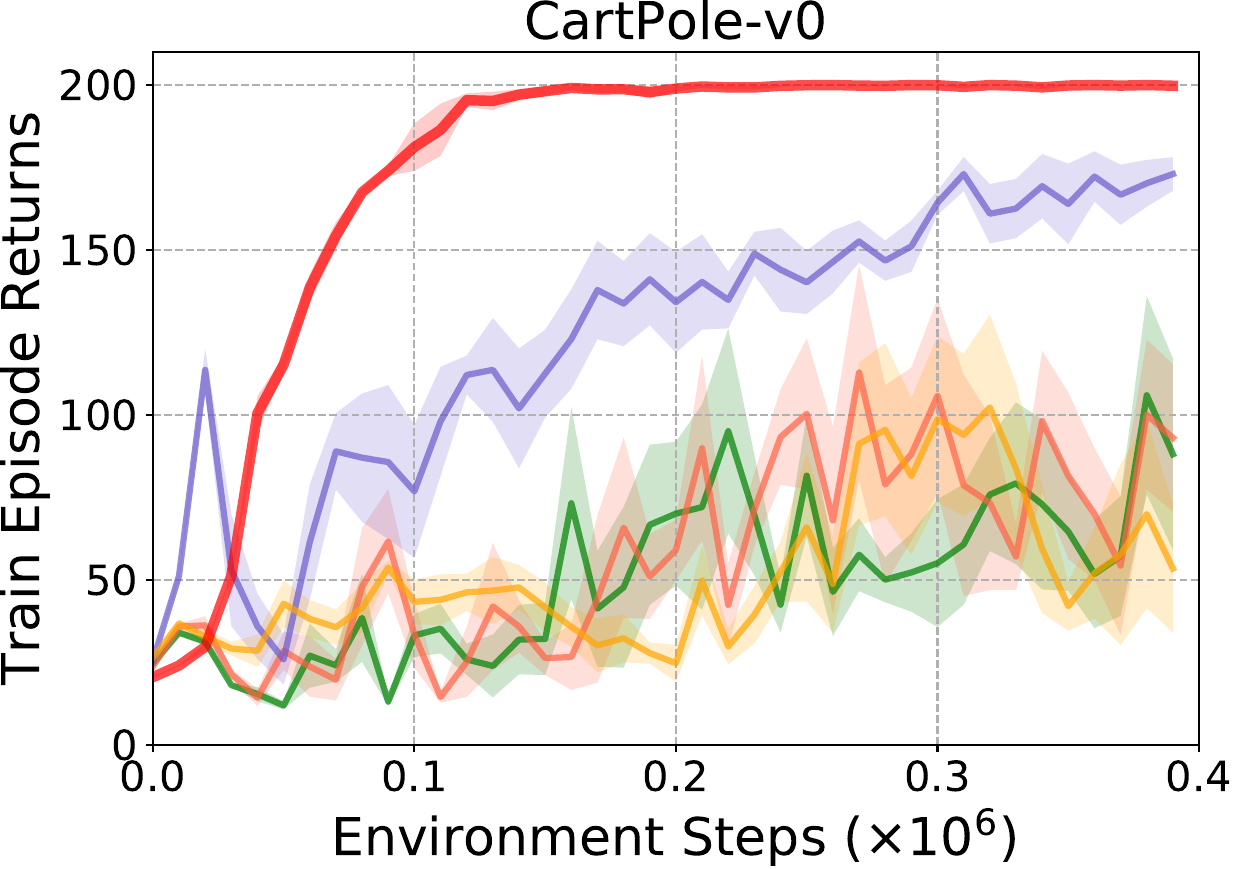}
			\includegraphics[width=0.252\textwidth]{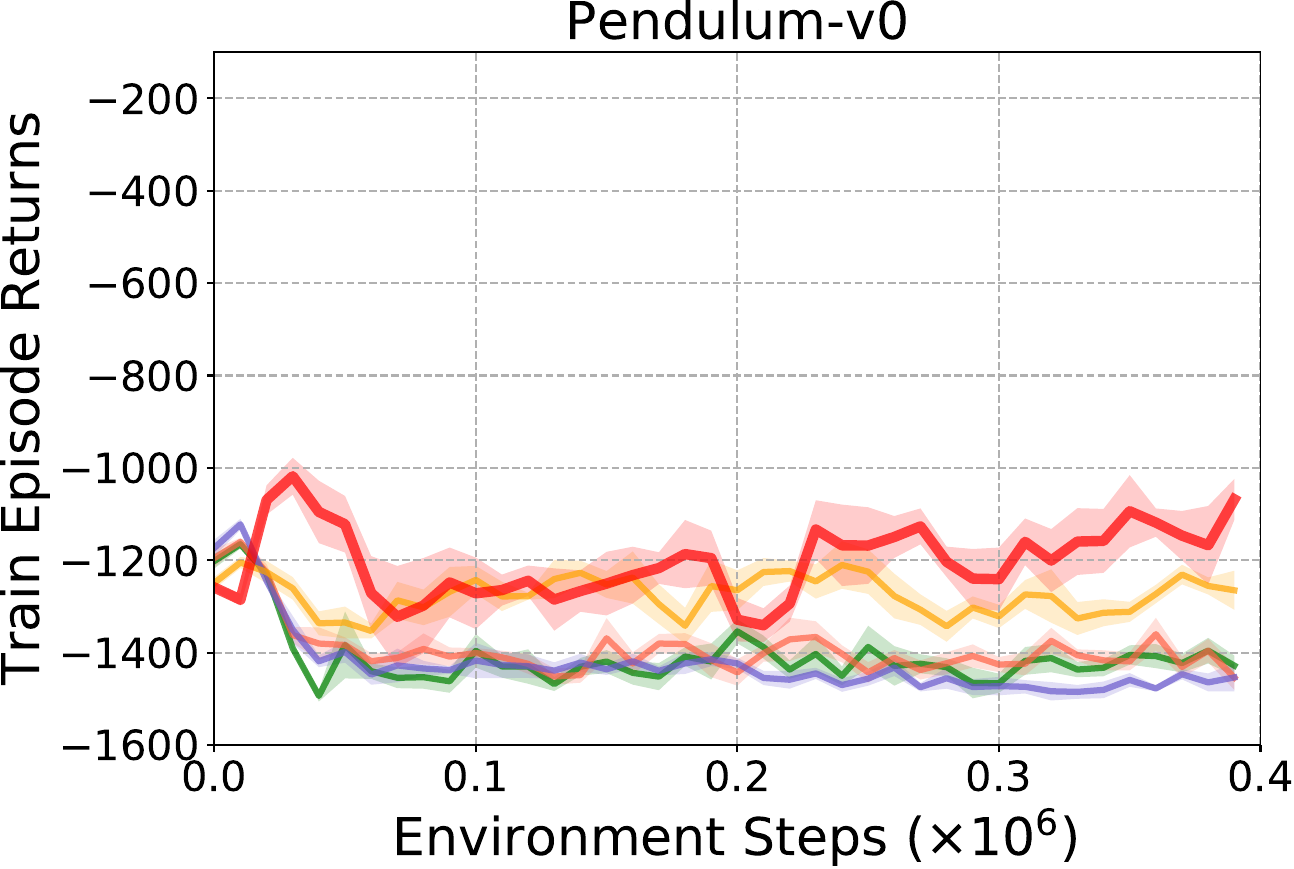}
			\includegraphics[width=0.24\textwidth]{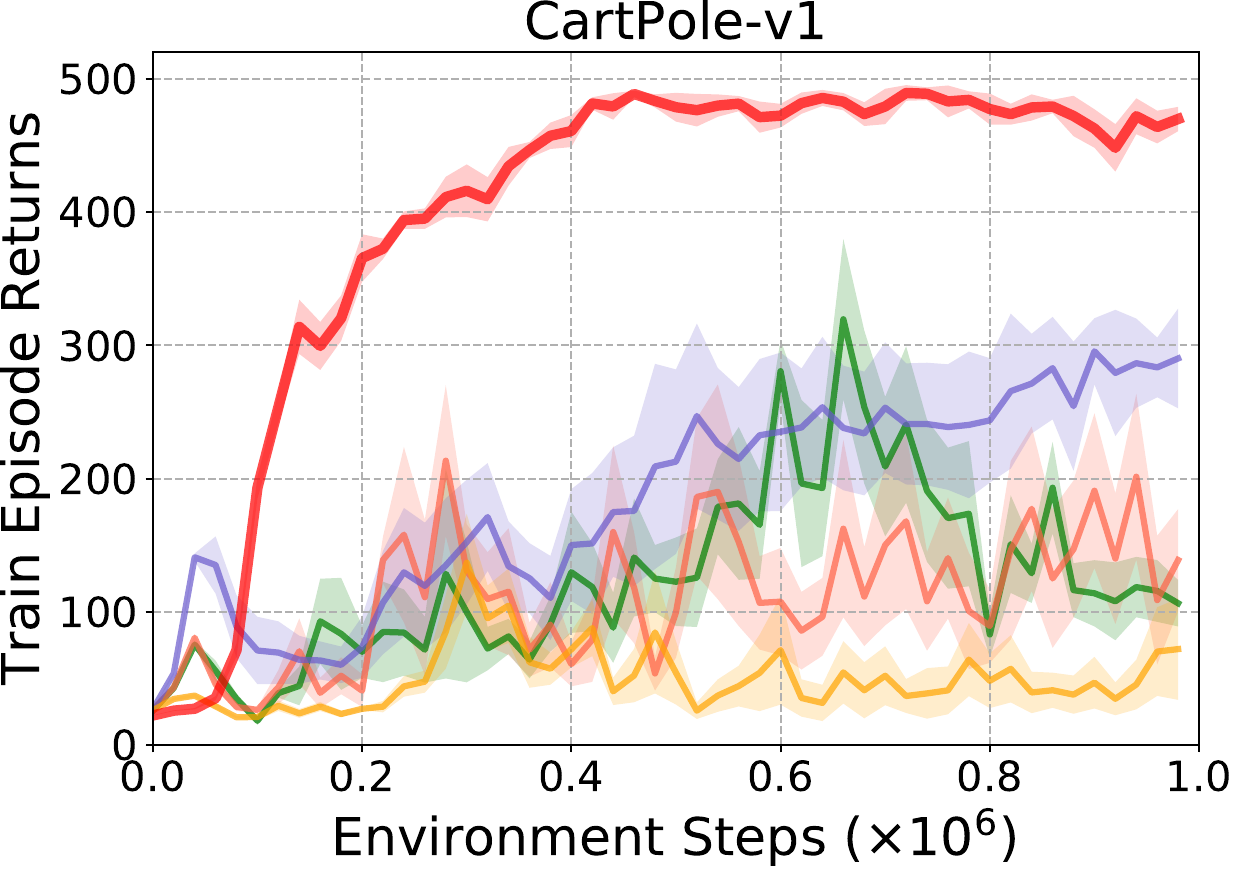}
			\includegraphics[width=0.247\textwidth]{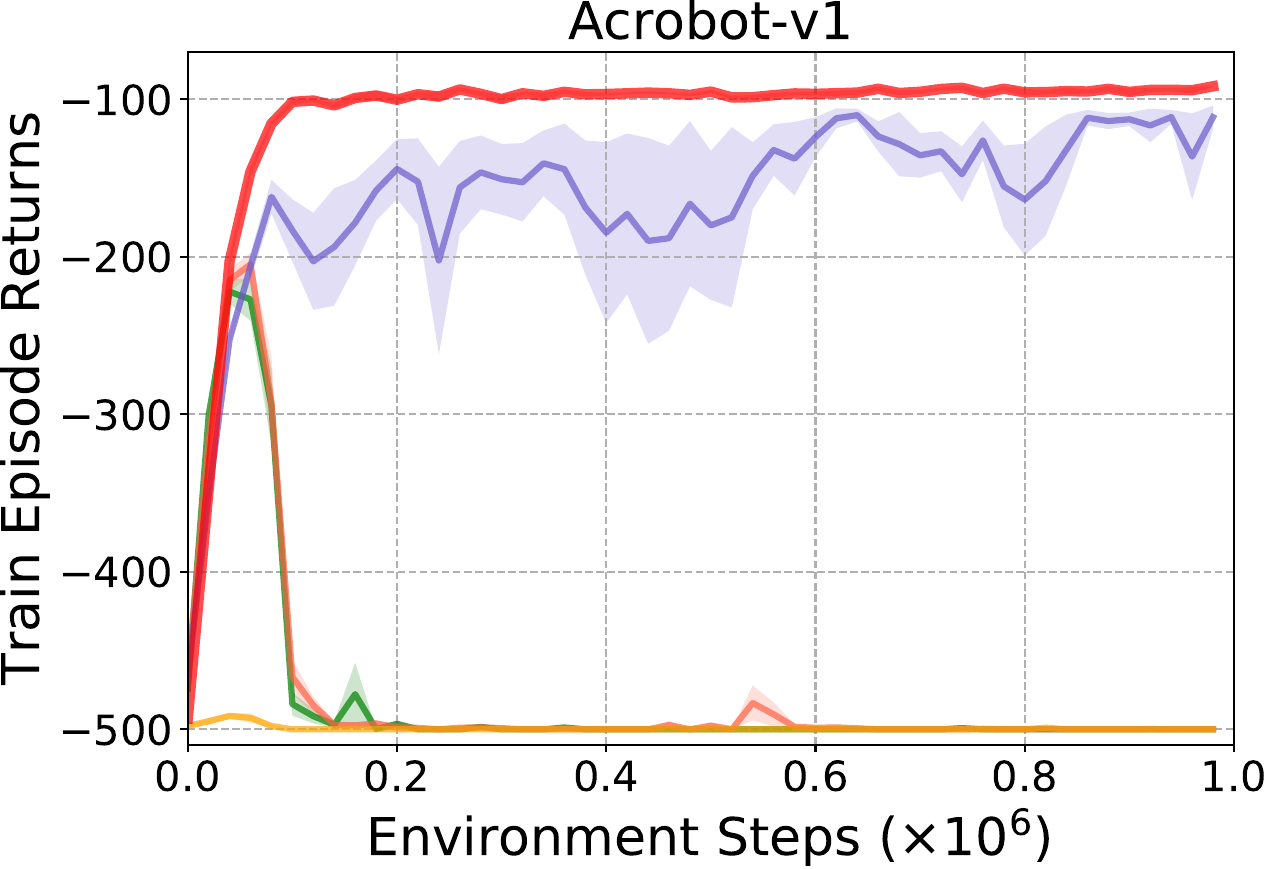}
		\end{minipage}
    }
    \caption{Learning curves on classic control tasks with different replay buffer capacities $N$. Here and in related figures below, the solid lines and shaded regions denote the means and standard deviations of rewards, respectively, across five runs.}
    \label{fig:Classic_Control}
\end{figure*}

\begin{table*}[t]
\centering
\setlength{\tabcolsep}{0.4mm}
\caption{Numerical results in terms of the highest cumulative return achieved during training of all methods implemented\\ in the classic control tasks (based on the performance of five runs in Fig. \ref{fig:Classic_Control}. Here and in\\ related tables below, the best performance is marked in boldface.)}
\label{table:highest_cumulative_score_classic_control}
\begin{tabular}{c|ccc|ccc|ccc|ccc|ccc}
\toprule
\specialrule{0em}{1pt}{1pt}
Method                    & \multicolumn{3}{c|}{DQN}                 & \multicolumn{3}{c|}{DQN + SRNN}         & \multicolumn{3}{c|}{DQN + DSOM}             & \multicolumn{3}{c|}{DQN + TCNN}        & \multicolumn{3}{c}{DQN + IQ (ours)}                  \\ \hline
N                         & 1         & 100      & 50,000            & 1         & 100      & 50,000           & 1         & 100           & 50,000          & 1          & 100      & 50,000         & 1              & 100           & 50,000              \\ \hline
\textit{CartPole-v0}      & $106.0$   & $185.4$  & \bm{$200.0$}      & $112.9$   & $182.9$  & \bm{$200.0$}     & $173.0$   & \bm{$200.0$}  & \bm{$200.0$}     & $102.3$    & $174.1$  & \bm{$200.0$}   & \bm{$200.0$}   & \bm{$200.0$}  & \bm{$200.0$}          \\
\textit{Pendulum-v0}      & $-1165.3$ & $-375.2$ & \bm{$-134.9$}     & $-1160.3$ & $-567.1$ & $-144.2$         & $-1122.1$ & $-834.7$      & $-146.2$        & $-1204.8$  & $-438.0$ & $-142.0$       & \bm{$-1018.1$} & \bm{$-209.1$} & $-140.0$              \\
\textit{CartPole-v1}      & $188.4$   & $421.5$  & $497.7$           & $250.1$   & $445.6$  & $498.6$          & $295.5$   & $490.0$       & $500.0$         & $136.8$    & $341.1$  & $494.2$        & \bm{$489.3$}   & \bm{$498.5$}  & \bm{$500.0$}           \\
\textit{Acrobot-v1}       & $-222.1$  & $-89.2$  & $-85.4$           & $-204.7$  & $-89.4$  & $-83.2$          & $-110.0$  & $-218.2$      & $-91.4$         & $-491.5$   & $-426.5$ & $-89.8$        & \bm{$-91.7$}   & \bm{$-85.7$}  & \bm{$-82.9$} \\
\specialrule{0em}{1pt}{1pt}
\bottomrule
\end{tabular}
\end{table*}

\begin{figure*}[t]
    \flushright
    \setlength{\abovecaptionskip}{5pt}
    \subfigure[$N=1,000,000$ \label{fig:Atari_games_a}]{
        \begin{minipage}[b]{0.48\textwidth}
        \flushright
			\includegraphics[width=0.49\textwidth]{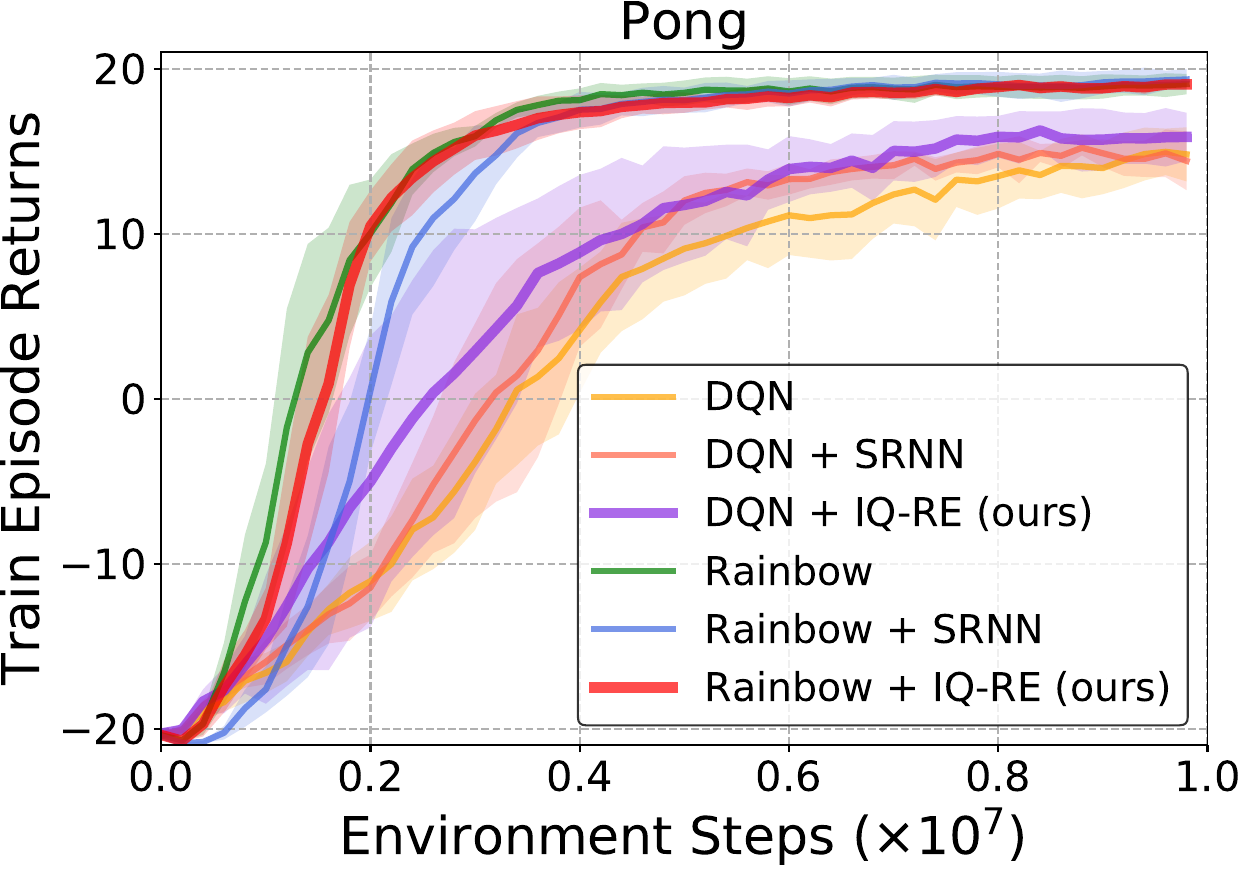}
			\includegraphics[width=0.49\textwidth]{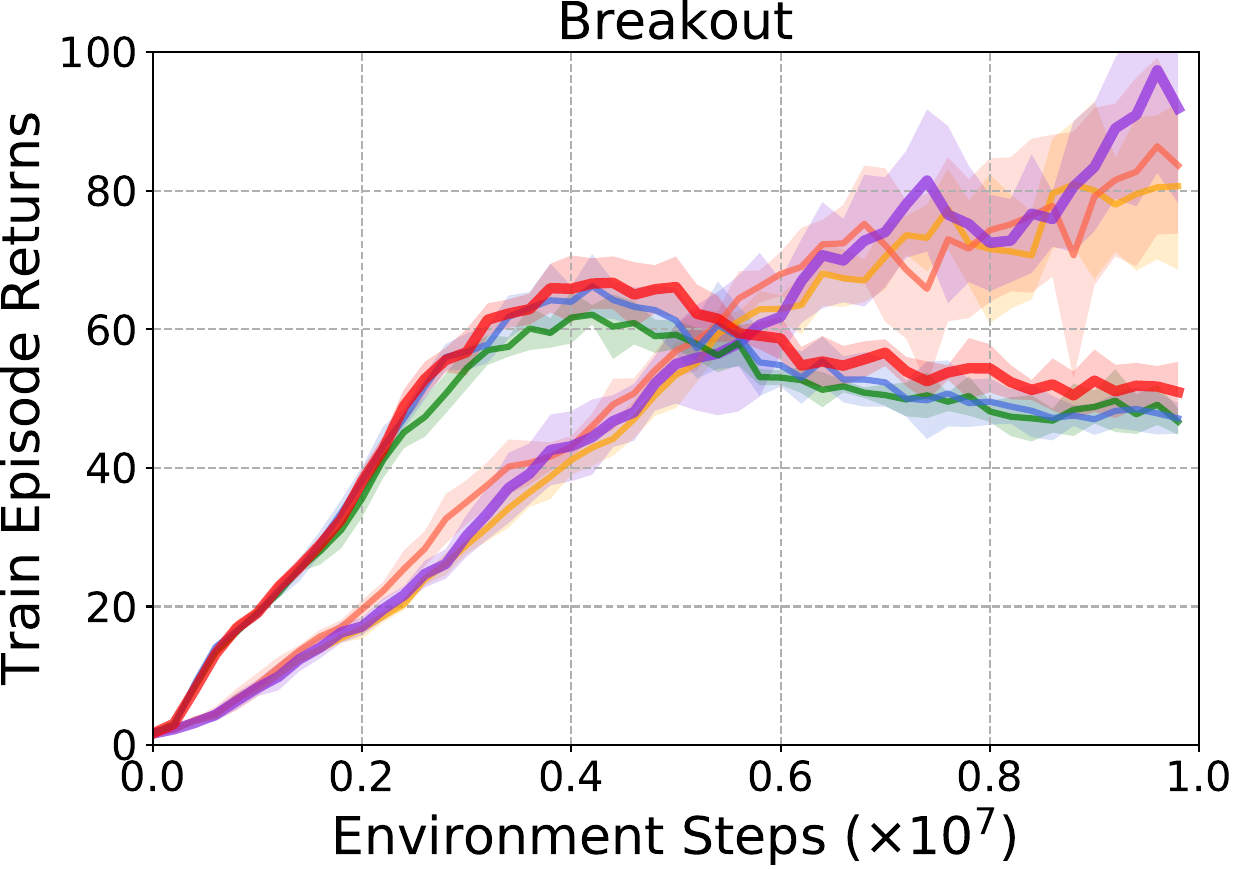} \\
			\vspace{0.3cm}
			\includegraphics[width=0.485\textwidth]{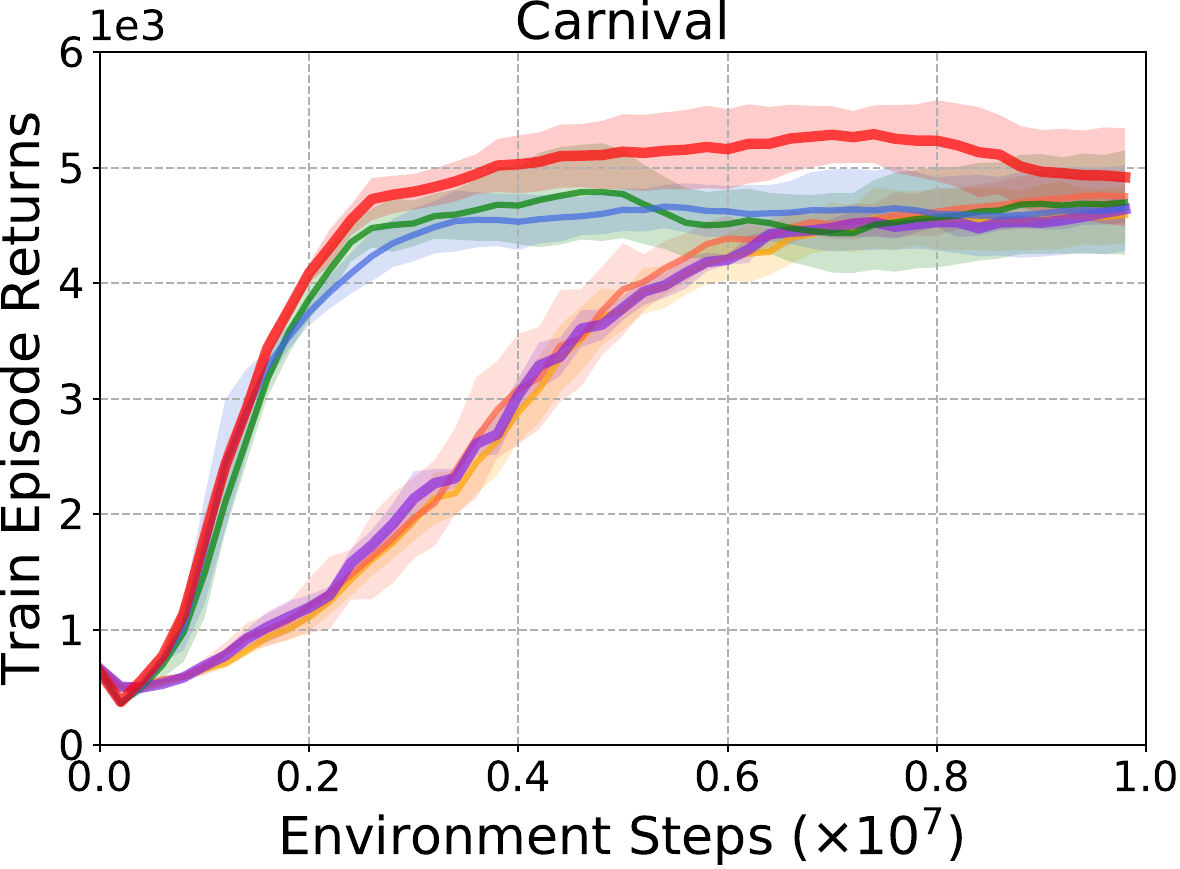}
			\includegraphics[width=0.495\textwidth]{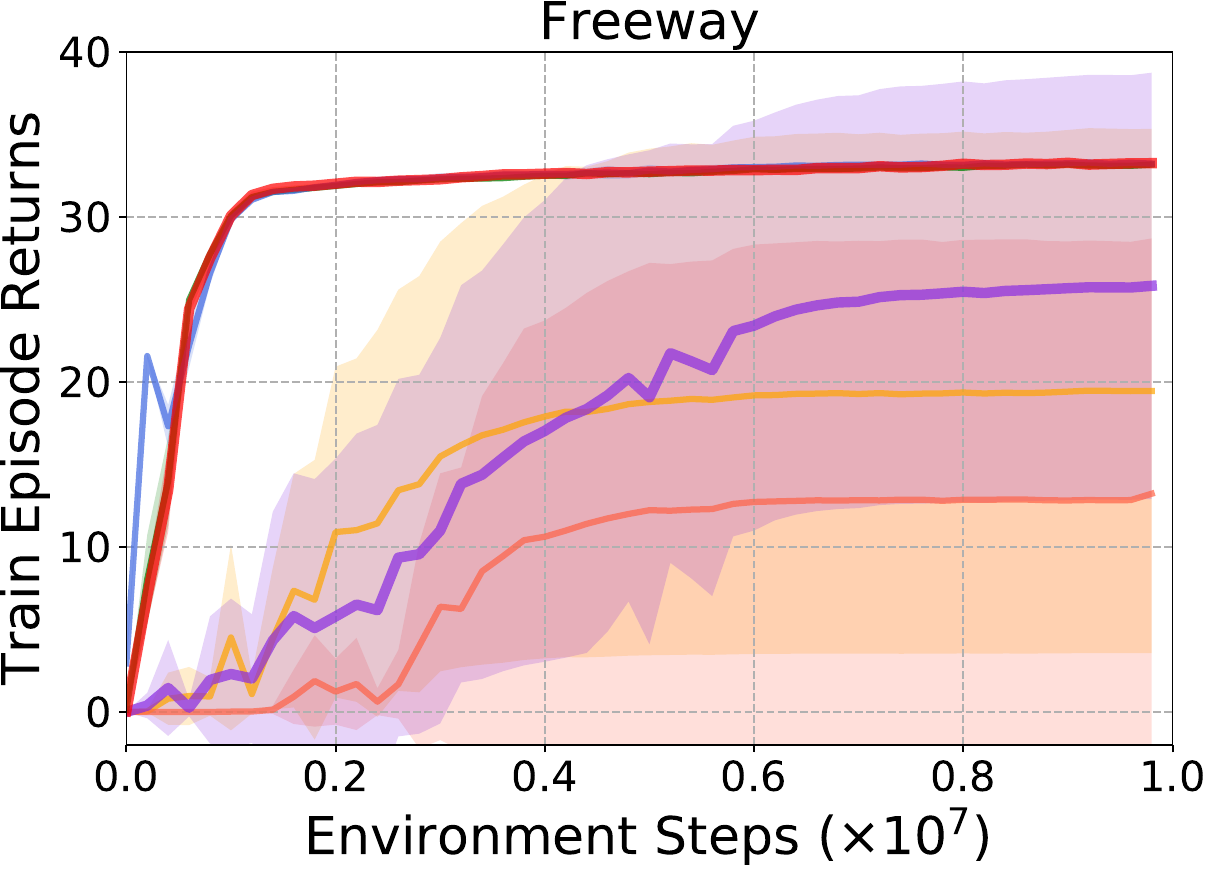} \\
			\vspace{0.3cm}
			\includegraphics[width=0.485\textwidth]{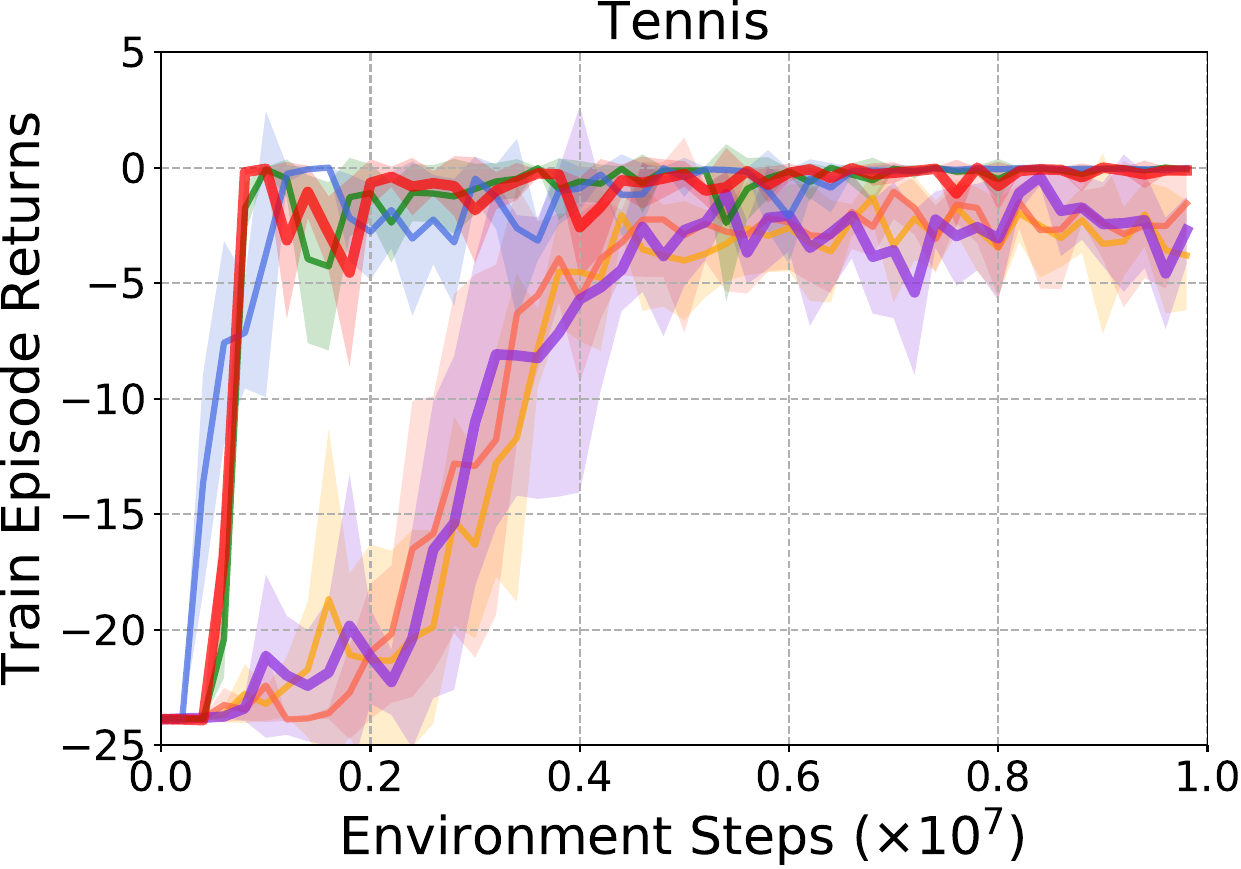}
			\includegraphics[width=0.495\textwidth]{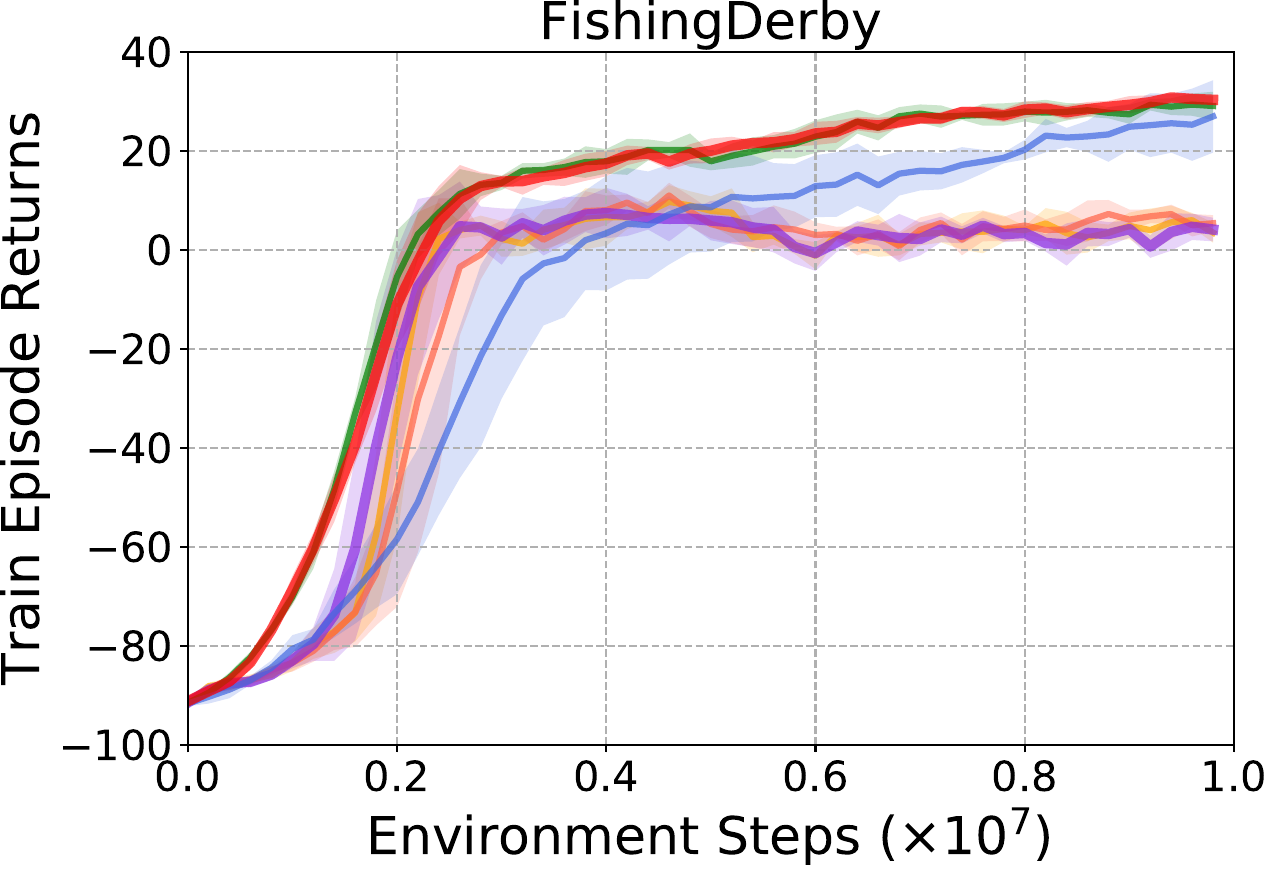}
		\end{minipage}
    }
    \subfigure[$N=10,000$ \label{fig:Atari_games_b}]{
        \begin{minipage}[b]{0.48\textwidth}
        \flushright
			\includegraphics[width=0.49\textwidth]{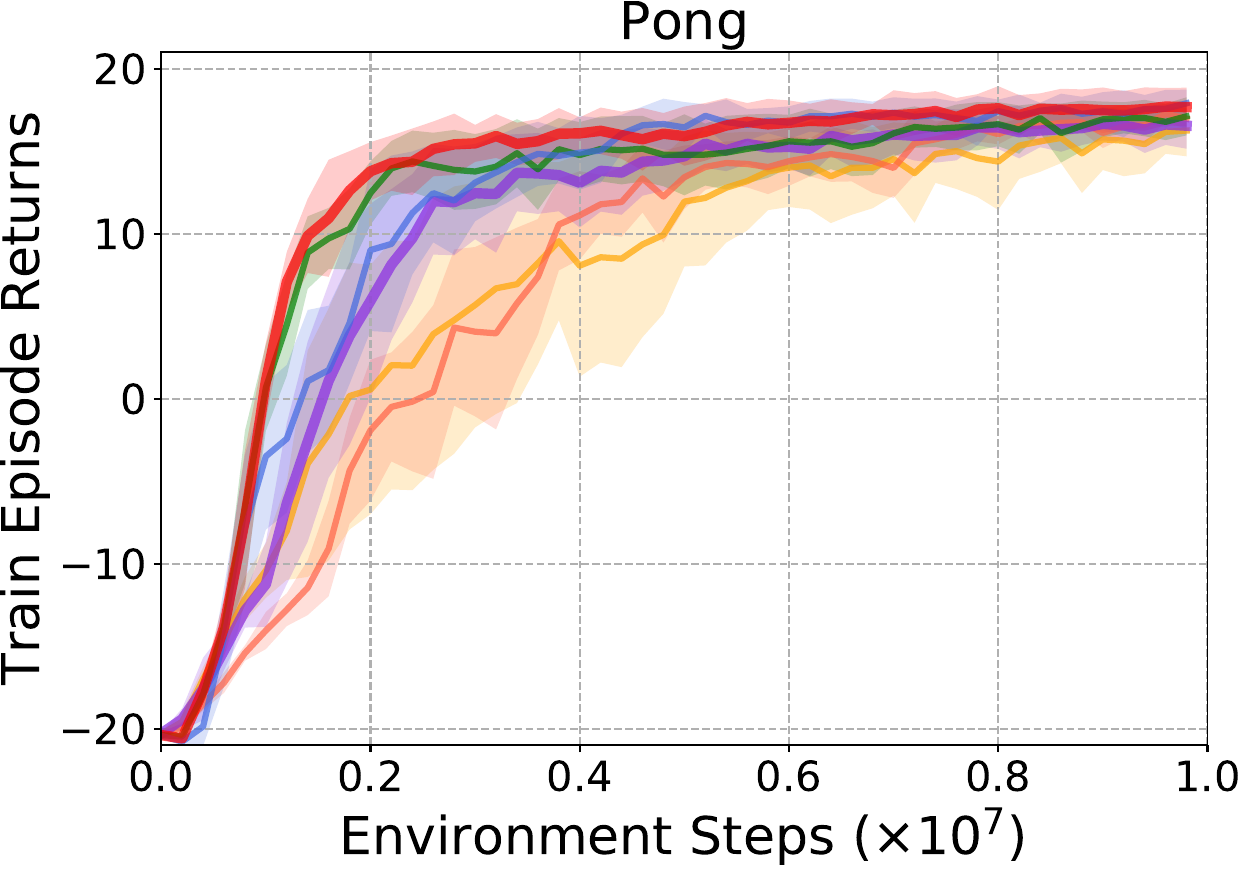}
			\includegraphics[width=0.49\textwidth]{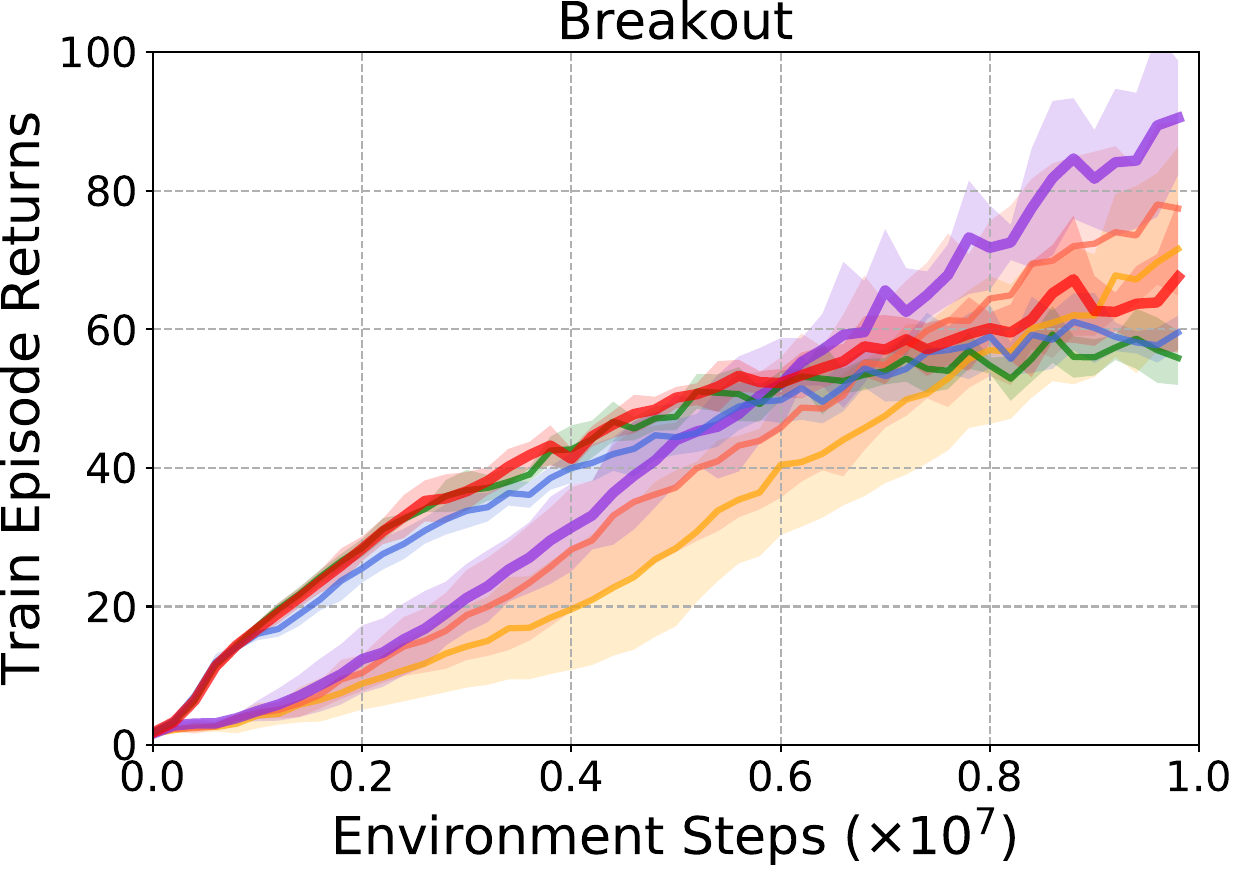} \\
			\vspace{0.3cm}
			\includegraphics[width=0.485\textwidth]{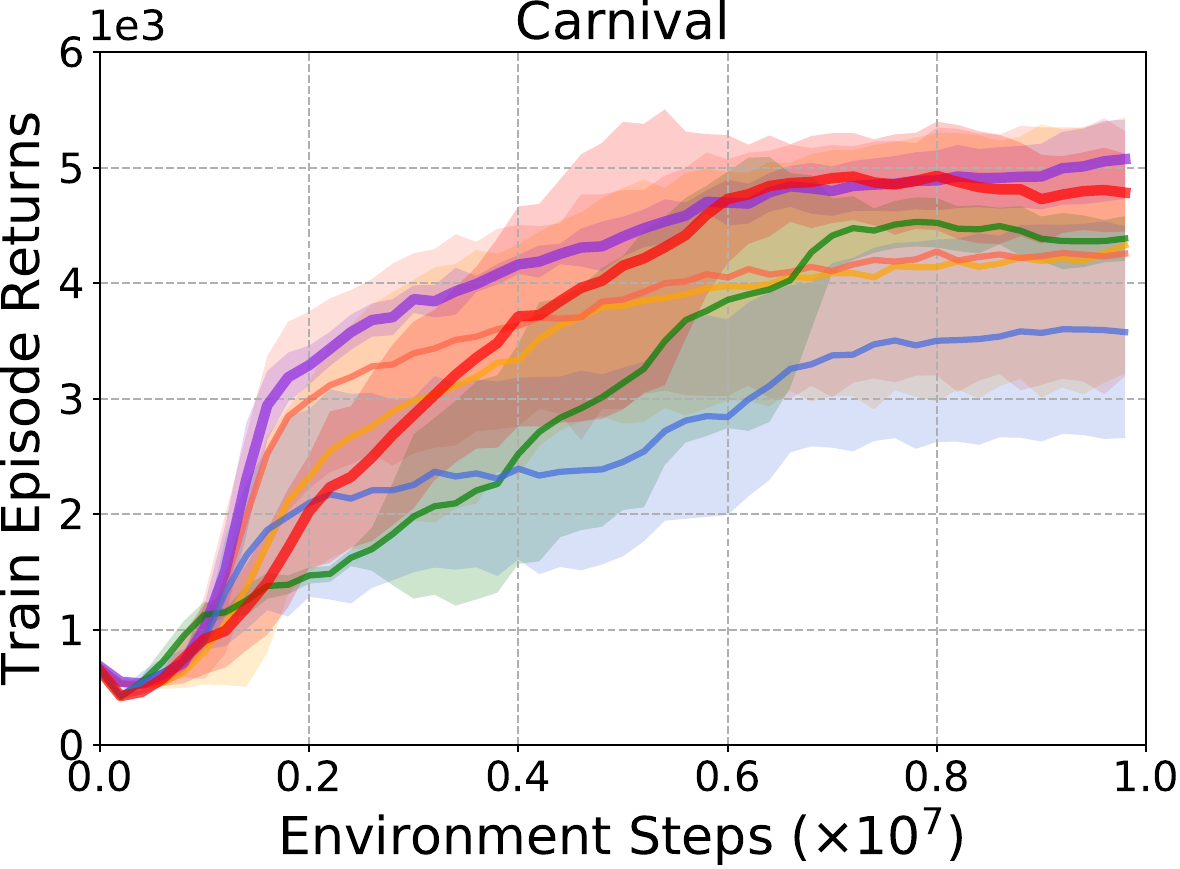}
			\includegraphics[width=0.495\textwidth]{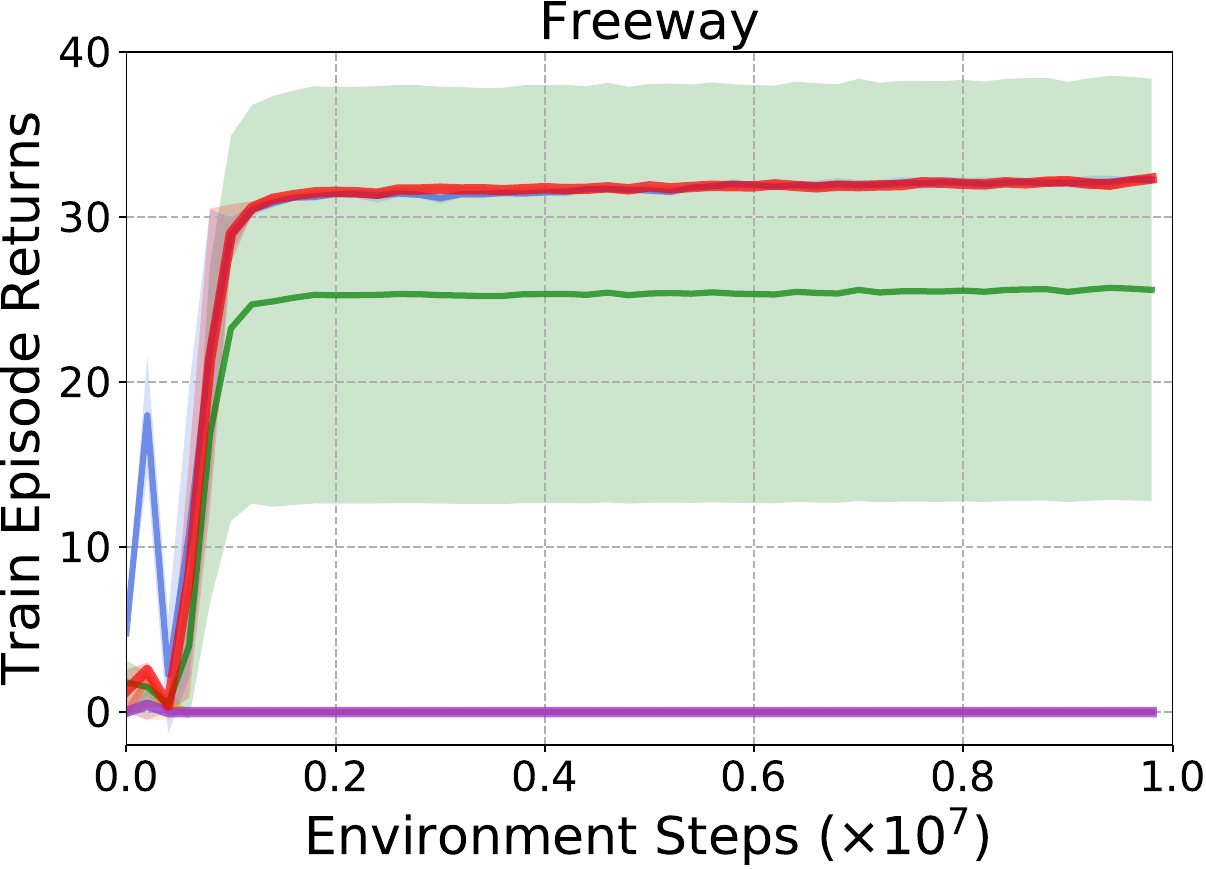} \\
			\vspace{0.3cm}
		    \includegraphics[width=0.485\textwidth]{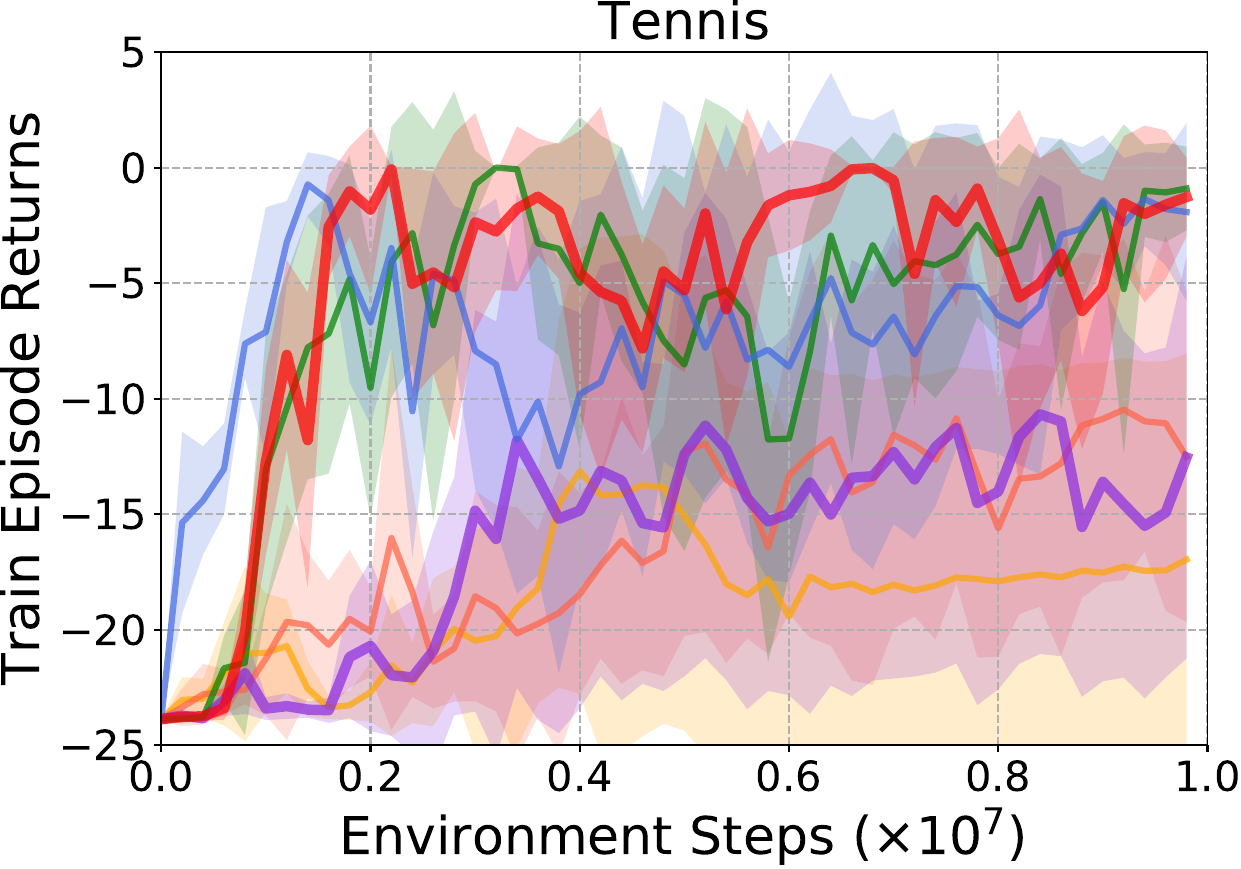}
			\includegraphics[width=0.495\textwidth]{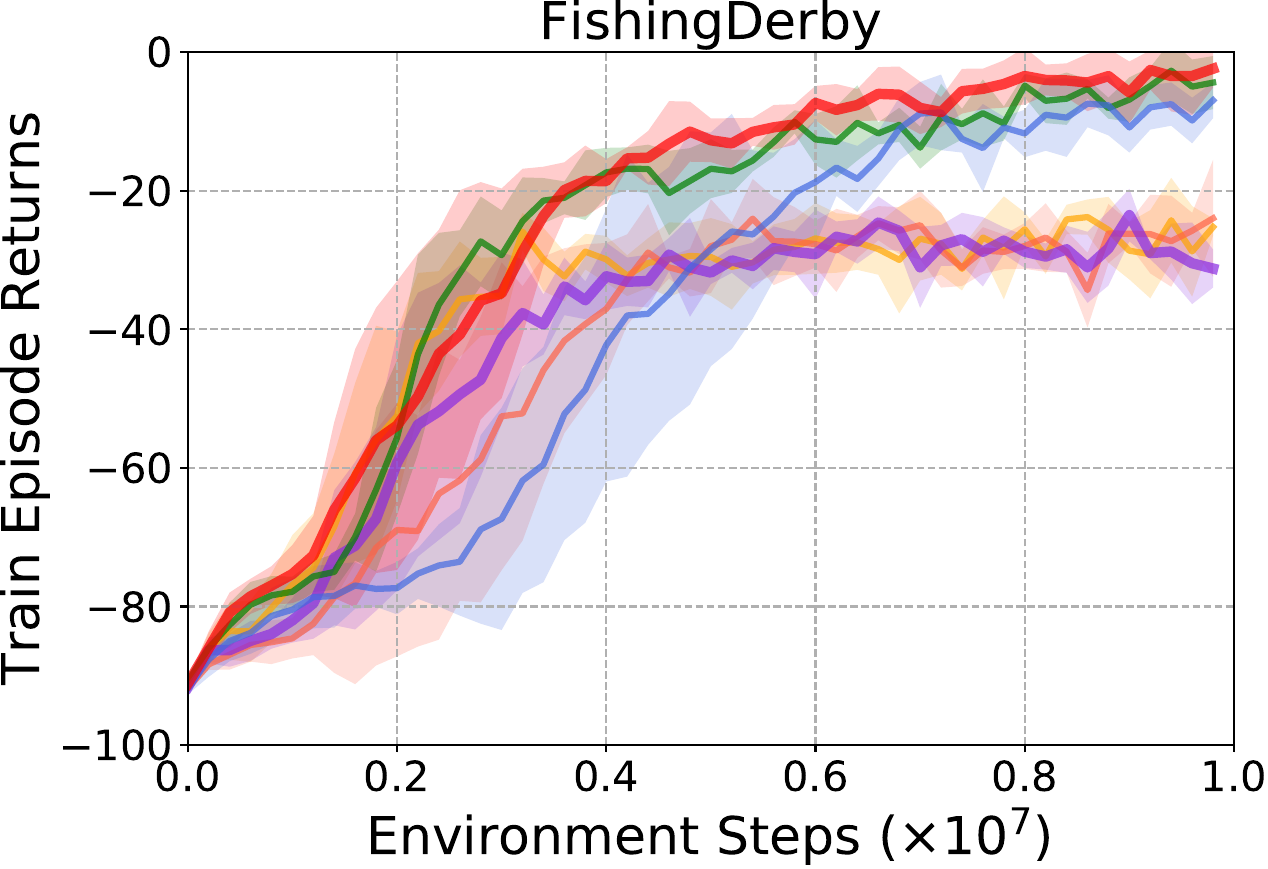}
		\end{minipage}
    }
    \caption{Learning curves on Atari games with different replay buffer capacities $N$. It is worth noting that the green line and the red line of Freeway in (a) and the purple, orange and pink lines of Freeway in (b) are overlapping, respectively.}
    \label{fig:Atari_games}
\end{figure*}

\begin{table*}[t]
\centering
\setlength{\tabcolsep}{1.3mm}
\caption{Numerical results in terms of the highest cumulative return achieved during training of all methods\\ implemented in the Atari games (based on the performance of five runs in Fig. \ref{fig:Atari_games}.)}
\label{table:highest_cumulative_score_atari}
\begin{tabular}{c|cccccc|cccccc}
\toprule
\specialrule{0em}{1pt}{1pt}
Method                & \multicolumn{2}{c}{DQN}      & \multicolumn{2}{c}{DQN + SRNN} & \multicolumn{2}{c|}{DQN + IQ-RE (ours)}    & \multicolumn{2}{c}{Rainbow}    & \multicolumn{2}{c}{Rainbow + SRNN}   & \multicolumn{2}{c}{Rainbow + IQ-RE (ours)}     \\ \hline
N                     & 10,000        & 1,000,000    & 10,000        & 1,000,000      & 10,000           & 1,000,000        & 10,000      & 1,000,000        & 10,000        & 1,000,000            & 10,000         & 1,000,000                  \\ \hline
\textit{Pong}         & $16.2$        & $15.0$       & \bm{$17.3$}   & $15.2$         & $16.7$           & \bm{$16.3$}      & $17.1$      & $19.1$           & \bm{$17.9$}   & \bm{$19.3$}          & $17.7$         & $19.0$                   \\ 
\textit{Breakout}     & $71.6$        & $80.9$       & $78.0$        & $86.4$         & \bm{$90.5$}      & \bm{$97.4$}      & $59.3$      & $62.1$           & $61.1$        & $66.2$               & \bm{$67.6$}    & \bm{$66.7$}              \\ 
\textit{Carnival}     & $4327.0$      & $4589.0$     & $4273.6$      & \bm{$4758.0$}  & \bm{$5073.1$}    & $4639.8$         & $4529.7$    & $4788.0$         & $3599.6$      & $4662.2$             & \bm{$4928.2$}  & \bm{$5289.0$}            \\ 
\textit{Freeway}      & \bm{$0.5$}    & $19.5$       & \bm{$0.5$}    & $13.2$         & \bm{$0.5$}       & \bm{$25.8$}      & $25.7$      & $33.2$           & \bm{$32.3$}   & \bm{$33.3$}          & \bm{$32.3$}    & \bm{$33.3$}              \\ 
\textit{Tennis}       & $-13.2$       & $-1.3$       & \bm{$-10.5$}  & $-1.0$         & $-10.7$          & \bm{$-0.4$}      & \bm{$0.0$}  & \bm{$0.0$}       & $-0.7$        & \bm{$0.0$}           & \bm{$0.0$}     & \bm{$0.0$}               \\ 
\textit{FishingDerby} & $-23.8$       & $9.7$        & $-24.0$       & \bm{$11.0$}    & \bm{$-23.5$}     & $7.5$            & $-2.7$      & $29.4$           & $-7.0$        & $27.0$               & \bm{$-2.4$}    & \bm{$30.8$}              \\ 
\specialrule{0em}{1pt}{1pt}
\bottomrule
\end{tabular}
\end{table*}

\subsection{Results}
To address the effectiveness of our proposed method, we present primary results of IQ incorporated with DQN and all baselines implemented on 4 control tasks. Fig. \ref{fig:Classic_Control} shows the learning curves of average episodic return during training for each task with three levels of replay buffer capacity, and Table \ref{table:highest_cumulative_score_classic_control} reports the numerical results in terms of the highest cumulative return achieved in corresponding curves. In general, IQ is clearly superior to all baselines both in terms of the stability and the maximum achieved cumulative reward, especially when the replay buffer capacity is small ({\em e.g.}, $N=100$) or even without experience replay ({\em i.e.}, $N=1$). In most tasks, IQ achieves near optimal performance as well as good stability even without any experience replay. For {\em Pendulum-v0} and {\em Acrobot-v1}, a large replay buffer ({\em e.g.}, $N=50,000$) can help DQN, SRNN and TCNN escape from catastrophic interference. However, this is not the case for two {\em CartPole} tasks where the agents exhibit fast initial learning but then encounter collapse in performance. DSOM performs comparably to IQ with large replay buffers, but is significantly inferior to IQ with the other two smaller capacities. We also conduct experimental comparisons with all baselines in terms of the degree of interference (according to Eq. \eqref{AEI}), corresponding to all task settings in Fig. \ref{fig:Classic_Control}. The experimental results are presented in Appendix \ref{Measuring Interference}, from which we can further confirm that IQ can substantially reduce the negative interference encountered by the base RL agents during the learning progress.

Moreover, from a macro perspective, Fig. \ref{fig:Classic_Control} shows that: 1) DQN, SRNN and TCNN agents exhibit high sensitivity to the replay buffer capacity. They generally perform well with a large buffer (except on {\em CartPole-v1}), but their performance deteriorates significantly when the buffer capacity is reduced. The reason for this phenomenon is that DQN primarily relies on experience replay to obtain approximately {\em i.i.d.} training data to avoid possible interference in training, which cannot be guaranteed when the replay buffer capacity is small. Since SRNN just reshapes the constraint term on the basis of DQN loss, while TCNN only increases the input dimension of DQN, both techniques can only alleviate interference to a certain extent. 2) The overall performance of DSOM is better than the above three baselines, as it optimizes the data distribution-specific representation modules to circumvent the interference caused by the shared representation layer. Nevertheless, since DSOM shares the same output layer, it still suffers from interference on most tasks. 3) By contrast, IQ features a shared representation module and multiple data distribution-specific output heads, and employs the knowledge distillation technique to prevent interference caused by shared representation layers, achieving significantly better performance than baselines. Note that in some cases ({\em e.g.}, {\em CartPole-v1} settings), IQ learns more slowly than baselines during the early stages of training. A possible explanation is that IQ learns context division in a fully online manner and the partitions may not be accurate enough back then, but it can quickly surpass the baselines as the training progress.

Additionally, to demonstrate the scalability and flexibility of our method, we also provide the results of IQ-RE with DQN and Rainbow separately, on 6 Atari games. The learning curves are shown in Fig. \ref{fig:Atari_games} and the highest cumulative returns achieved during training are summarized in Table \ref{table:highest_cumulative_score_atari}. Overall, for high-dimensional image inputs, the training performance of the underlying RL algorithms can be noticeably improved with our IQ-RE scheme, while SRNN provides little contribution to both underlying RL methods. Specifically, in Fig. \ref{fig:Atari_games}, with DQN as the underlying RL method, IQ-RE significantly outperforms DQN and SRNN on 7 out of 12 tasks, being comparable with DQN and SRNN on the rest 5 tasks. Similarly, with Rainbow as the underlying RL method, IQ-RE outperforms baselines on 8 out of 12 tasks, while being comparable with baselines on the rest 4 tasks. The two-sample Kolmogorov-Smirnov test in Appendix \ref{ks_test} further confirms that the improvement brought by IQ-RE is statistically significant. Furthermore, as shown in Table \ref{table:highest_cumulative_score_atari}, IQ-RE achieves higher maximum cumulative scores in most tasks than its counterparts. Among the 24 training settings, the maximum cumulative scores achieved by IQ-RE are slightly lower than those of baselines in only 4 cases and 2 cases when combined with DQN and Rainbow, respectively. It is worth noting that, even with a large memory ($N=1,000,000$), IQ-RE still shows certain advantages over the baselines.

In summary, the proposed techniques containing context division based on the clustering of all experienced states, and knowledge distillation in multi-head neural networks can effectively eliminate catastrophic interference caused by data drift in the single-task RL, whilst reduce the requirement of the replay buffer capacity for off-policy RL. In addition, our method leverages a fixed randomly initialized encoder to characterize the similarity among states in the low-dimensional representation space, which can be used to partition contexts effectively for high-dimensional environments. 

\subsection{Analysis}
\label{Analysis}

\begin{figure*}[t]
    \centering
    \setlength{\abovecaptionskip}{5pt}
        \begin{minipage}[b]{0.98\textwidth}
			\includegraphics[width=0.24\textwidth]{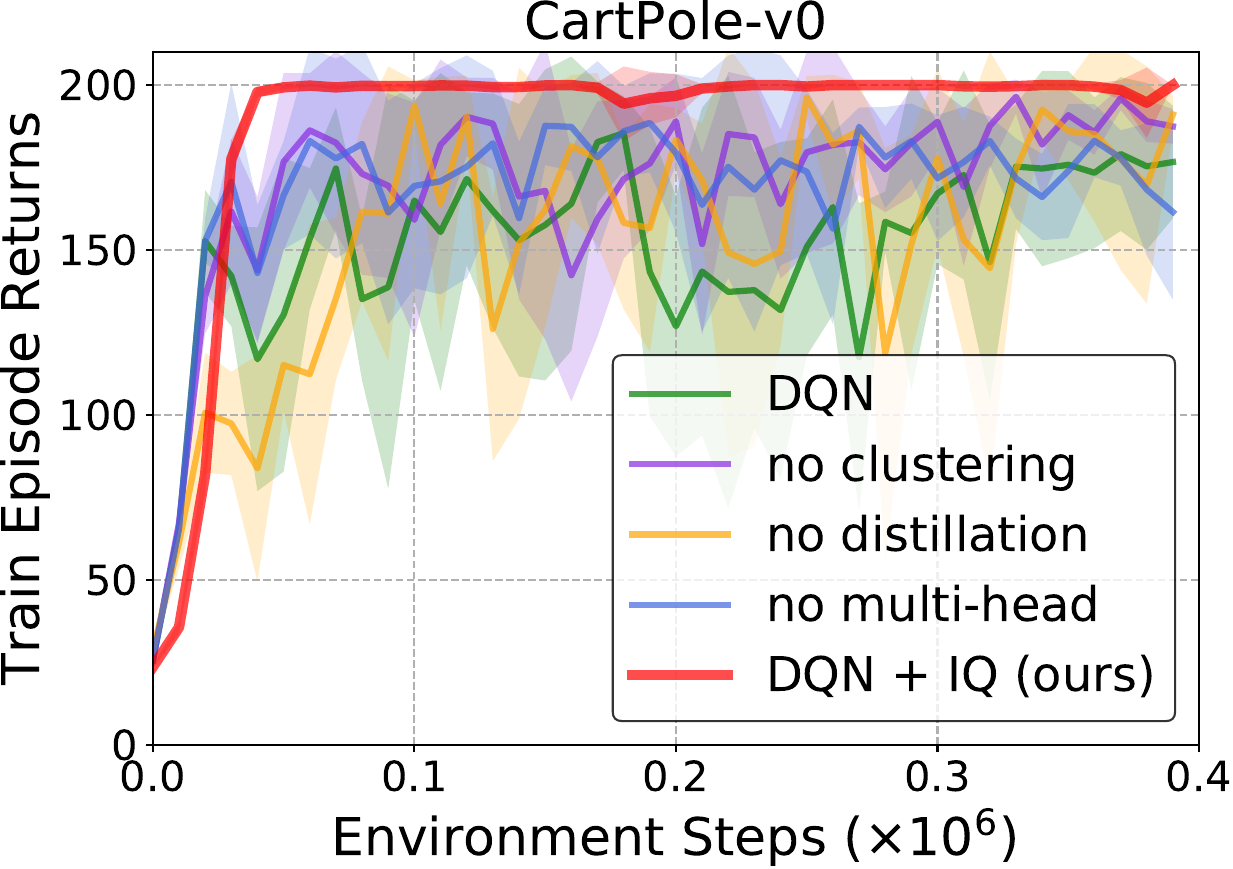}
			\includegraphics[width=0.252\textwidth]{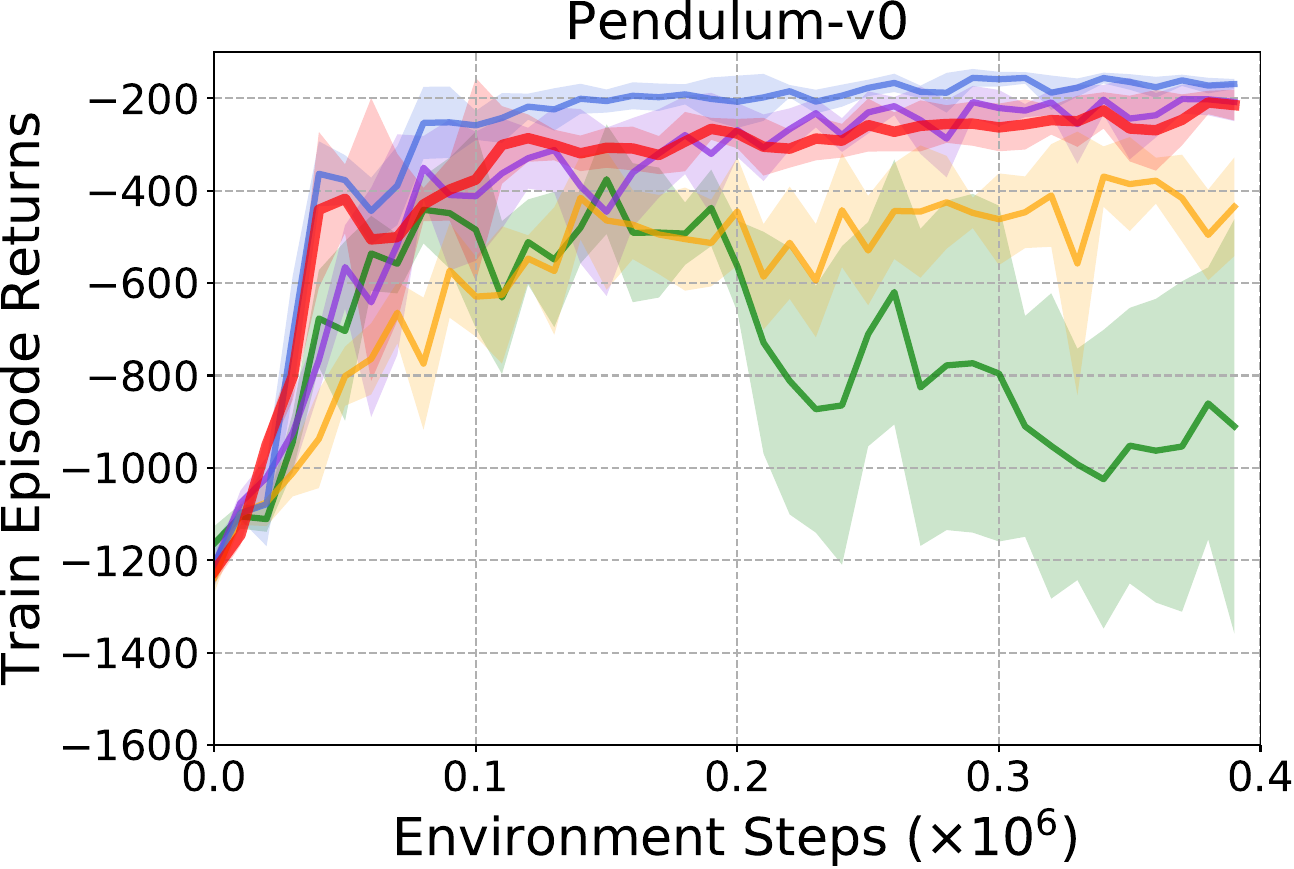}
			\includegraphics[width=0.24\textwidth]{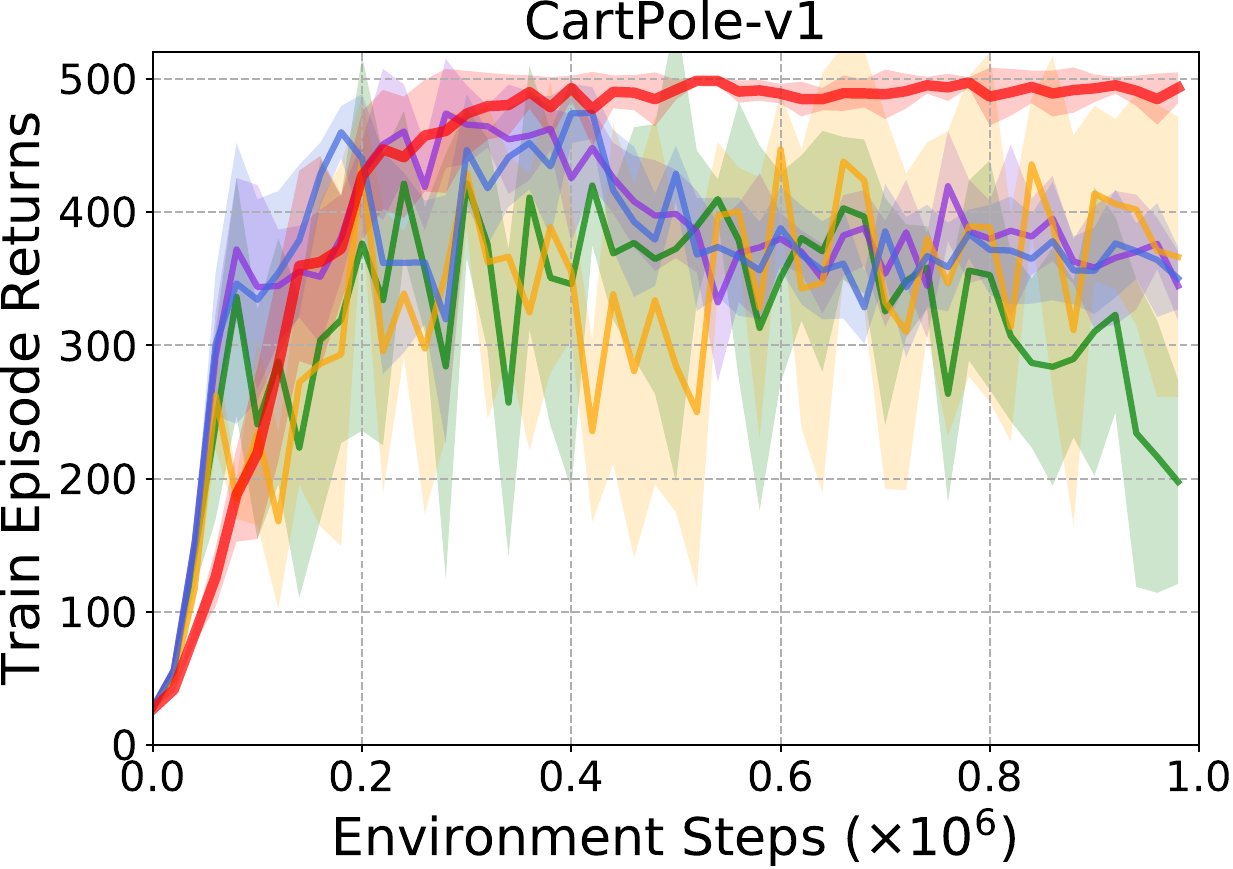}
			\includegraphics[width=0.247\textwidth]{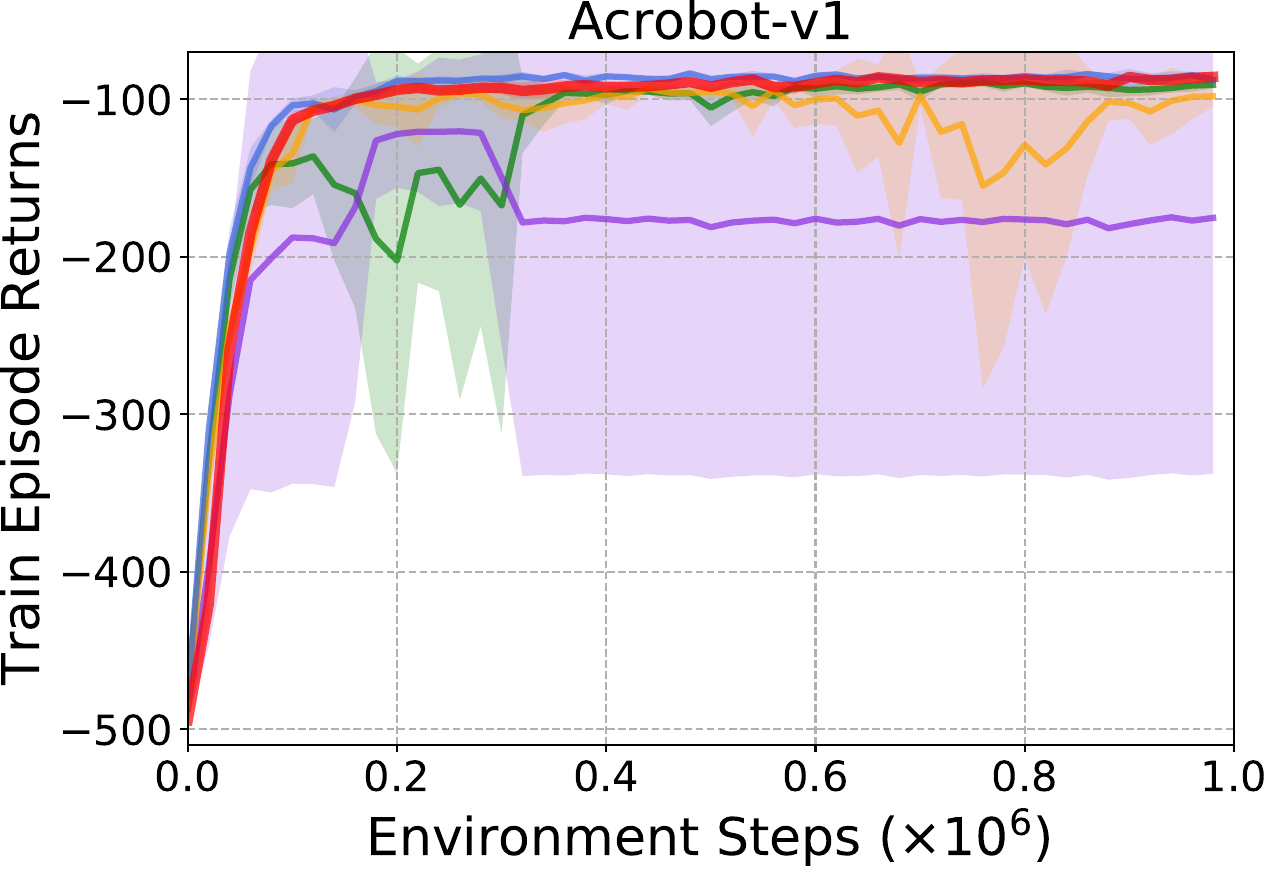}
		\end{minipage}
    \caption{Comparisons of IQ (red) with DQN (green) and its three different ablations (other colors), on each individual task ($N=100$).}
    \label{fig:ablation_study}
\end{figure*}

\begin{figure*}[t]
    \centering
    \setlength{\abovecaptionskip}{5pt}
    \begin{minipage}[b]{0.98\textwidth}
		\includegraphics[width=0.24\textwidth]{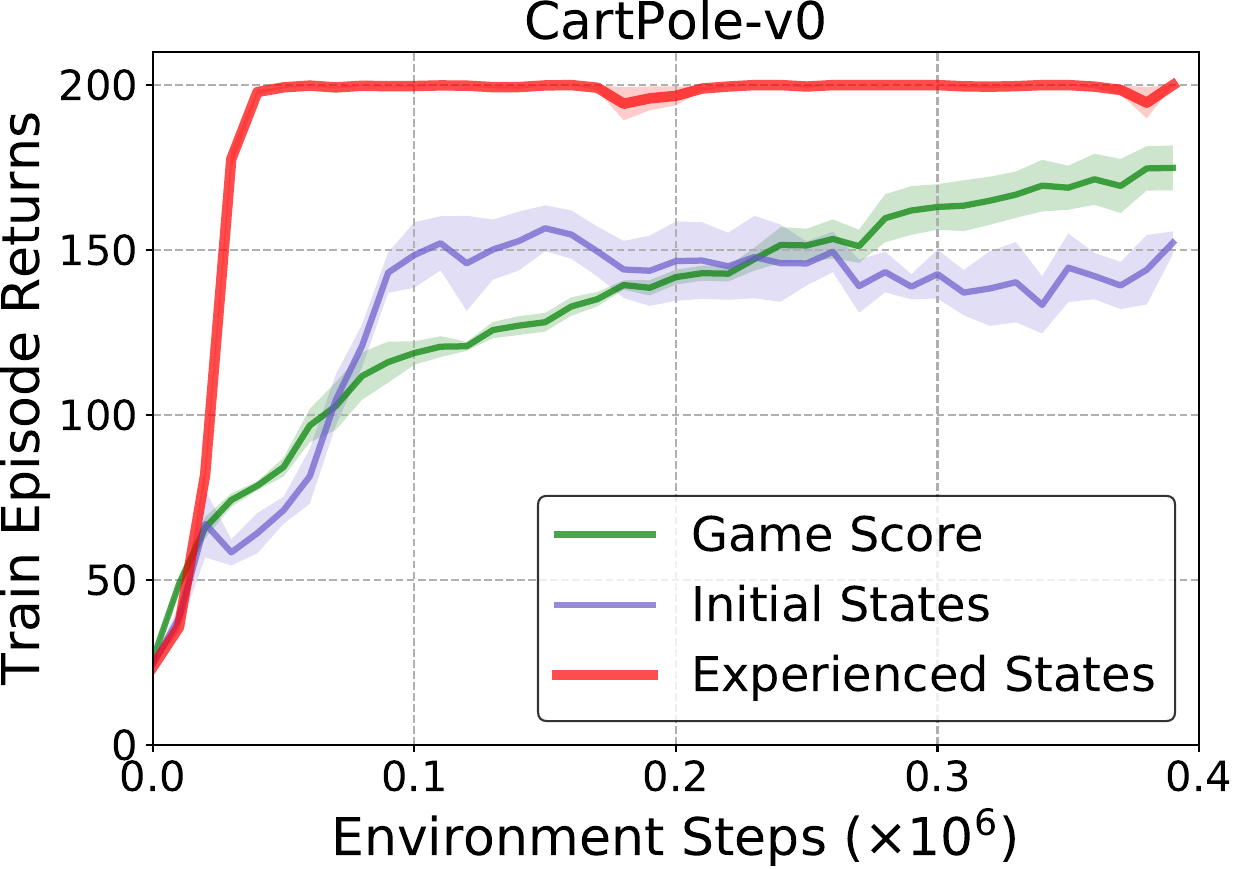}
		\includegraphics[width=0.252\textwidth]{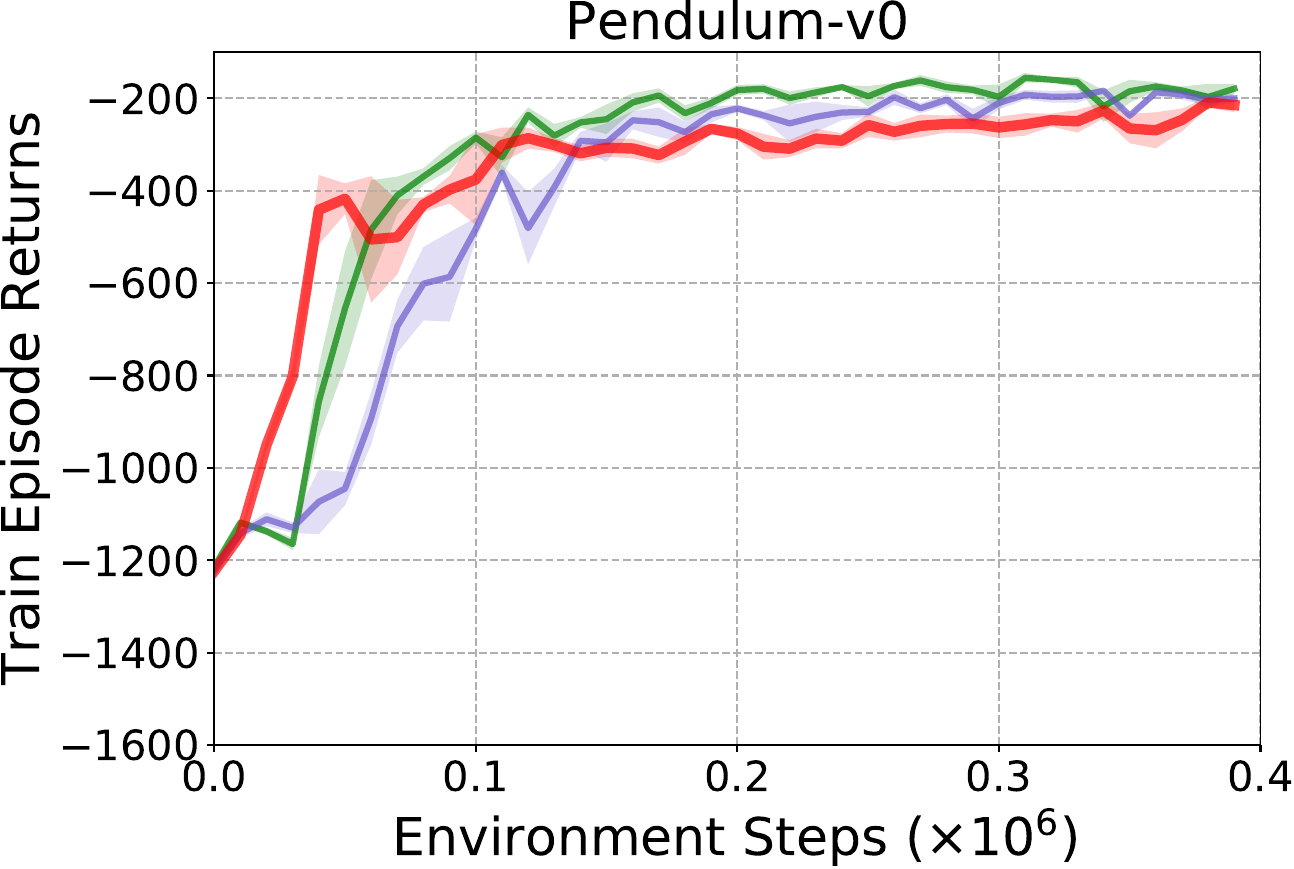}
		\includegraphics[width=0.24\textwidth]{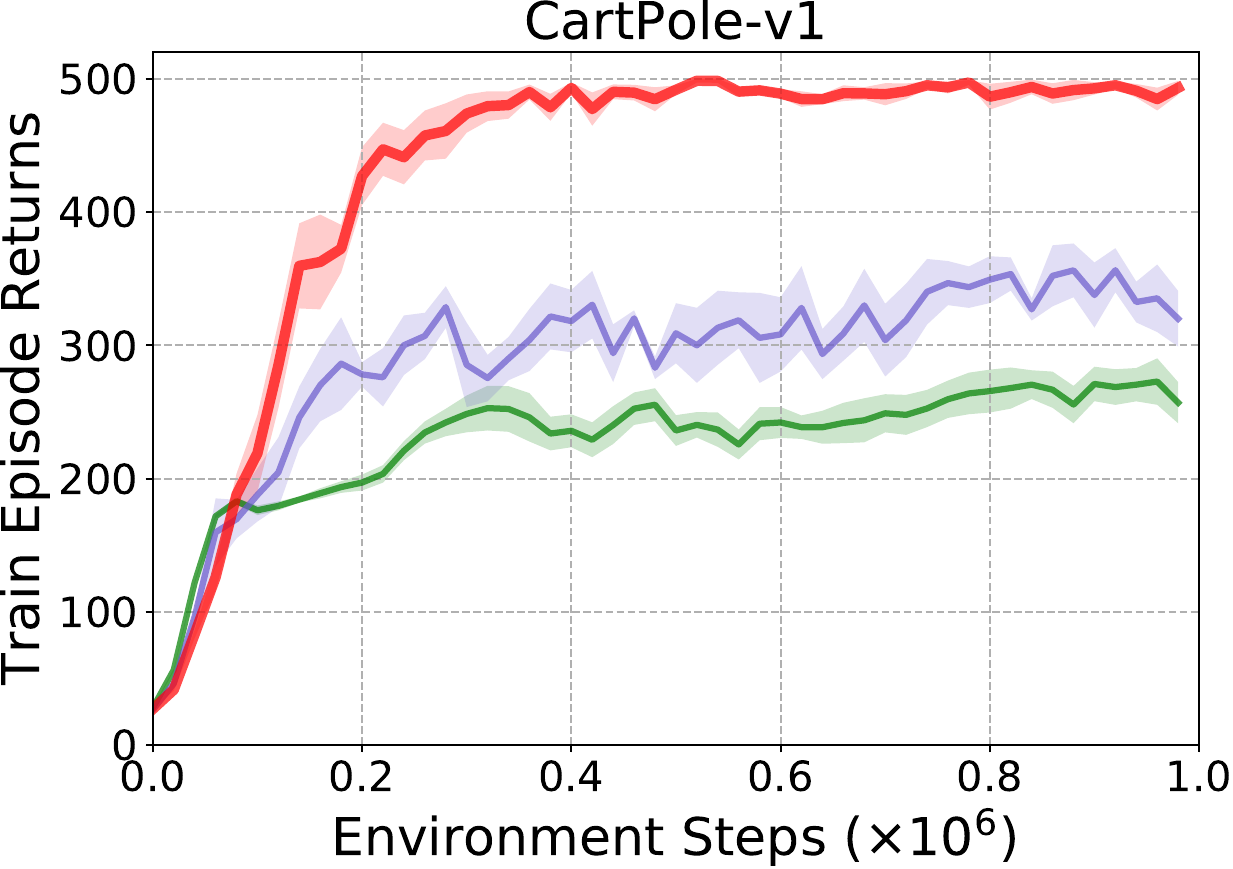}
		\includegraphics[width=0.247\textwidth]{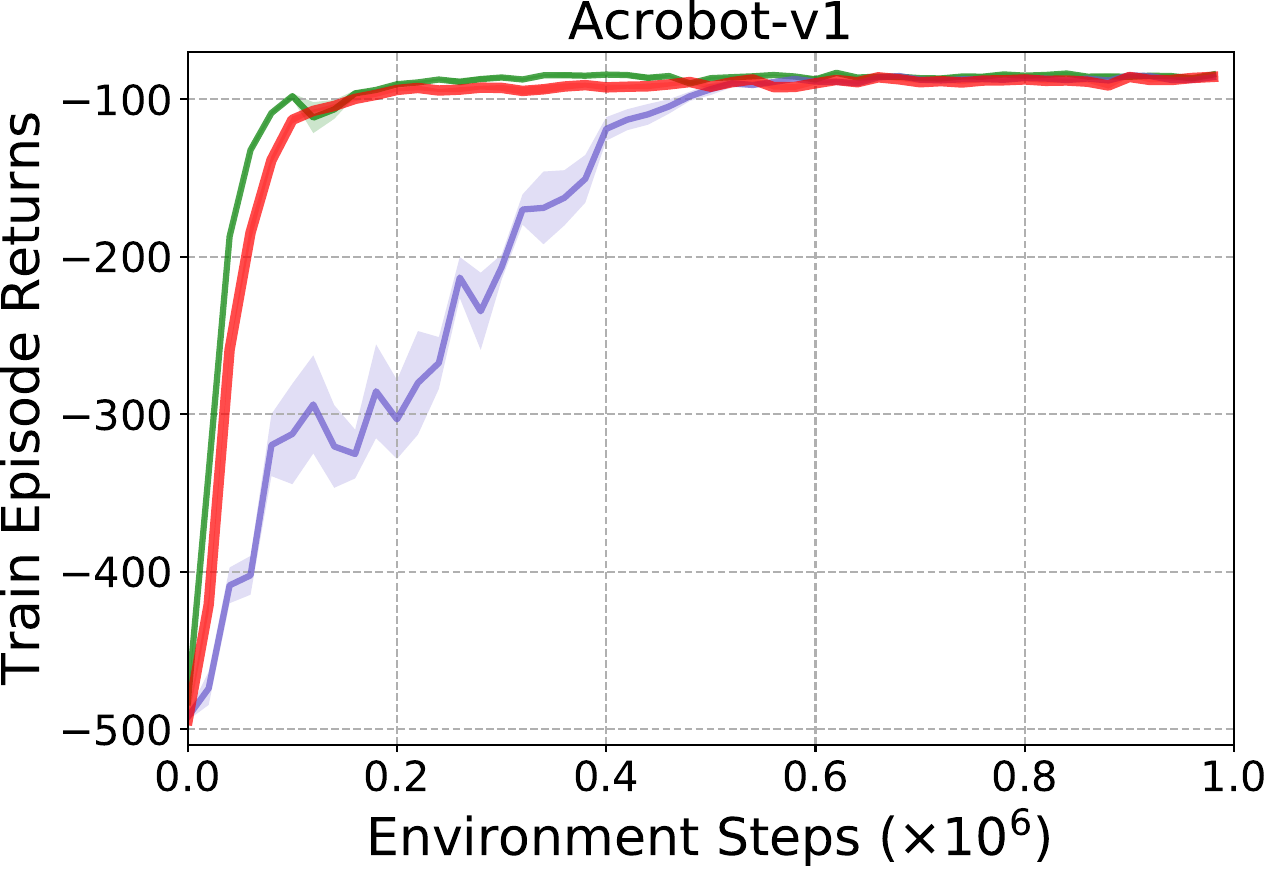}
	\end{minipage}
    \caption{Comparisons of IQ (red, performs context division on all experienced states) with other two context division strategies (other colors) ($N=100$).}
    \label{fig:cd_strategy}
\end{figure*}

{\em 1) Ablation Study}: Since our method can be regarded as an extension to existing RL methods ({\em e.g.}, DQN \cite{mnih2015human}), with three novel components ({\em i.e.}, adaptive context division by online clustering, knowledge distillation, and the multi-head neural network), the ablation experiments are designed as follows:
\begin{itemize}
\item {\bf No clustering} means using a random partition of the raw state space before learning instead of adaptive context division by online clustering.
\item {\bf No distillation} means removing the distillation loss function $\mathcal{L}_\mathcal{D}(\theta_\mathcal{S},\theta_\mathcal{F})$ from Eq. \eqref{eq:joint_optimization_loss} ({\em i.e.}, $\lambda=0$).
\item {\bf No multi-head} means removing the context division module and optimize the neural network with a single-head output ({\em i.e.}, $k=1$). Here, the distillation term is represented as the distillation of the network before each update of the output head.
\end{itemize}

The results of ablation experiments are shown in Fig. \ref{fig:ablation_study}, using classic control tasks for the convenience of validation. From Fig. \ref{fig:ablation_study}, the following facts can be observed: 1) Across all settings, the overall performance of DQN is the worst, showing the effectiveness of the three components introduced for coping with catastrophic interference in the single-task RL, although the contribution of each component varies substantially per task; 2) Removing online clustering from the context division module is likely to damage the performance in most cases;  3) Removing knowledge distillation makes the performance deteriorate on almost all tasks, indicating that knowledge distillation is a key element in our method; 4) Without the multi-head component, our model is equivalent to a DQN with an extra distillation loss, which performs better than DQN alone but worse than our proposed IQ in general. Note that, on {\em Pendulum-v0}, the single-head network performs better than our method during training, which means that the additional distillation constraint is sufficient to mitigate the interference faced by the base RL under this setting, without the need of further context division. By contrast, our method learns context division in a fully online manner and the partitions may not be accurate enough on this task. However, this is not the case in other settings (See Appendix \ref{Additional results of ablation experiments} for more experimental results).

{\em 2) Context Division Strategy}: By introducing the context variables during learning, IQ bears a resemblance to some existing settings \cite{ghosh2018divide, jain2020algorithmic} that partition contexts using game score and initial states. Therefore, we further compare our context division based on all experienced states with the following context division strategies:
\begin{itemize}
\item {\bf Game Score} \cite{jain2020algorithmic} splits a game into contexts based on the undiscounted cumulative game score.
\item {\bf Initial States} \cite{ghosh2018divide} partitions the initial state space into ``slices" by the k-means clustering procedure.
\end{itemize}

From the experimental results in Fig. \ref{fig:cd_strategy}, we can observe that our method is distinctly superior to the above two strategies in general, although it performs slightly worse on {\em Pendulum-v0}. This is due to the fact that, on {\em Pendulum-v0}, the game scores or initial states have a perfect correspondence to different state distributions. However, these two baselines are primarily designed to decompose complex tasks, rather than mitigating catastrophic interference in the single-task RL settings and cannot guarantee the complete decoupling among different state distributions. This conclusion is further confirmed by the additional experimental results in Appendix \ref{Additional Results of Context Division Strategy}.

{\em 3) Parameter Analysis}: There are two critical parameters in IQ: $\lambda$ and $k$. By its nature, $\lambda$ is related to the training progress. Since we need to preserve the learned good policies during training, it is intuitive to gradually increase $\lambda$ until its value reaches 1. The reason is that, in early-stage training, the model has not learned any sufficiently useful information, so the distillation constraint can be ignored. With the progress of training, the model starts to acquire more and more valuable information and needs to pay serious attention to interference to protect the learned good policies while learning further. In our experiments, we recommend to set $\lambda$ to be inversely proportional to the exploration proportion $\epsilon$, and the results in Figs. \ref{fig:Classic_Control} and \ref{fig:Atari_games} have demonstrated the simplicity and effectiveness of this setting. 

\begin{figure}[!t]
    \centering
    \setlength{\abovecaptionskip}{5pt}
    \begin{minipage}[b]{0.475\textwidth}
    \flushright
		\includegraphics[width=0.475\textwidth]{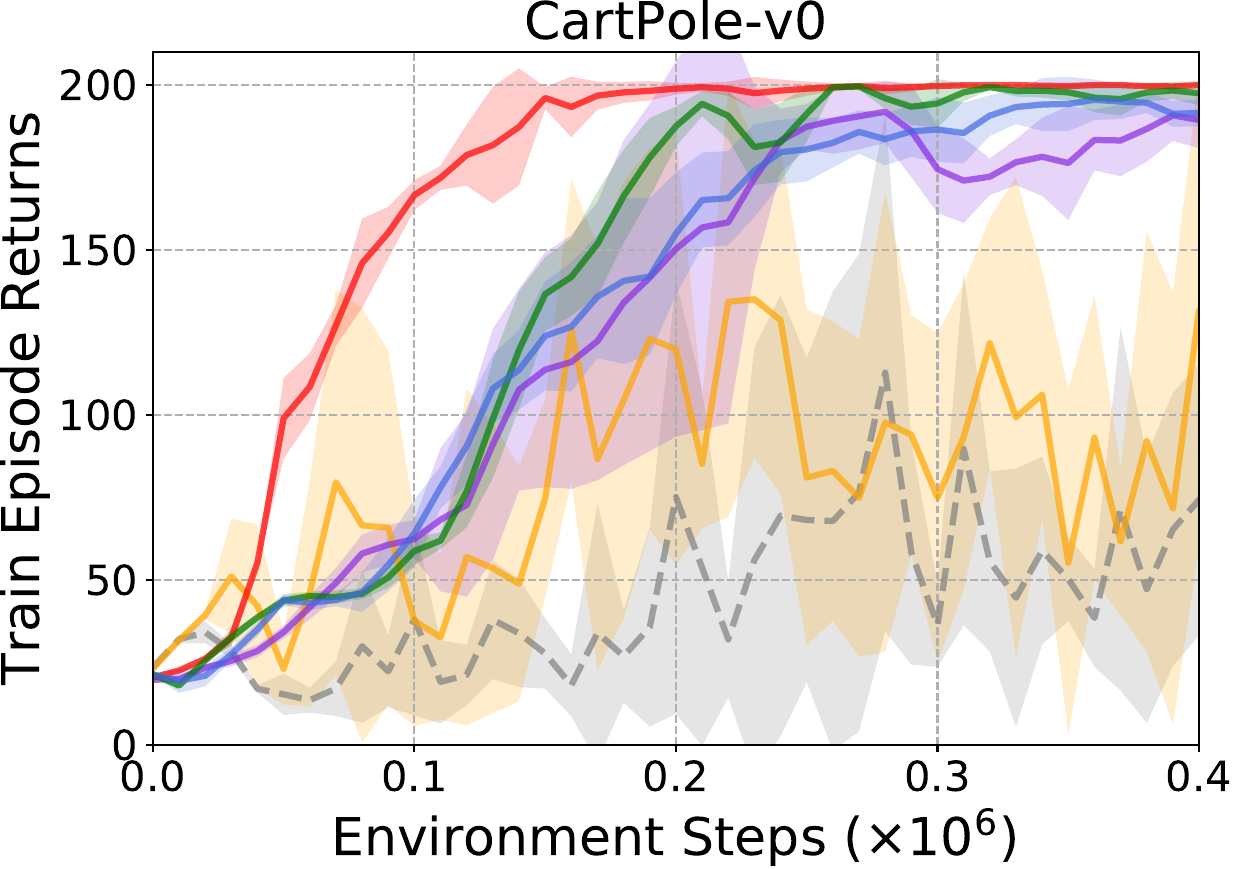}
		\includegraphics[width=0.502\textwidth]{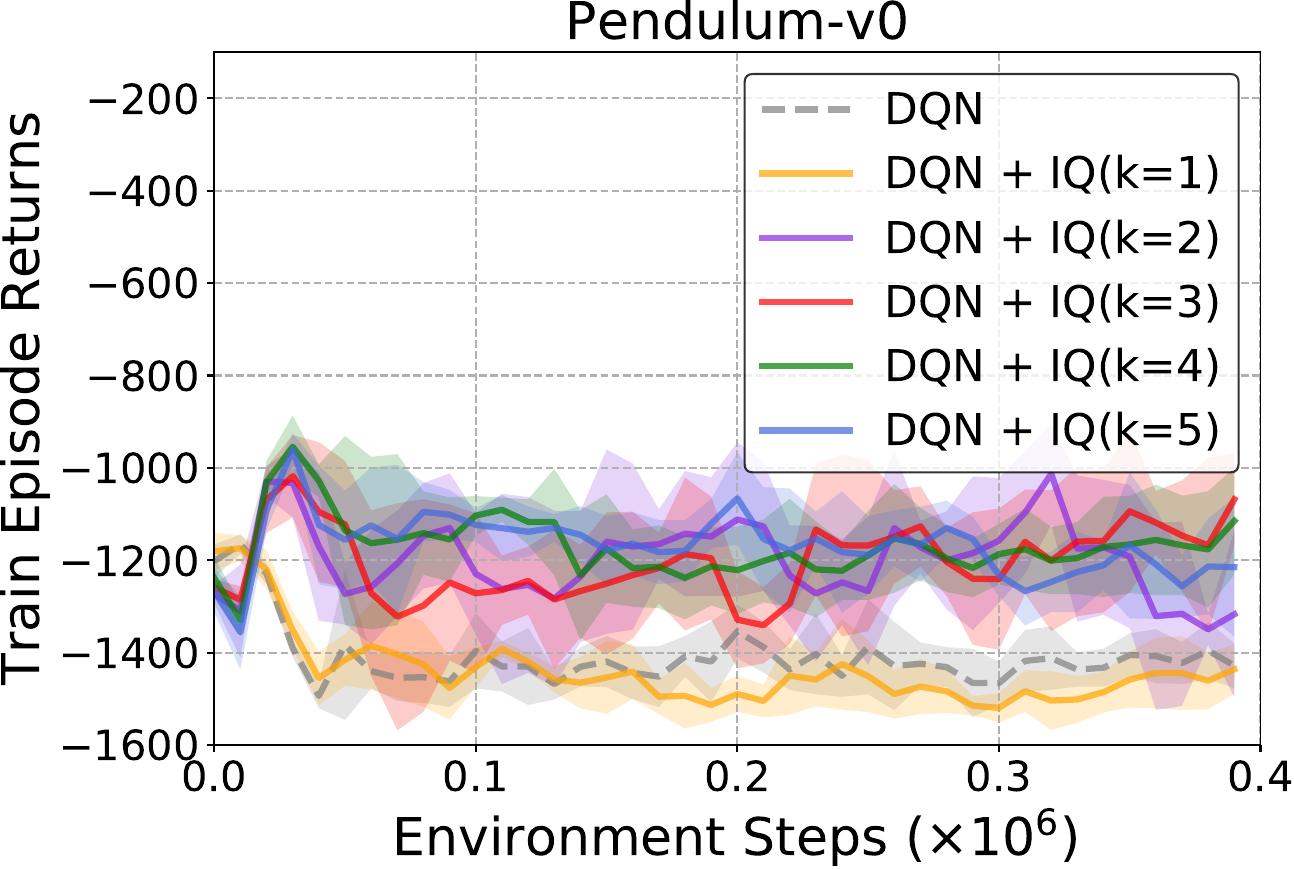}\\
		\vspace{0.2cm}
		\includegraphics[width=0.48\textwidth]{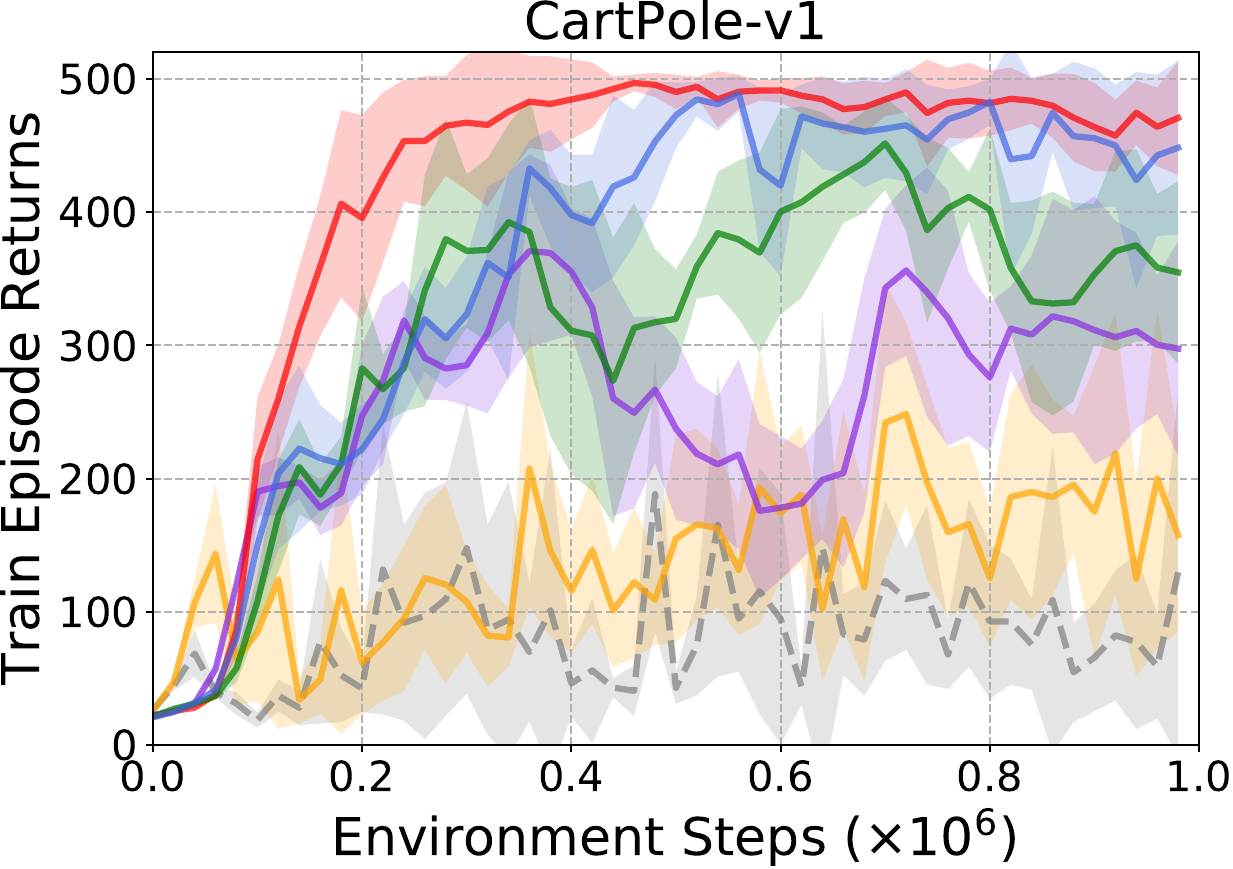}
		\includegraphics[width=0.497\textwidth]{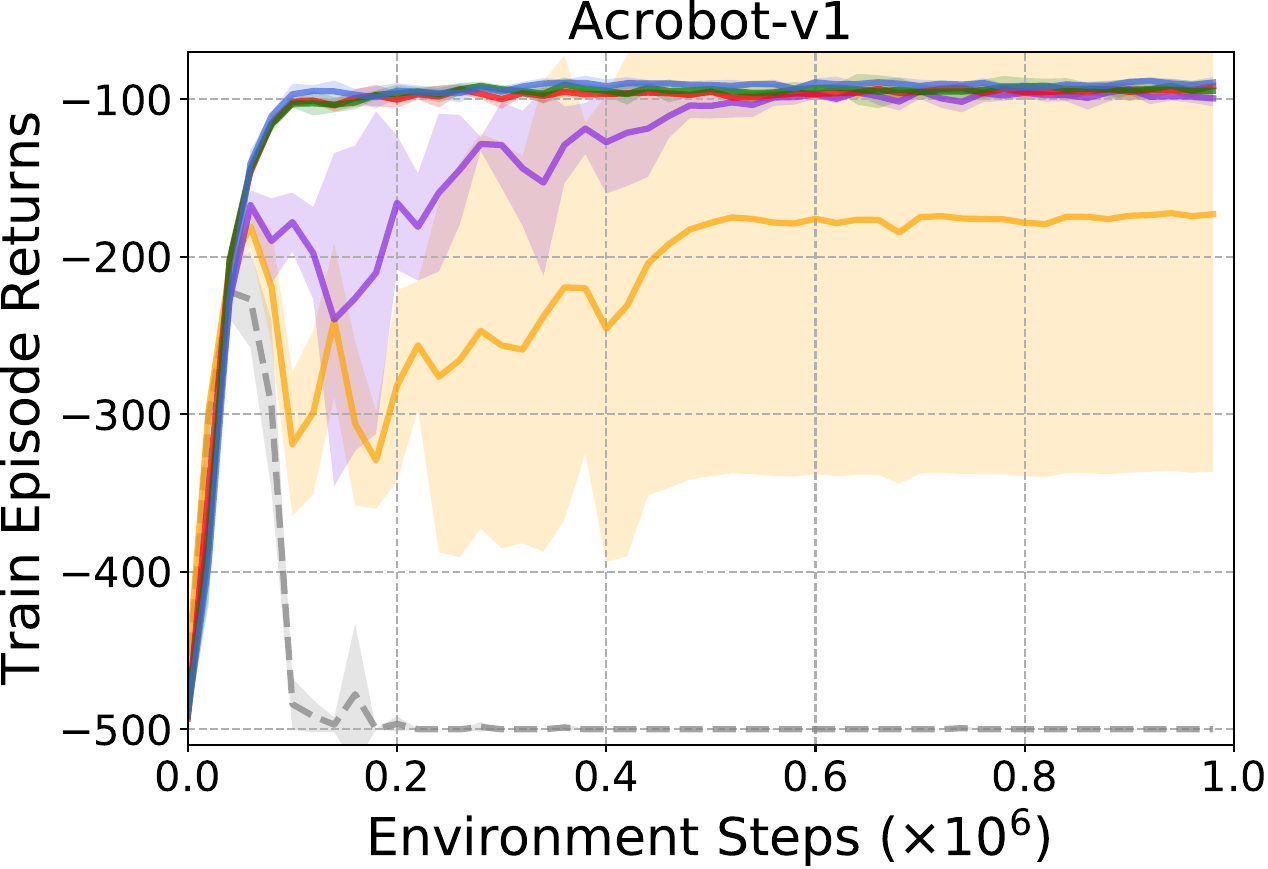}
	\end{minipage}
    \caption{Parameter sensitivity analysis {\em w.r.t} the number of contexts $k$. Experiments are conducted with different $k$ values ($N=1$).}
    \label{fig:parameter_study}
\end{figure}

\begin{figure*}[t]
    \centering
    \setlength{\abovecaptionskip}{5pt}
    \subfigure[\label{fig:average_loss_value}]
        {\includegraphics[width=0.3\linewidth]{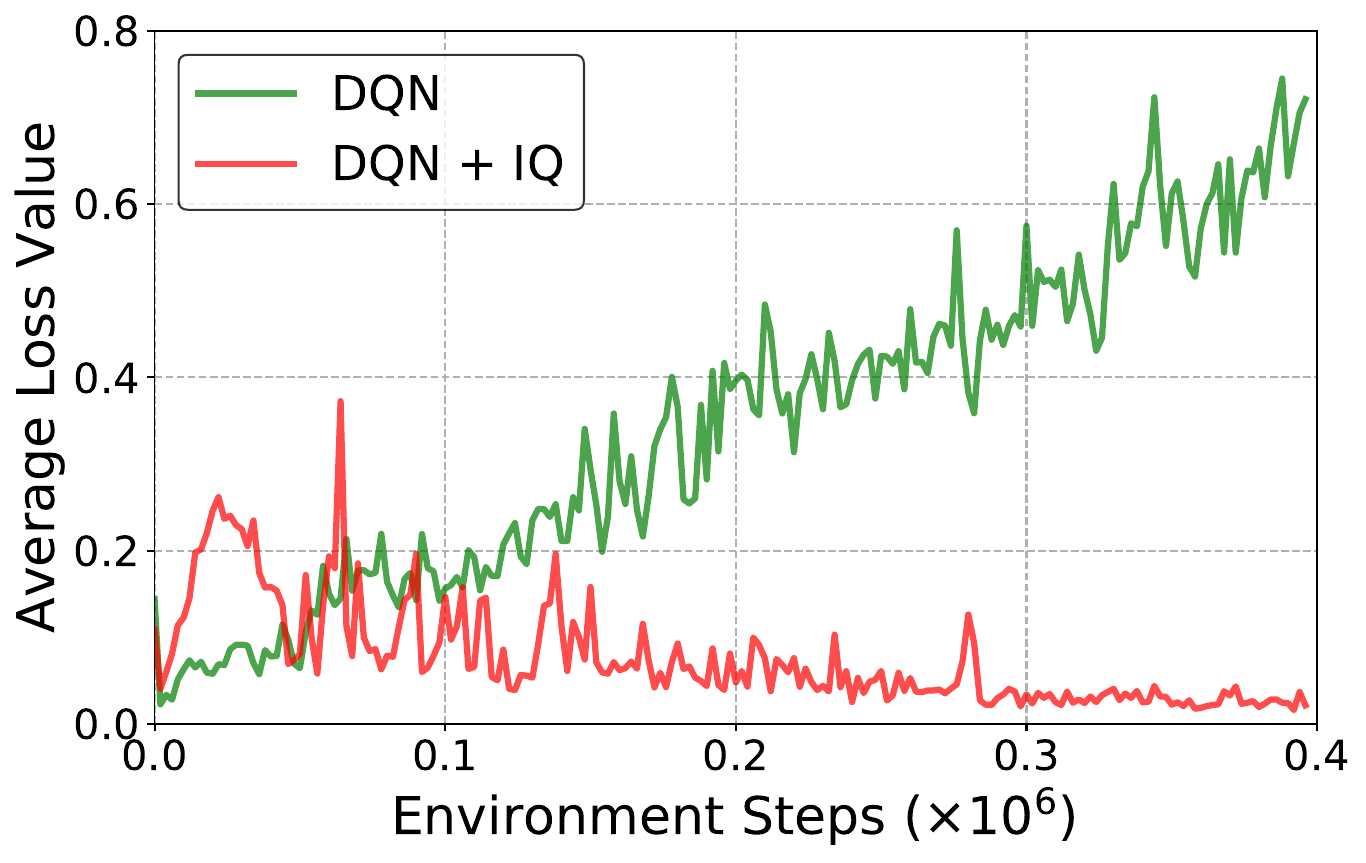}}
    \hspace{4mm}
    \subfigure[\label{fig:average_Q_methods}]
        {\includegraphics[width=0.3\linewidth]{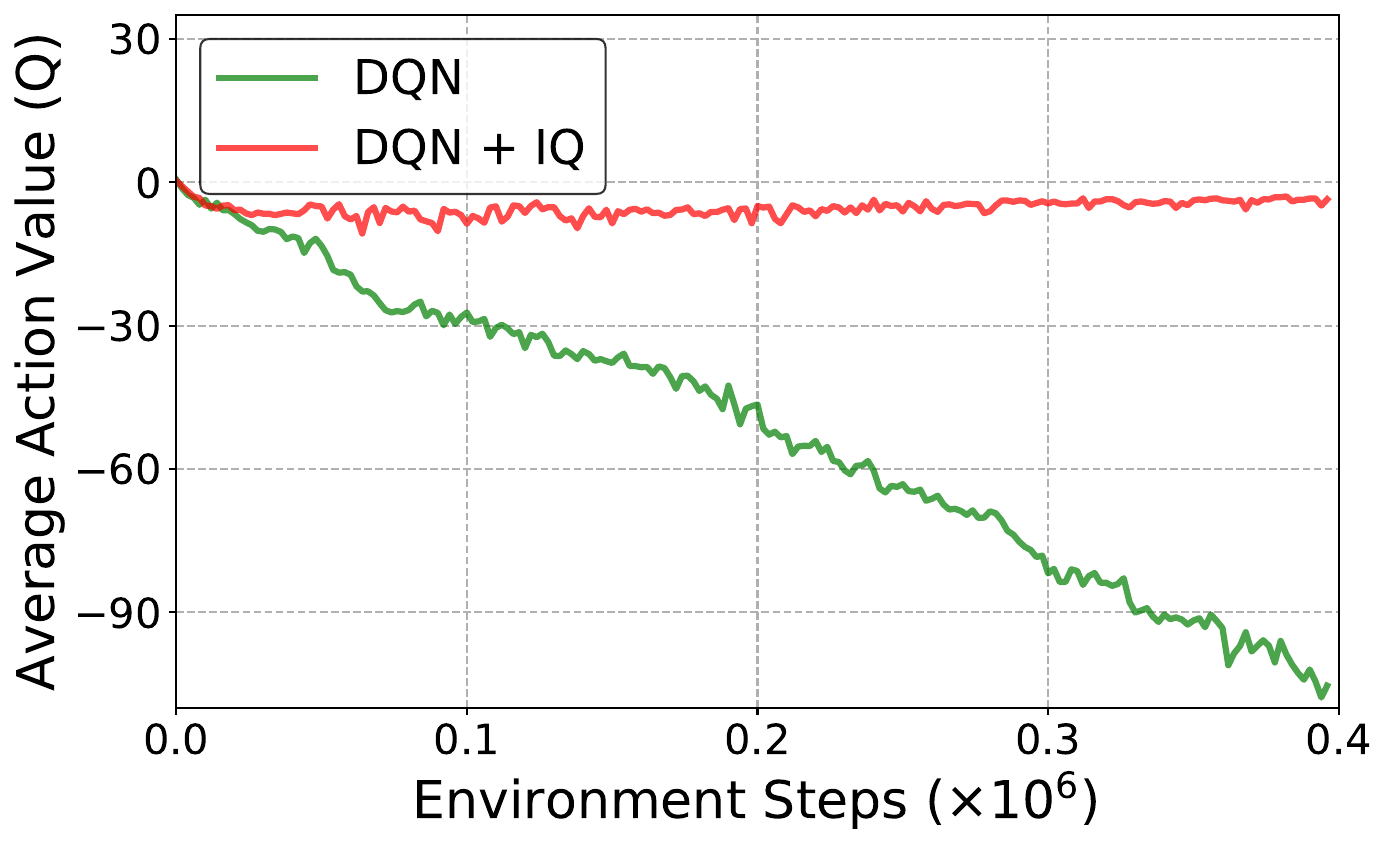}}
    \hspace{4mm}
    \subfigure[\label{fig:average_Q_heads}]
        {\includegraphics[width=0.3\linewidth]{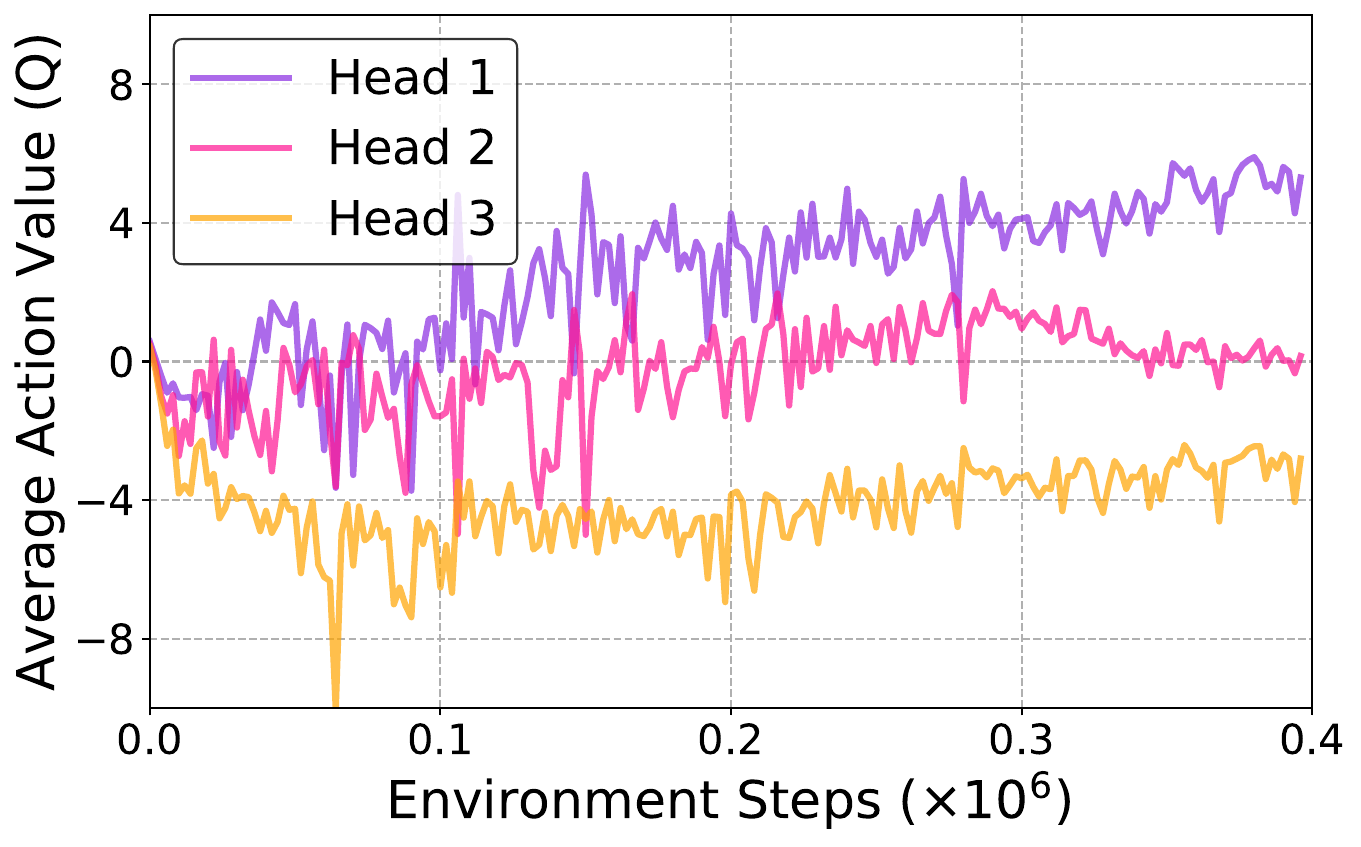}}
\caption{Training curves tracking the agent's average loss and average predicted action-value for 400K environment steps in {\em Pendulum-v0} ($N=100$ and $k=3$, see Fig. \ref{fig:Classic_Control_b} for corresponding curves). (a) Each point is the average loss achieved per training iteration; (b) Average maximum predicted action-value of agents on a held-out set of states; (c) Average maximum predicted action-value of each output head in IQ.}
\label{fig:convergence_analysis}
\end{figure*}

For the parameter $k$, it needs to be specified before training, which may be suboptimal without a good knowledge of the state space structure of the environment. To investigate the effect of $k$, we conduct experiments with different $k$ values ($k\in\{1,2,3,4,5\}$) and the results are shown in Fig. \ref{fig:parameter_study}. In our experiments, $k=3$ is a reasonably good choice for {\em CartPole-v0} and {\em CartPole-v1}, while $k=5$ is best for IQ on {\em Acrobot-v1}. It is worth noting that, on {\em Pendulum-v0}, our method achieves similar performance with $k$ set to 2, 3, 4, and 5, respectively, but without any satisfactory result. A possible explanation is that the agents failed to learn any useful information due to the limited exploration in the early training, leading to the failure of further learning. 

In summary, we can make the following statements: 1) The performance of our method is obviously better than the base RL baseline regardless of the specific $k$ value, confirming the effectiveness of IQ even with inaccurate $k$ estimation; 2) For $k>1$, better performance of IQ can be expected. However, large $k$ values are not always desirable as it will result in more fine-grained context divisions and more complex neural networks with a large amount of output heads, making the model unlikely to converge satisfactorily within a limited number of training steps. Thus, we recommend to set the value of $k$ by taking into consideration the state-space structure of specific tasks. In practice, we recommend to explore the environment using an appropriate random policy and conduct initial density-based clustering for the obtained states before training. Thereafter, the initial centroids of {\em SKM} and $k$ value can be estimated according to this initial clustering result.

{\em 4) Convergence Analysis}: To analyze convergence, we track the agent's average loss and average predicted action-value during training progress. According to Fig. \ref{fig:convergence_analysis}, we can conclude that: 1) Our method has better convergence and stability in face of interference compared with original RL algorithms (See Figs. \ref{fig:average_loss_value} and \ref{fig:average_Q_methods}); 2) For a held-out set of states\footnote{It refers to a fixed set of states \cite{mnih2013playing, mnih2015human} of the environment, which can be obtained by performing exploration in the environment and then used to track the average predicted action value changes during training.}, the average maximum predicted action-value of each output head reflects the difference as expected (See Fig. \ref{fig:average_Q_heads}), and the final output of IQ is synthesised based on all of them.

\begin{figure}[!t]
    \centering
    \setlength{\abovecaptionskip}{4pt}
    \subfigure[Training time \label{fig:training_time}]
        {\includegraphics[width=0.425\linewidth]{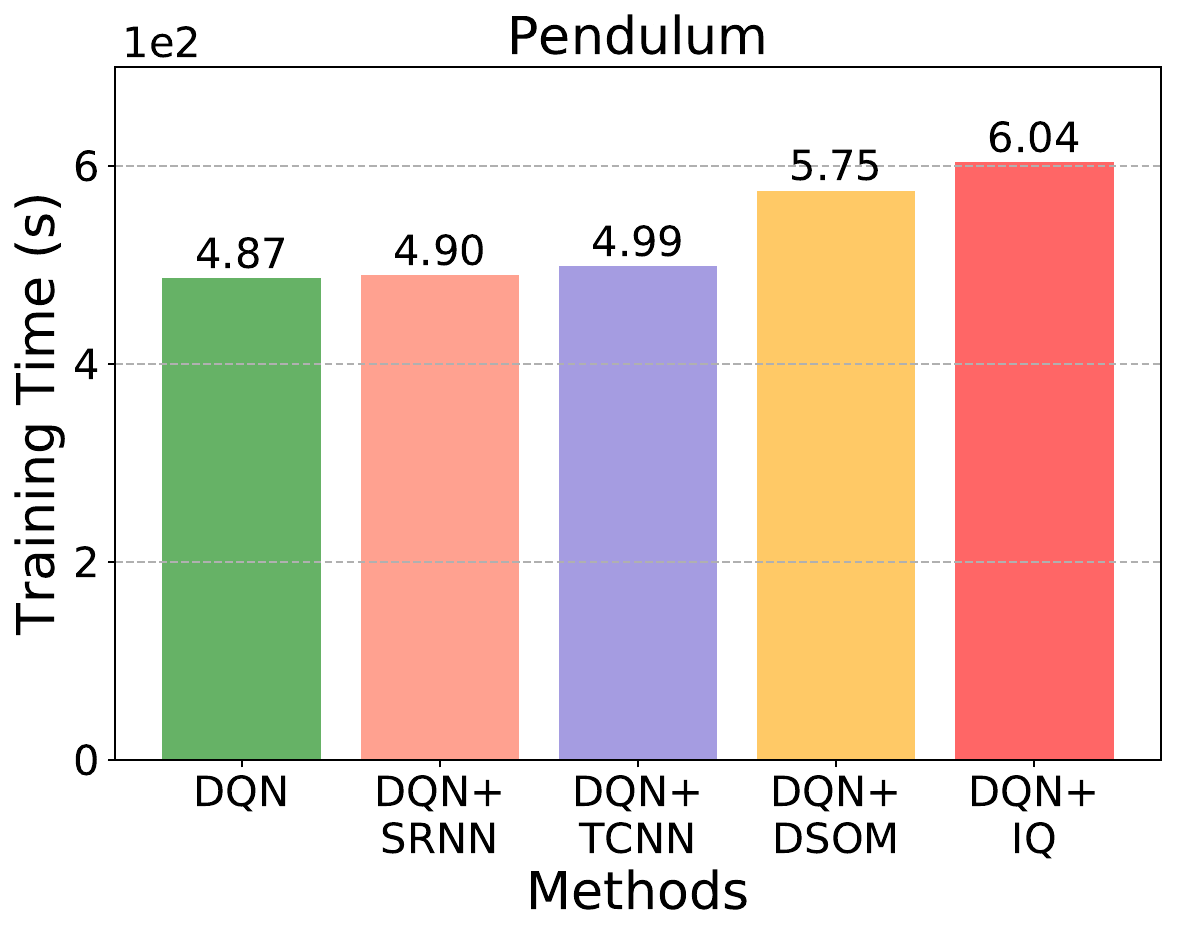}}
    \subfigure[Computational complexity \label{fig:FLOPs}]
        {\includegraphics[width=0.555\linewidth]{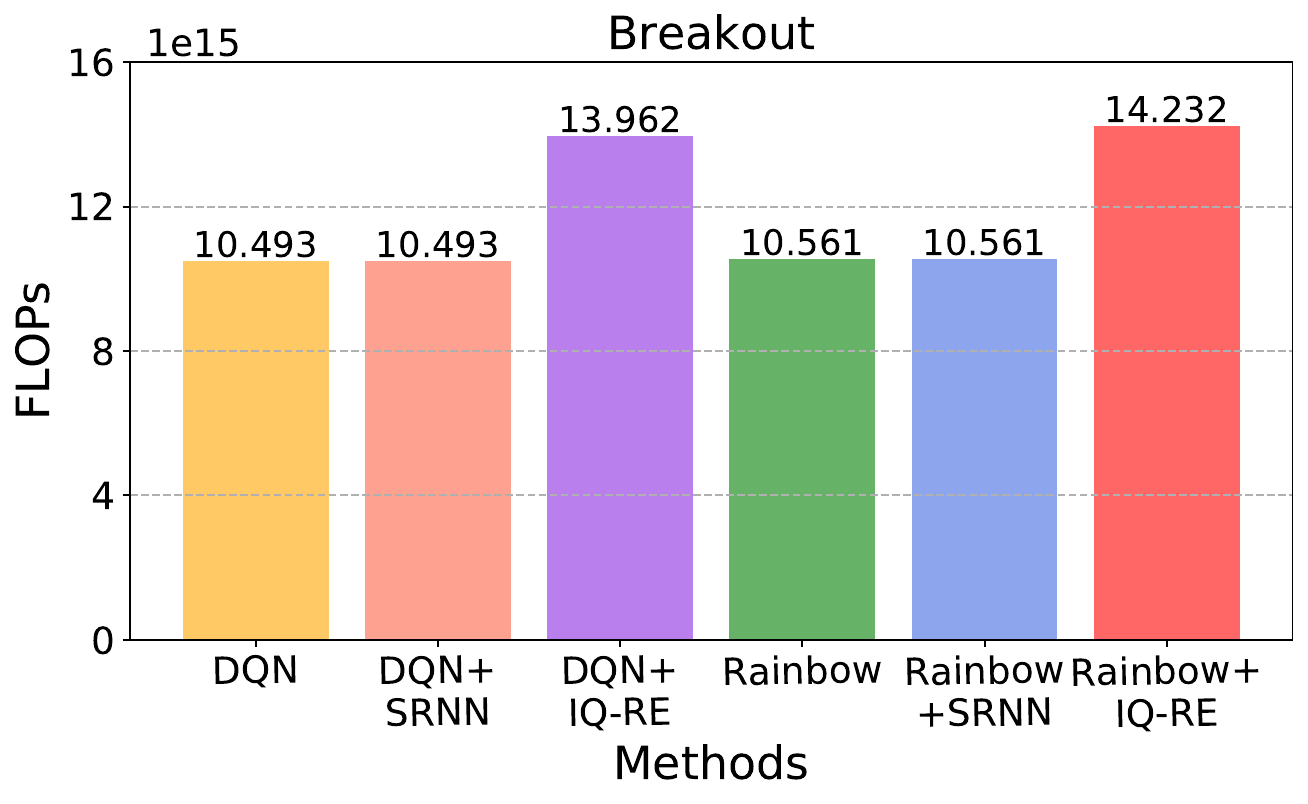}}
\caption{Comparison of computational efficiency: (a) Training time of each agent to achieve its performance for 400K environment steps in {\em Pendulum-v0} ($N=100$, see Fig. \ref{fig:Classic_Control_b} for corresponding learning curves); (b) Number of FLOPs used by each agent at 10M environment steps in {\em Breakout}. Here, we only take into account forward and backward passes through neural network layers (See Fig. \ref{fig:Atari_games} for corresponding learning curves).}
\label{fig:computational_efficiency}
\end{figure}
{\em 5) Computational Efficiency}: Our methods greatly improve training performance of the existing RL algorithms, which are computationally efficient in that: 1) In each time step, the extra context division module only needs to compute the distances between the current state and $k$ context centroids, which is computationally negligible {\em w.r.t.} the SGD complexity of the large parameter vector updated in each iteration of RL itself; 2) Only $k-1$ extra output heads are added to the neural network, in which the increased computation is acceptable {\em w.r.t.} the representation complexity; 3) There are no gradient updates through the random encoder; 4) There is no unnecessary distance computation for finding the corresponding context at every update step as the context label for each state is stored in the replay buffer. Fig. \ref{fig:computational_efficiency} shows the training time of each agent on {\em Pendulum} and the floating point operations (FLOPs) executed by agents on {\em Breakout}, respectively.

\section{Conclusion and Future Work}
In this paper, we propose a competent scheme IQ to tackle the inherent challenge of catastrophic interference in the single-task RL. The core idea is to partition all states experienced during training into a set of contexts using online clustering techniques and simultaneously estimate the context-specific value function with a multi-head neural network as well as a knowledge distillation loss to mitigate the interference across contexts. Furthermore, we introduced a random encoder to enhance the context division for high-dimensional complex tasks. Our method can effectively decouple the correlations among differently distributed states and can be easily incorporated into various value-based RL models. Experiments on several benchmarks show that our method can significantly outperform state-of-the-art RL methods and dramatically reduce the memory requirement of existing RL methods.

In the future, we aim to incorporate our method into policy-based RL models to reduce the interference during training by applying weight or functional regularization on policies. Furthermore, we will investigate a more challenging setting called continual RL in non-stationary environments \cite{lomonaco2020continual}. This setting is a more realistic representation of the real-world scenarios and includes abrupt changes or smooth transitions on dynamics, or even the dynamics itself is shuffled.

\section*{Acknowledgment}
The work presented in this paper was supported by the
National Natural Science Foundation of China (U1713214).


\bibliographystyle{IEEEtran}
\bibliography{IEEEexample}

\begin{thebibliography}{10}
\providecommand{\url}[1]{#1}
\csname url@samestyle\endcsname
\providecommand{\newblock}{\relax}
\providecommand{\bibinfo}[2]{#2}
\providecommand{\BIBentrySTDinterwordspacing}{\spaceskip=0pt\relax}
\providecommand{\BIBentryALTinterwordstretchfactor}{4}
\providecommand{\BIBentryALTinterwordspacing}{\spaceskip=\fontdimen2\font plus
\BIBentryALTinterwordstretchfactor\fontdimen3\font minus
  \fontdimen4\font\relax}
\providecommand{\BIBforeignlanguage}[2]{{%
\expandafter\ifx\csname l@#1\endcsname\relax
\typeout{** WARNING: IEEEtran.bst: No hyphenation pattern has been}%
\typeout{** loaded for the language `#1'. Using the pattern for}%
\typeout{** the default language instead.}%
\else
\language=\csname l@#1\endcsname
\fi
#2}}
\providecommand{\BIBdecl}{\relax}
\BIBdecl

\bibitem{riemer2019learning}
M.~Riemer, I.~Cases, R.~Ajemian, M.~Liu, I.~Rish \emph{et~al.}, ``Learning to
  learn without forgetting by maximizing transfer and minimizing
  interference,'' in \emph{Proceedings of the 7$^{th}$ International Conference
  on Learning Representations}, 2019.

\bibitem{sutton2018reinforcement}
R.~S. Sutton and A.~G. Barto, \emph{Reinforcement learning: An
  introduction}.\hskip 1em plus 0.5em minus 0.4em\relax MIT press, 2018.

\bibitem{li2019deep}
H.~Li, Q.~Zhang, and D.~Zhao, ``Deep reinforcement learning-based automatic
  exploration for navigation in unknown environment,'' \emph{IEEE Transactions
  on Neural Networks and Learning Systems}, vol.~31, no.~6, pp. 2064--2076,
  2019.

\bibitem{mnih2013playing}
V.~Mnih, K.~Kavukcuoglu, D.~Silver, A.~Graves, I.~Antonoglou, D.~Wierstra, and
  M.~Riedmiller, ``Playing atari with deep reinforcement learning,'' in
  \emph{Proceedings of the 27$^{th}$ Conference on Neural Information
  Processing Systems}, 2013.

\bibitem{mnih2015human}
V.~Mnih, K.~Kavukcuoglu, D.~Silver, A.~A. Rusu, J.~Veness, M.~G. Bellemare,
  A.~Graves, M.~Riedmiller, A.~K. Fidjeland, G.~Ostrovski \emph{et~al.},
  ``Human-level control through deep reinforcement learning,'' \emph{Nature},
  vol. 518, no. 7540, pp. 529--533, 2015.

\bibitem{faust2018prm}
A.~Faust, K.~Oslund, O.~Ramirez, A.~Francis, L.~Tapia, M.~Fiser, and
  J.~Davidson, ``{PRM-RL}: Long-range robotic navigation tasks by combining
  reinforcement learning and sampling-based planning,'' in \emph{Proceedings of
  the International Conference on Robotics and Automation}, 2018, pp.
  5113--5120.

\bibitem{chiang2019learning}
H.-T.~L. Chiang, A.~Faust, M.~Fiser, and A.~Francis, ``Learning navigation
  behaviors end-to-end with autorl,'' \emph{IEEE Robotics and Automation
  Letters}, vol.~4, no.~2, pp. 2007--2014, 2019.

\bibitem{francis2020long}
A.~Francis, A.~Faust, H.-T.~L. Chiang, J.~Hsu, J.~C. Kew, M.~Fiser, and
  T.-W.~E. Lee, ``Long-range indoor navigation with prm-rl,'' \emph{IEEE
  Transactions on Robotics}, vol.~36, no.~4, pp. 1115--1134, 2020.

\bibitem{mnih2016asynchronous}
V.~Mnih, A.~P. Badia, M.~Mirza, A.~Graves, T.~Lillicrap, T.~Harley, D.~Silver,
  and K.~Kavukcuoglu, ``Asynchronous methods for deep reinforcement learning,''
  in \emph{Proceedings of the 33$^{rd}$ International Conference on Machine
  Learning}, vol.~48, 2016, pp. 1928--1937.

\bibitem{bellemare2017distributional}
M.~G. Bellemare, W.~Dabney, and R.~Munos, ``A distributional perspective on
  reinforcement learning,'' in \emph{Proceedings of the 34$^{nd}$ International
  Conference on Machine Learning}, 2017, pp. 449--458.

\bibitem{espeholt2018impala}
L.~Espeholt, H.~Soyer, R.~Munos, K.~Simonyan, V.~Mnih, T.~Ward, Y.~Doron,
  V.~Firoiu, T.~Harley, I.~Dunning \emph{et~al.}, ``{IMPALA}: Scalable
  distributed deep-rl with importance weighted actor-learner architectures,''
  in \emph{Proceedings of the 35$^{th}$ International Conference on Machine
  Learning}.\hskip 1em plus 0.5em minus 0.4em\relax PMLR, 2018, pp. 1407--1416.

\bibitem{mccloskey1989catastrophic}
M.~McCloskey and N.~J. Cohen, ``Catastrophic interference in connectionist
  networks: The sequential learning problem,'' in \emph{Psychology of learning
  and motivation}.\hskip 1em plus 0.5em minus 0.4em\relax Elsevier, 1989,
  vol.~24, pp. 109--165.

\bibitem{kirkpatrick2017overcoming}
J.~Kirkpatrick, R.~Pascanu, N.~Rabinowitz, J.~Veness, G.~Desjardins, A.~A. Rusu
  \emph{et~al.}, ``Overcoming catastrophic forgetting in neural networks,'' in
  \emph{Proceedings of the National Academy of Sciences}, vol. 114, no.~13,
  2017, pp. 3521--3526.

\bibitem{fernando2017pathnet}
C.~Fernando, D.~Banarse, C.~Blundell, Y.~Zwols, D.~Ha, A.~A. Rusu, A.~Pritzel,
  and D.~Wierstra, ``Pathnet: Evolution channels gradient descent in super
  neural networks,'' \emph{arXiv preprint arXiv:1701.08734}, 2017.

\bibitem{lopez2017gradient}
D.~Lopez-Paz and M.~Ranzato, ``Gradient episodic memory for continual
  learning,'' in \emph{Proceedings of the 31$^{st}$ Conference on Neural
  Information Processing Systems}, 2017, pp. 6467--6476.

\bibitem{rebuffi2017icarl}
S.-A. Rebuffi, A.~Kolesnikov, G.~Sperl, and C.~H. Lampert, ``icarl: Incremental
  classifier and representation learning,'' in \emph{Proceedings of the IEEE
  Conference on Computer Vision and Pattern Recognition}, 2017, pp. 2001--2010.

\bibitem{mallya2018packnet}
A.~Mallya and S.~Lazebnik, ``Packnet: Adding multiple tasks to a single network
  by iterative pruning,'' in \emph{Proceedings of the IEEE Conference on
  Computer Vision and Pattern Recognition}, 2018, pp. 7765--7773.

\bibitem{delange2021continual}
M.~Delange, R.~Aljundi, M.~Masana, S.~Parisot, X.~Jia, A.~Leonardis,
  G.~Slabaugh, and T.~Tuytelaars, ``A continual learning survey: Defying
  forgetting in classification tasks,'' \emph{IEEE Transactions on Pattern
  Analysis and Machine Intelligence}, 2021.

\bibitem{kessler2020unclear}
S.~Kessler, J.~Parker-Holder, P.~Ball, S.~Zohren, and S.~J. Roberts,
  ``{UNCLEAR}: A straightforward method for continual reinforcement learning,''
  in \emph{Proceedings of the 37$^{th}$ International Conference on Machine
  Learning}, 2020.

\bibitem{khetarpal2020towards}
K.~Khetarpal, M.~Riemer, I.~Rish, and D.~Precup, ``Towards continual
  reinforcement learning: A review and perspectives,'' \emph{arXiv preprint
  arXiv:2012.13490}, 2020.

\bibitem{ghiassian2020improving}
S.~Ghiassian, B.~Rafiee, Y.~L. Lo, and A.~White, ``Improving performance in
  reinforcement learning by breaking generalization in neural networks,'' in
  \emph{Proceedings of the 19$^{th}$ International Conference on Autonomous
  Agents and Multiagent Systems}, 2020.

\bibitem{bengio2020interference}
E.~Bengio, J.~Pineau, and D.~Precup, ``Interference and generalization in
  temporal difference learning,'' in \emph{Proceedings of the 37$^{th}$
  International Conference on Machine Learning}, 2020, pp. 767--777.

\bibitem{schaul2019ray}
T.~Schaul, D.~Borsa, J.~Modayil, and R.~Pascanu, ``Ray interference: a source
  of plateaus in deep reinforcement learning,'' \emph{arXiv preprint
  arXiv:1904.11455}, 2019.

\bibitem{fedus2020catastrophic}
W.~Fedus, D.~Ghosh, J.~D. Martin, M.~G. Bellemare, Y.~Bengio, and
  H.~Larochelle, ``On catastrophic interference in atari 2600 games,''
  \emph{arXiv preprint arXiv:2002.12499}, 2020.

\bibitem{lo2019overcoming}
Y.~L. Lo and S.~Ghiassian, ``Overcoming catastrophic interference in online
  reinforcement learning with dynamic self-organizing maps,'' in
  \emph{Proceedings of the 33$^{rd}$ Conference on Neural Information
  Processing Systems}, 2019.

\bibitem{liu2019utility}
V.~Liu, R.~Kumaraswamy, L.~Le, and M.~White, ``The utility of sparse
  representations for control in reinforcement learning,'' in \emph{Proceedings
  of the 33$^{rd}$ AAAI Conference on Artificial Intelligence}, vol.~33,
  no.~01, 2019, pp. 4384--4391.

\bibitem{schaul2016prioritized}
T.~Schaul, J.~Quan, I.~Antonoglou, and D.~Silver, ``Prioritized experience
  replay,'' in \emph{Proceedings of the 4$^{th}$ International Conference on
  Learning Representations}, 2016.

\bibitem{fedus2020revisiting}
W.~Fedus, P.~Ramachandran, R.~Agarwal, Y.~Bengio, H.~Larochelle, M.~Rowland,
  and W.~Dabney, ``Revisiting fundamentals of experience replay,'' in
  \emph{Proceedings of the 37$^{th}$ International Conference on Machine
  Learning}, 2020, pp. 3061--3071.

\bibitem{dias2008skm}
J.~G. Dias and M.~J. Cortinhal, ``The skm algorithm: A k-means algorithm for
  clustering sequential data,'' in \emph{Proceedings of the Ibero-American
  Conference on Artificial Intelligence}, 2008, pp. 173--182.

\bibitem{zenke2017continual}
F.~Zenke, B.~Poole, and S.~Ganguli, ``Continual learning through synaptic
  intelligence,'' in \emph{Proceedings of the 34$^{th}$ International
  Conference on Machine Learning}, 2017, pp. 3987--3995.

\bibitem{golkar2019continual}
S.~Golkar, M.~Kagan, and K.~Cho, ``Continual learning via neural pruning,''
  \emph{arXiv preprint arXiv:1903.04476}, 2019.

\bibitem{seo2021state}
Y.~Seo, L.~Chen, J.~Shin, H.~Lee, P.~Abbeel, and K.~Lee, ``State entropy
  maximization with random encoders for efficient exploration,'' \emph{arXiv
  preprint arXiv:2102.09430}, 2021.

\bibitem{bellemare2013arcade}
M.~G. Bellemare, Y.~Naddaf, J.~Veness, and M.~Bowling, ``The arcade learning
  environment: An evaluation platform for general agents,'' \emph{Journal of
  Artificial Intelligence Research}, vol.~47, pp. 253--279, 2013.

\bibitem{lesort2020continual}
T.~Lesort, V.~Lomonaco, A.~Stoian, D.~Maltoni, D.~Filliat, and
  N.~D{\'\i}az-Rodr{\'\i}guez, ``Continual learning for robotics: Definition,
  framework, learning strategies, opportunities and challenges,''
  \emph{Information Fusion}, vol.~58, pp. 52--68, 2020.

\bibitem{isele2018selective}
D.~Isele and A.~Cosgun, ``Selective experience replay for lifelong learning,''
  in \emph{Proceedings of the 32$^{nd}$ AAAI Conference on Artificial
  Intelligence}, vol.~32, no.~1, 2018.

\bibitem{rolnick2019experience}
D.~Rolnick, A.~Ahuja, J.~Schwarz, T.~P. Lillicrap, and G.~Wayne, ``Experience
  replay for continual learning,'' in \emph{Proceedings of the 33$^{th}$
  Conference on Neural Information Processing Systems}, 2019.

\bibitem{atkinson2021pseudo}
C.~Atkinson, B.~McCane, L.~Szymanski, and A.~Robins, ``Pseudo-rehearsal:
  Achieving deep reinforcement learning without catastrophic forgetting,''
  \emph{Neurocomputing}, vol. 428, pp. 291--307, 2021.

\bibitem{hinton2015distilling}
G.~Hinton, O.~Vinyals, and J.~Dean, ``Distilling the knowledge in a neural
  network,'' in \emph{Workshop of the Conference on Neural Information
  Processing Systems}, 2015.

\bibitem{rusu2016policy}
A.~A. Rusu, S.~G. Colmenarejo, C.~Gulcehre, G.~Desjardins, J.~Kirkpatrick,
  R.~Pascanu \emph{et~al.}, ``Policy distillation,'' in \emph{Proceedings of
  the 5$^{th}$ International Conference on Learning Representations}, 2016.

\bibitem{gangwani2018policy}
T.~Gangwani and J.~Peng, ``Policy optimization by genetic distillation,'' in
  \emph{Proceedings of the 7$^{th}$ International Conference on Learning
  Representations}, 2018.

\bibitem{traore2019discorl}
R.~Traor{\'e}, H.~Caselles-Dupr{\'e}, T.~Lesort, T.~Sun, G.~Cai \emph{et~al.},
  ``Discorl: Continual reinforcement learning via policy distillation,'' in
  \emph{Workshop of the Conference on Neural Information Processing Systems},
  2019.

\bibitem{rusu2016progressive}
A.~A. Rusu, N.~C. Rabinowitz, G.~Desjardins, H.~Soyer, J.~Kirkpatrick,
  K.~Kavukcuoglu, R.~Pascanu, and R.~Hadsell, ``Progressive neural networks,''
  \emph{arXiv preprint arXiv:1606.04671}, 2016.

\bibitem{van2016deep}
H.~Van~Hasselt, A.~Guez, and D.~Silver, ``Deep reinforcement learning with
  double q-learning,'' in \emph{Proceedings of the 30$^{th}$ AAAI Conference on
  Artificial Intelligence}, vol.~30, no.~1, 2016.

\bibitem{hessel2018rainbow}
M.~Hessel, J.~Modayil, H.~Van~Hasselt, T.~Schaul, G.~Ostrovski, W.~Dabney,
  D.~Horgan, B.~Piot, M.~Azar, and D.~Silver, ``Rainbow: Combining improvements
  in deep reinforcement learning,'' in \emph{Proceedings of the 32$^{nd}$ AAAI
  Conference on Artificial Intelligence}, vol.~32, no.~1, 2018.

\bibitem{padakandla2020reinforcement}
S.~Padakandla, K.~Prabuchandran, and S.~Bhatnagar, ``Reinforcement learning
  algorithm for non-stationary environments,'' \emph{Applied Intelligence},
  vol.~50, no.~11, pp. 3590--3606, 2020.

\bibitem{lomonaco2020continual}
V.~Lomonaco, K.~Desai, E.~Culurciello, and D.~Maltoni, ``Continual
  reinforcement learning in 3d non-stationary environments,'' in
  \emph{Proceedings of the IEEE/CVF Conference on Computer Vision and Pattern
  Recognition Workshops}, 2020, pp. 248--249.

\bibitem{jain2020algorithmic}
V.~Jain, W.~Fedus, H.~Larochelle, D.~Precup, and M.~G. Bellemare, ``Algorithmic
  improvements for deep reinforcement learning applied to interactive
  fiction,'' in \emph{Proceedings of the 34$^{th}$ AAAI Conference on
  Artificial Intelligence}, vol.~34, no.~04, 2020, pp. 4328--4336.

\bibitem{milan2016forget}
K.~Milan, J.~Veness, J.~Kirkpatrick, M.~Bowling, A.~Koop, and D.~Hassabis,
  ``The forget-me-not process,'' in \emph{Proceedings of the 30$^{th}$
  Conference on Neural Information Processing Systems}, vol.~29, 2016, pp.
  3702--3710.

\bibitem{rao2019continual}
D.~Rao, F.~Visin, A.~A. Rusu, Y.~W. Teh, R.~Pascanu, and R.~Hadsell,
  ``Continual unsupervised representation learning,'' in \emph{Proceedings of
  the 33$^{th}$ Conference on Neural Information Processing Systems}, 2019.

\bibitem{ghosh2018divide}
D.~Ghosh, A.~Singh, A.~Rajeswaran, V.~Kumar, and S.~Levine,
  ``Divide-and-conquer reinforcement learning,'' in \emph{Proceedings of the
  7$^{th}$ International Conference on Learning Representations}, 2018.

\bibitem{hadsell2020embracing}
R.~Hadsell, D.~Rao, A.~A. Rusu, and R.~Pascanu, ``Embracing change: Continual
  learning in deep neural networks,'' \emph{Trends in Cognitive Sciences},
  2020.

\bibitem{liu2020towards}
V.~Liu, A.~White, H.~Yao, and M.~White, ``Towards a practical measure of
  interference for reinforcement learning,'' \emph{arXiv preprint
  arXiv:2007.03807}, 2020.

\bibitem{brockman2016openai}
G.~Brockman, V.~Cheung, L.~Pettersson, J.~Schneider, J.~Schulman, J.~Tang, and
  W.~Zaremba, ``Openai gym,'' \emph{arXiv preprint arXiv:1606.01540}, 2016.

\bibitem{castro18dopamine}
P.~S. Castro, S.~Moitra, C.~Gelada, S.~Kumar, and M.~G. Bellemare, ``Dopamine:
  {A} {R}esearch {F}ramework for {D}eep {R}einforcement {L}earning,''
  \emph{arXiv preprint arXiv:1812.06110}, 2018.

\bibitem{kingma2014adam}
D.~P. Kingma and J.~Ba, ``Adam: A method for stochastic optimization,'' in
  \emph{Proceedings of the 3$^{rd}$ International Conference on Learning
  Representations}, 2014.

\bibitem{pan2019fuzzy}
Y.~Pan, K.~Banman, and M.~White, ``Fuzzy tiling activations: A simple approach
  to learning sparse representations online,'' in \emph{Proceedings of the
  9$^{th}$ International Conference on Learning Representations}, 2021.

\end{thebibliography}

\vspace{-5cm}
\begin{IEEEbiography}[{\includegraphics[width=1in,height=1.25in,clip,keepaspectratio]{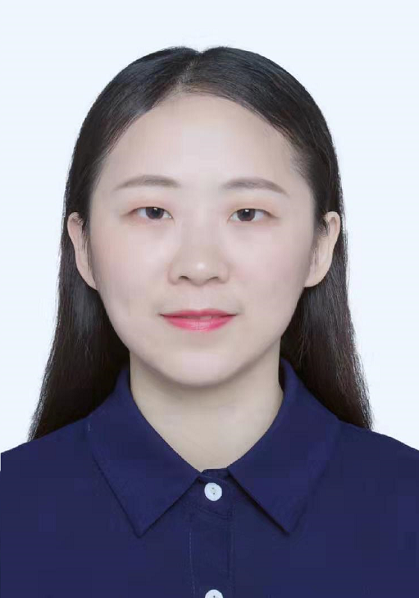}}]{Tiantian Zhang}
is currently working toward her Ph.D. degree in control science and engineering with the Department of Automation, Tsinghua University, Beijing, China. She received her B.Sc. degree in automation from the Department of Information Science and Technology, Central South University, Changsha, China, in 2015, and the M.Sc. degree in control engineering with the Department of Automation, Tsinghua University, Beijing, China, in 2018. Her research interests include data science, decision making, and reinforcement learning.
\end{IEEEbiography}

\vspace{-5cm} 

\begin{IEEEbiography}[{\includegraphics[width=1in,height=1.25in,clip,keepaspectratio]{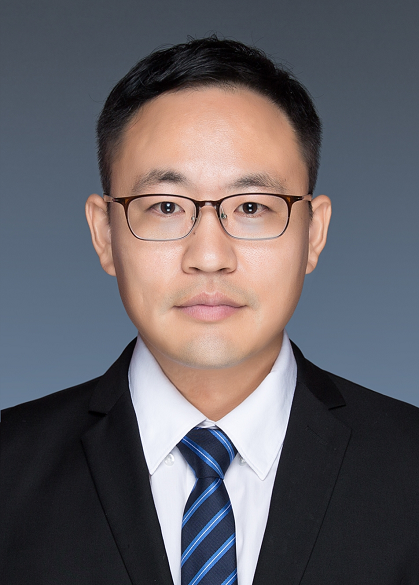}}]{Xueqian Wang}
(Member, IEEE) received his M.Sc and Ph.D. degrees in control science and engineering from Harbin Institute of Technology (HIT), Harbin, China, in 2005 and 2010, respectively. From June 2010 to February 2014, he was a Postdoctoral Researcher at HIT. From March 2014 to November 2019, he was an Associate Professor with the Division of Informatics, Shenzhen International Graduate School, Tsinghua University, Shenzhen, China. He is currently a Professor and the leader of the Center for Artificial Intelligence and Robotics, Shenzhen International Graduate School, Tsinghua University. His Research interests include Robot dynamics and control, teleoperation, intelligent decision-making and game-playing, and fault diagnosis.
\end{IEEEbiography}

\newpage
\begin{IEEEbiography}[{\includegraphics[width=1in,height=1.25in,clip,keepaspectratio]{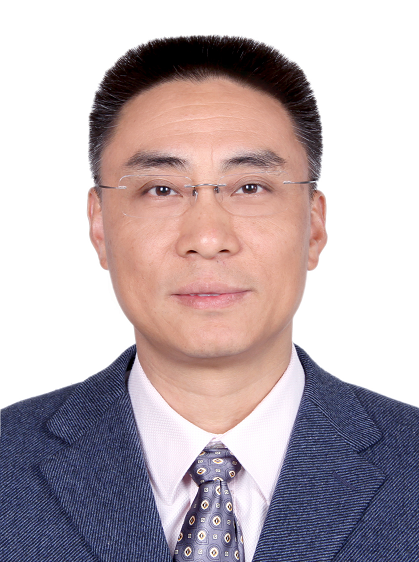}}]{Bin Liang}
(Senior Member, IEEE) is currently a Professor with the Research Center for Navigation and Control, Department of Automation, Tsinghua University, Beijing, China. He received his B.Sc. and M.Sc. degrees in control engineering from the Honors College of Northwestern Polytechnical University, Xi’an, China, in 1989 and 1991, respectively, and Ph.D. degree in control engineering from the Department of Precision Instrument, Tsinghua University, Beijing, China, in 1994. From 1994 to 2003, he held his positions as a Postdoctoral Researcher, an Associate Researcher, and a Researcher with China Academy of Space Technology (CAST), Beijing, China. From 2003 to 2007, he held his positions as a researcher and an Assistant Chief Engineer with the China Aerospace Science and Technology Corporation, Beijing. His research interests include modeling and control of intelligent robotic systems, teleoperation, intelligent sensing technology.
\end{IEEEbiography}

\vspace{-13.5cm} 
\begin{IEEEbiography}[{\includegraphics[width=1in,height=1.25in,clip,keepaspectratio]{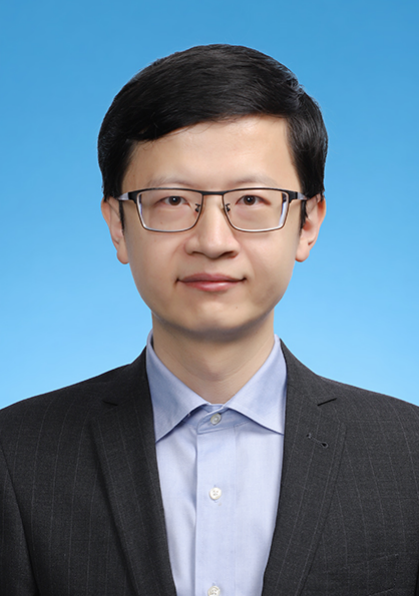}}]{Bo Yuan}
(Senior Member, IEEE) received the B.E. degree in computer science from the Nanjing University of Science and Technology, Nanjing, China, in 1998, and the M.Sc. and Ph.D. degrees in computer science from The University of Queensland (UQ), St Lucia, QLD, Australia, in 2002 and 2006, respectively. From 2006 to 2007, he was a Research Officer on a project funded by the Australian Research Council, UQ. He is currently an Associate Professor with the Division of Informatics, Shenzhen International Graduate School, Tsinghua University, Shenzhen, China. He has authored or coauthored more than 110 articles in refereed international conferences and journals. His research interests include data science, evolutionary computation, and reinforcement learning.
\end{IEEEbiography}

\newpage
\appendices
\section{Illustration of Context Division}
\subsection{Sequential K-Means Clustering}
\label{Sequential K-Means Clustering}
The process of {\em Sequential K-Means Clustering} for the current state $s$ is shown in Algorithm \ref{alg:Sequential_K-Means_Clustering}. Each context centroid $c_i$ is the average of all of the states $s$ closest to $c_i$. In order to get a better initialization of $\mathcal{C}$, we can perform offline {\em K-Means} clustering on all states experienced before training starts and set the results of centroids as the initial $\mathcal{C}$. Then, {\em Sequential K-Means Clustering} is performed in subsequent time steps.

\begin{algorithm}[ht]
    \caption{SKM: Sequential K-Means Clustering}
    \label{alg:Sequential_K-Means_Clustering}
    \textbf{Input}: Current state \textbf{$s$};
    \newline \hspace*{0.96cm} Initial context centroids $\mathcal{C}=\{c_1,c_2,\dots,c_k\}$.\\
    \textbf{Output}: Updated centroids $\mathcal{C}$.
\begin{algorithmic}[1] 
    \STATE Count the number of samples in each context:
    \newline \hspace*{1.6cm} $\mathcal{N}=\{n_1,n_2,\dots,n_k\}$
    \IF{$s$ is closest to centroid $c_i$}
        \STATE Increment $n_i$: $n_i=n_i+1$;
        \STATE Update $c_i$: $c_i=c_i+(1/n_i)*(s-c_i)$;
    \ENDIF
\STATE \textbf{return} $\mathcal{C}$.
\end{algorithmic}
\end{algorithm}

\begin{figure}[ht]
    \centering
    \setlength{\abovecaptionskip}{5pt}
    \subfigure[ISC\label{fig:ISC}]
        {\includegraphics[width=0.47\linewidth]{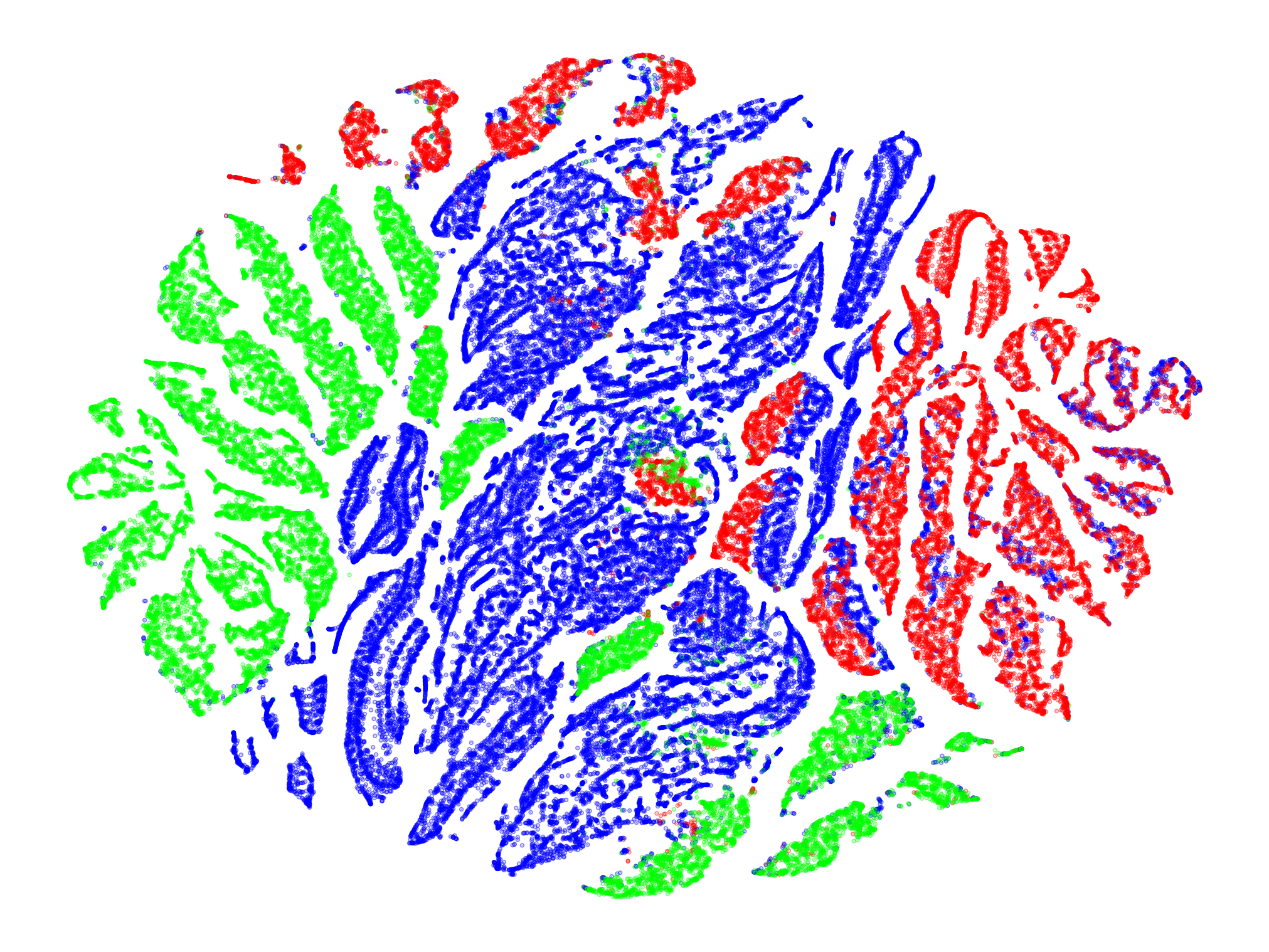}}
    \hspace{2mm}
    \subfigure[ESC\label{fig:ESC}]
        {\includegraphics[width=0.47\linewidth]{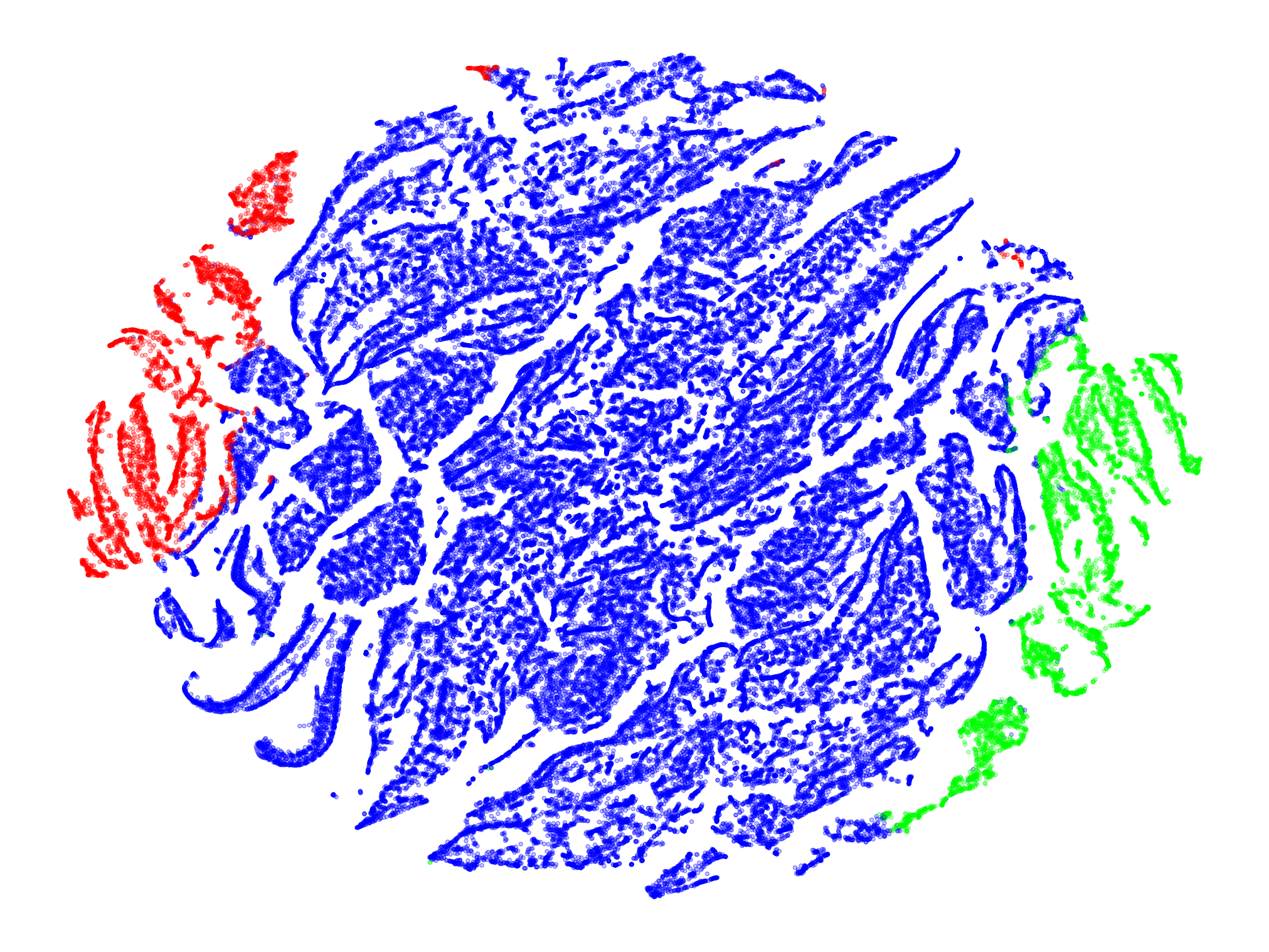}}
    \caption{Two-dimensional t-SNE results of context division according to different kinds of clustering objects on {\em CartPole-v0}: (a) ISC vs (b) ESC. Here, the colors represent different contexts of which the points represent the states within the corresponding context.}
    \label{fig:CD_objects_comparison}
\end{figure}

\begin{algorithm}[t]
    \caption{IQ-RE: Interference-aware Deep Q-learning with Random Encoder}
    \label{alg:IQ-RE}
    \textbf{Input}: Initial replay buffer $\mathcal{B}$ with capacity $N$;
    \newline \hspace*{0.95cm} Initial $Q$-function $f_\theta$ with random weights $\theta$;
    \newline \hspace*{0.95cm} Initial target $\hat{Q}$-function $f_{\theta^-}$ with weights $\theta^-=\theta$;
    \newline \hspace*{0.95cm} Initial random encoder $f_{\theta_{re}}$ with weights $\theta_{re}$;
    \newline \hspace*{0.95cm} Initial context centroids $\mathcal{C}=\{c_1,c_2,\dots,c_k\}$;
    \newline \hspace*{0.95cm} Initial target context centroids $\hat{\mathcal{C}}=\mathcal{C}$.\\
    \textbf{Parameter}: Total training steps $T$, the number of contexts $k$,
    \newline \hspace*{1.88cm}the output dimension of random encoder $d$, target
    \newline \hspace*{1.88cm}update period $C$, learning rate $\alpha$. \\
    \textbf{Output}: Updated $\mathcal{C}$ and $f_\theta$.
\begin{algorithmic}[1] 
    \STATE Initial state $s$;
    \FOR{$t = 1, T$}
        \STATE Interact with environment to obtain $\{s_t,a_t,r_t,s_{t+1}\}$.
        \STATE State encoding: get the fixed representation for $s_t$ and $s_{t+1}$, $y_t=f_{\theta_{re}}(s_t)$, $y_{t+1}=f_{\theta_{re}}(s_{t+1})$.
        \STATE States assignment: $\omega(s_t)\xleftarrow{\hat{\mathcal{C}}}y_t$, $\omega(s_{t+1})\xleftarrow{\hat{\mathcal{C}}}y_{t+1}$.
        \STATE Store transition $\{s_t,\omega(s_t),a_t,r_t,s_{t+1},\omega(s_{t+1})\}$ in $\mathcal{B}$.
        \STATE Context centroids update: $\mathcal{C}\leftarrow SKM(y_t,\mathcal{C})$.
        \STATE Joint optimization: \\
        \hspace*{0.2cm} Sample mini-batch $\{s_i,\omega(s_i),a_i,r_i,s_i^\prime,\omega(s_i^\prime)\}_{i=1}^{m}$; \\
        \hspace*{0.2cm} Calculate $\mathcal{L}_{ori}$, $\mathcal{L_D}$ according to Eqs. \eqref{eq:dqn_loss} and \eqref{eq:distillation_loss}; \\
        \hspace*{0.2cm} Update parameter: $\theta\leftarrow\theta-\alpha\nabla_{\theta}(\mathcal{L}_{ori}+\lambda\mathcal{L_D})$.
        \IF{$t\mod {C}==0$} 
            \STATE $\theta^-=\theta$, $\hat{\mathcal{C}}=\mathcal{C}$
        \ENDIF
    \ENDFOR
\end{algorithmic}
\end{algorithm}

\begin{figure}[!t]
    \centering
    \setlength{\abovecaptionskip}{0pt}
    \subfigure[base RL: DQN]{
        \begin{minipage}[b]{0.48\textwidth}
            \centering
            \includegraphics[width=0.46\linewidth]{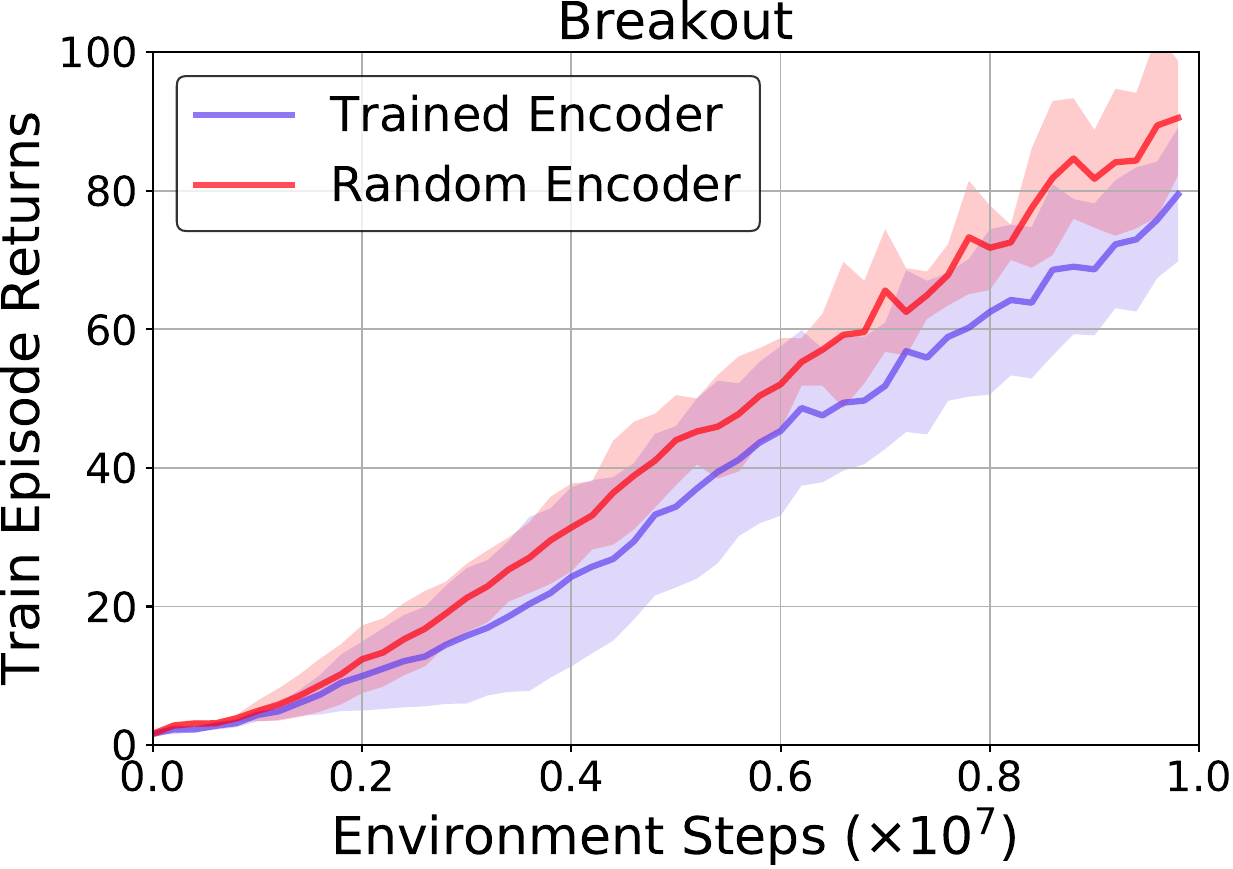}
            \hspace{0.1cm}
            \includegraphics[width=0.44\linewidth]{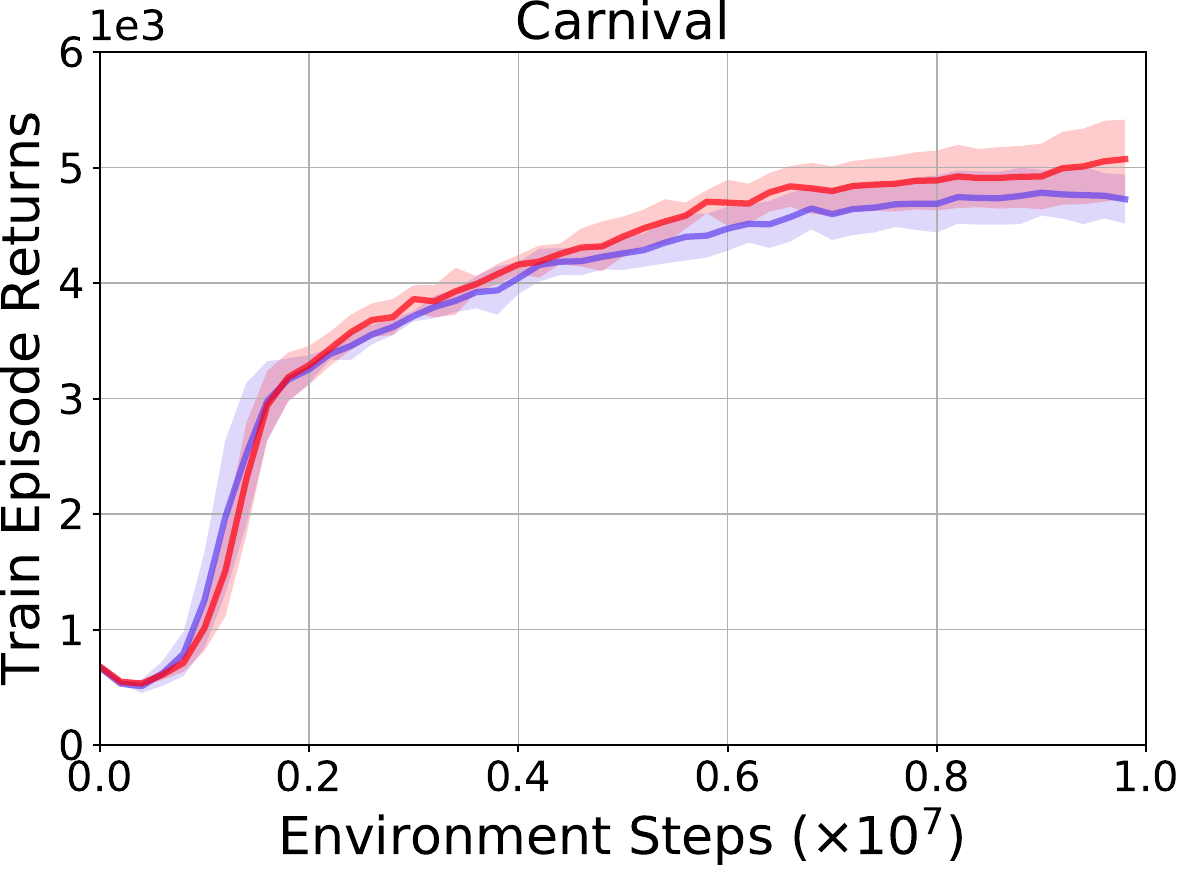}
        \end{minipage}
    }\\
    \subfigure[base RL: Rainbow]{
        \begin{minipage}[b]{0.48\textwidth}
            \centering
            \includegraphics[width=0.46\linewidth]{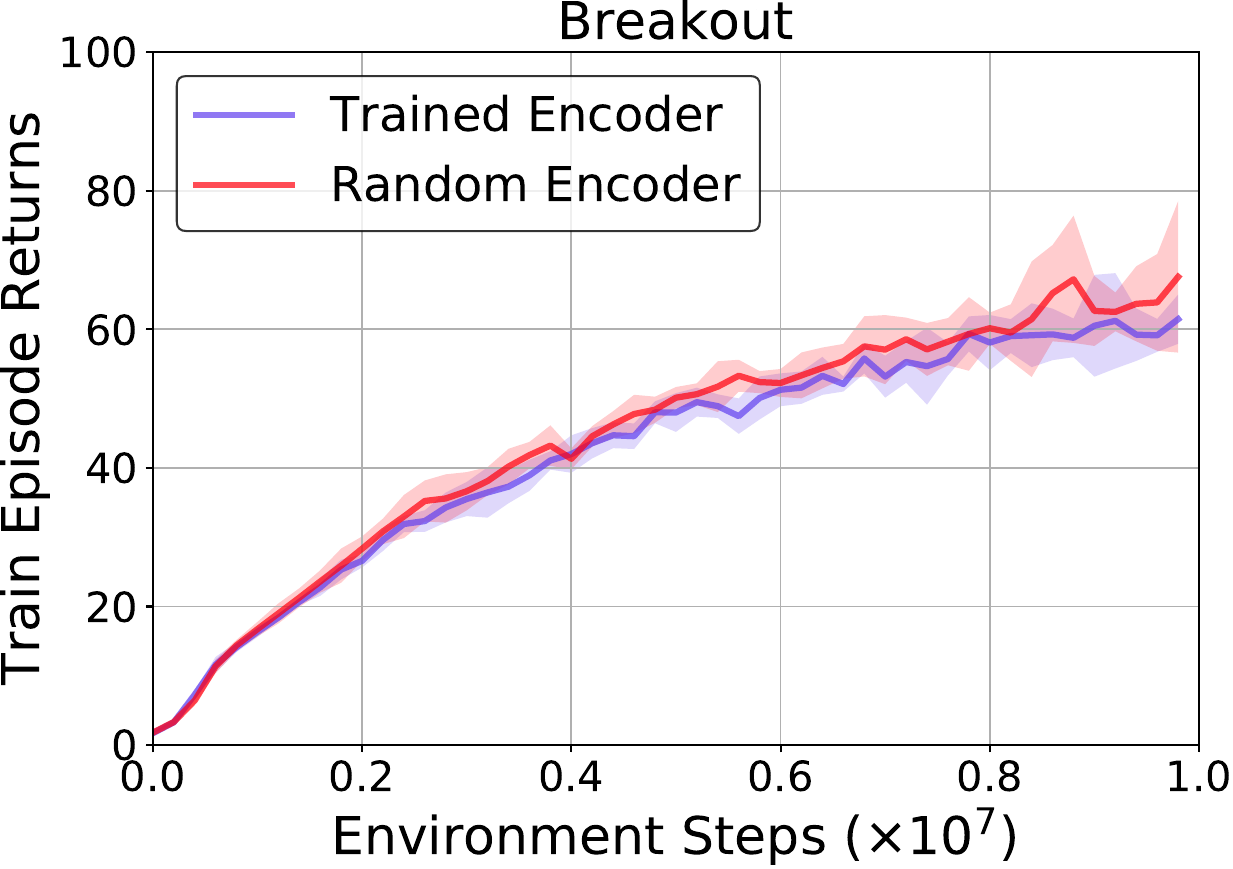}
            \hspace{0.1cm}
            \includegraphics[width=0.44\linewidth]{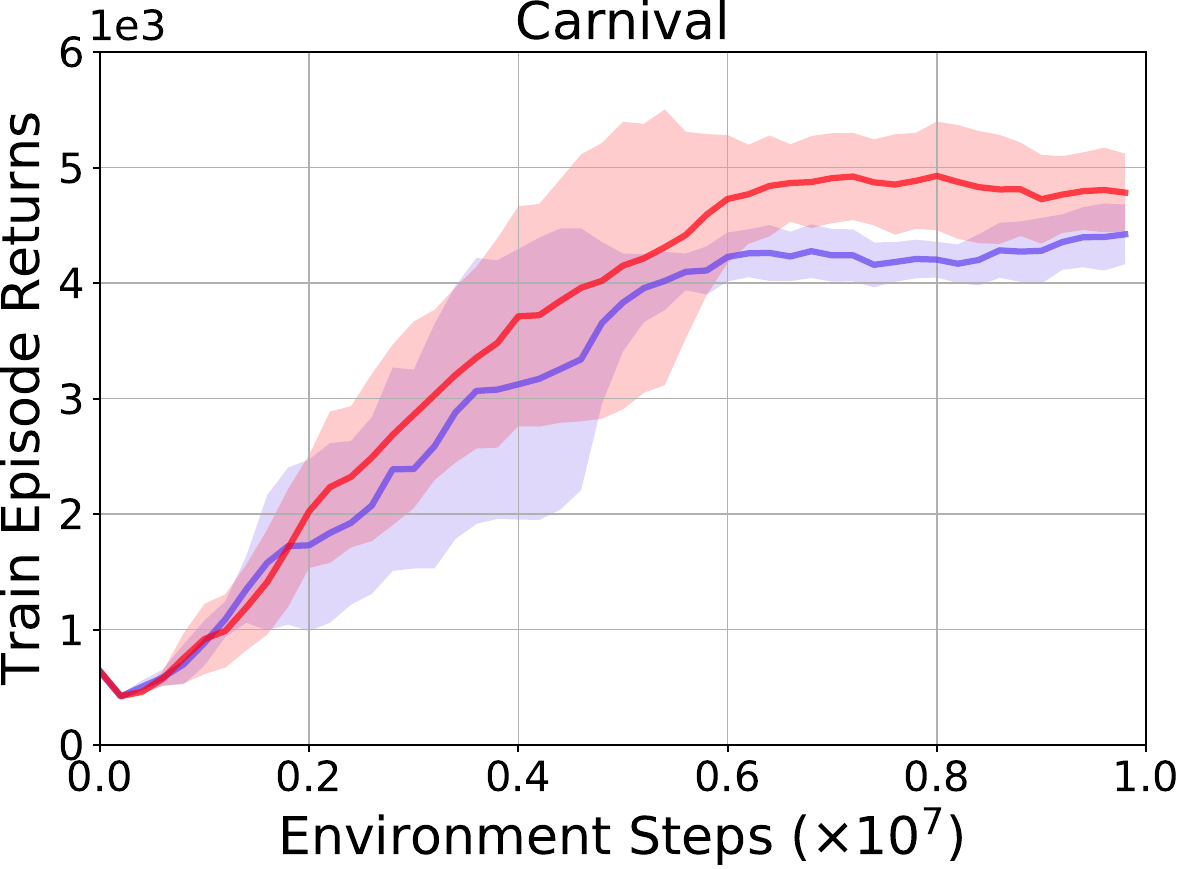}
        \end{minipage}  
    }
    \caption{Learning curves for IQ with different encoders. Here, $N$ is set to $10,000$. The solid lines and shaded regions represent the mean and standard deviation, respectively, across five runs. In summary, the performance of IQ with the random encoder is significantly better than that of IQ with the underlying RL trained encoder.}
    \label{fig:RE_TE_Compare}
\end{figure}

\subsection{Initial States vs All Experienced States}
\label{ISC_vs_ESC}
We compare the effects of two kinds of context division techniques: 
\begin{itemize}
\item {\bf ISC} (Initial States Clustering) means performing context division using {\em K-Means} on the samples sampled from the initial states distribution, referring to \cite{ghosh2018divide};
\item {\bf ESC} (Experienced States Clustering) is our proposed method that performs context division using {\em Sequential K-Means Clustering} on all states experienced during training process.
\end{itemize}

We run the DQN agent incorporated with IQ with the above two clustering techniques on {\em CartPole-v0}, respectively, and visualize the two-dimensional t-SNE results of context division in Fig. \ref{fig:CD_objects_comparison}. It can be clearly observed that the contexts divided by ESC are relatively independent and there is almost no overlapping area among contexts, achieving effective decoupling among states with different distributions. By contrast, there are obvious overlapping areas among the contexts divided by ISC, since the trajectories starting from the initial states within different contexts have a high likelihood of overlapping in subsequent time steps, which is not desirable for reducing interference on neural network training among differently distributed states.

\begin{figure*}[t]
  \centering
  \setlength{\abovecaptionskip}{5pt}
    {\includegraphics[width=0.98\linewidth]{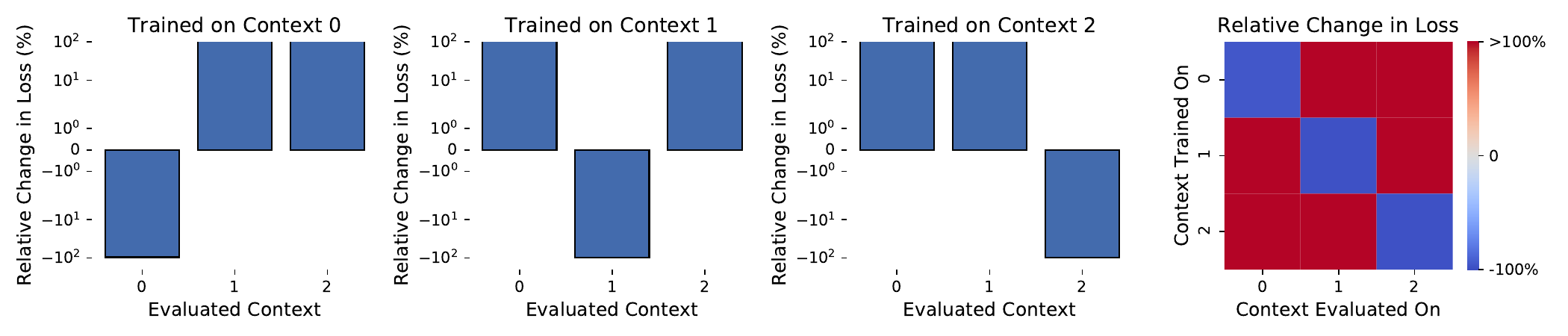}}
  \caption{Measuring the interference among contexts by clustering all experienced states when the agent is trained on {\em Pendulum-v0} for 400K environment steps ($k=3$). This result is obtained using the same experimental setup and method as for {\em CartPole-v0} in Fig. \ref{fig:Interference_btw_Context_cartpole}, and we can draw a completely consistent conclusion with that in Fig. \ref{fig:Interference_btw_Context_cartpole}.}
  \label{fig:Interference_btw_Context_pendulum}
\end{figure*}

\begin{figure*}[!t]
  \centering
  \setlength{\abovecaptionskip}{5pt}
  {\includegraphics[width=0.98\linewidth]{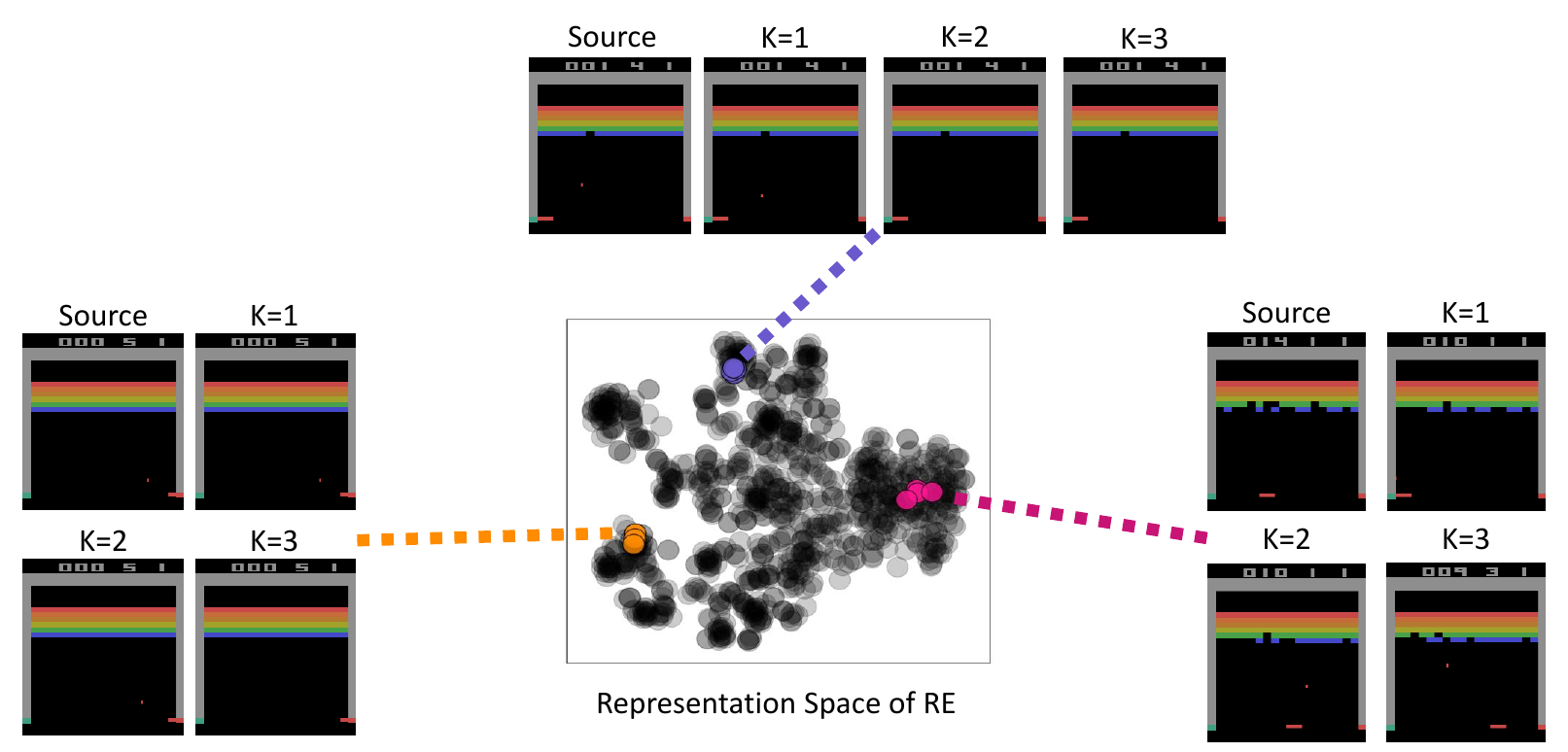}}
  \caption{Two-dimensional t-SNE visualization of {\em K-nearest neighbors} of states found by measuring distances in the representation space of a Random Encoder on {\em Breakout}. We observe that the representation space of a randomly initialized encoder effectively captures information about similarity between states.}
  \label{fig:RE}
\end{figure*}

\subsection{Interference Among Contexts on Pendulum-v0}
\label{Interference Among Contexts}
Similar to Fig. \ref{fig:Interference_btw_Context_cartpole}, we further investigate the interference among contexts obtained by our context division method on {\em Pendulum-v0}, and the result is shown in Fig. \ref{fig:Interference_btw_Context_pendulum}. The conclusion obtained from this additional experimental result is consistent with that in Fig. \ref{fig:Interference_btw_Context_cartpole}.

\section{Random Encoder}
\subsection{Representation Space Visualization of Random Encoder}
\label{Visualization Random Encoder}
To verify the feature that the random encoder can preserve the distance similarity of input samples, We find the {\em K-nearest neighbors} of some specific states by measuring the distances in the low-dimensional representation space produced by a randomly initialized encoder (Random Encoder) on {\em Breakout}. The results are shown in Fig. \ref{fig:RE} from which we can observe that the raw images corresponding to the {\em K-nearest neighbors} of the source image in the representation space of the random encoder demonstrate significant similarities.

\begin{table}[ht]
\centering
\caption{Number of FLOPs used by each agent at 10M\\ environment steps in {\em Breakout} (See\\ Fig. \ref{fig:RE_TE_Compare} for corresponding\\ learning curves.)}
\label{table:RE_TE_Compare}
\begin{tabular}{c|cc|cc}
\toprule
Base RL Method       & \multicolumn{2}{c|}{DQN}              & \multicolumn{2}{c}{Rainbow}        \\ \hline 
Encoder              & TE            & RE                   & TE           & RE                   \\ \hline 
PFOLPs               & 13.676        & 13.962               & 13.946       & 14.232               \\ 
\bottomrule
\end{tabular}
\end{table}

\begin{table*}[t]
\centering
\caption{The neural networks architecture of the underlying RL models ({\em i.e.}, DQN\\ and Rainbow) used in our experiments}
\label{table:architecture}
\begin{tabular}{ccccccccc}
\toprule
Tasks                                           & Layer & Input                      & Filter size    & Stride    & Num filters         & Activation    & Output                 \\   \hline  \specialrule{0em}{1pt}{1pt}
\multirow{2}{*}{Classic Control Tasks}          & FC1   & Dimension of state space   & -              & -         & $64$                & Tanh          & $64$                   \\
                                                & FC2   & $64$                       & -              & -         & Number of actions   & Linear        & Number of actions      \\    
\multirow{5}{*}{Atari Games}                    & Conv1 & $84\times84\times4$        & $8\times8$     & $4$       & $32$                & ReLU          & $20\times20\times32$   \\
                                                & Conv2 & $20\times20\times32$       & $4\times4$     & $2$       & $64$                & ReLU          & $9\times9\times64$     \\
                                                & Conv3 & $9\times9\times64$         & $3\times3$     & $1$       & $64$                & ReLU          & $7\times7\times64$     \\
                                                & FC4   & $7\times7\times64$         & -              & -         & $512$               & ReLU          & $512$                  \\
                                                & FC5   & $512$                      & -              & -         & $18$                & Linear        & $18$                   \\   
\bottomrule
\end{tabular}
\end{table*}

\begin{table*}[ht]
\centering
\caption{The common hyperparameters of the underlying RL models ({\em i.e.}, DQN\\ and Rainbow) used in our experiments}
\label{table:hyperparameters}
\begin{tabular}{ccc}
\toprule
Hyperparameter                  & Classic Control Tasks                                                                                                                                                & Atari Games          \\ \hline \specialrule{0em}{1pt}{1pt}
Training time step              & \begin{tabular}[c]{@{}c@{}}$400,000$ steps for {\em CartPole-v0} and {\em Pendulum-v0}\\ $1,000,000$ steps for {\em CartPole-v1} and {\em Acrobot-v1} \end{tabular}  & $10,000,000$ steps     \\
Training $\epsilon$             & $0.02$                                                                                                                                                               & $0.01$                 \\
$\epsilon$ decay schedule       & \begin{tabular}[c]{@{}c@{}}$40,000$ steps for {\em CartPole-v0} and {\em Pendulum-v0}\\  $100,000$ steps for {\em CartPole-v1} and {\em Acrobot-v1} \end{tabular}    & $250,000$ steps        \\
Min. history to start learning  & $1,000$ steps                                                                                                                                                        & $20,000$ steps         \\
Target network update frequency & $1,000$ steps                                                                                                                                                        & $8,000$ steps          \\
Batch size                      & $32$                                                                                                                                                                 & $32$                   \\
Learning rate $\alpha$          & $0.0005$                                                                                                                                                             & \begin{tabular}[c]{@{}c@{}}$0.00025$ for DQN\\ $0.0000625$ for Rainbow\end{tabular}                        \\ 
\bottomrule
\end{tabular}
\end{table*}

\subsection{Random Encoder vs. RL Trained Encoder}
\label{Random Encoder vs RL Trained Encoder}
For high-dimensional complex state spaces, a simple and intuitive idea is to perform clustering on the low-dimensional representation space of the underlying RL trained encoder directly. However, the RL trained encoder is constantly updated during learning, which means that the low-dimensional output results corresponding to the same input may be different with the progress of learning. Therefore, clustering based on the RL trained encoder is likely to bring extra inaccuracy and instability into context division. To complement the above analysis, additional experiments are conducted on two Atari games {\em Breakout} and {\em Carnival}. With the underlying RL trained encoder and the randomly initialized encoder respectively using two base RL algorithms ({\em i.e.,} DQN and Rainbow), and the results are shown in Fig. \ref{fig:RE_TE_Compare}. Furthermore, the numbers of FLOPs executed by both agents on {\em Breakout} are reported in Table \ref{table:RE_TE_Compare}. Compared with the RL trained encoder, IQ with the randomly initialized encoder achieves distinctly better performance with only about $2\%$ overhead. The full procedure of IQ with random encoder (named IQ-RE) is shown in Algorithm \ref{alg:IQ-RE}.

\section{Implementation Details}
\label{Implementation Details}
\subsection{Implementation Details for the Underlying RL Model}
To ensure the fairness of comparison, our results compare the agents based on the underlying RL model with the same hyperparameters and neural network architecture. We provide a full list of neural networks architecture of the underlying RL models in Table \ref{table:architecture} and summarize our choices for common key hyperparameters in Table \ref{table:hyperparameters}.

\subsection{Implementation Details for All Other Methods}
In our method and all baseline methods, the part of the underlying RL model follows the aforementioned network architecture and hyperparameter settings, and the method-specific parameter settings are summarized as follows:
\begin{itemize}
    \item SRNN: the sparsity level $\beta$ is set to 0.1, and the coefficient $\lambda_{KL}$ of distributional regularizes is also set to 0.1 for all tasks;
    \item DSOM: the DSOM learning rate $\epsilon$ is set to 0.25, the elasticity $\eta$ is set to 1.0, $\kappa$ is set to 0.5, and the number of DSOM weight vectors is set to 64 for all classic control tasks;
    \item TCNN: it uses binning operation to turn each raw input variable to 20 binary variables, and the results of all classic control tasks are generated with a single tiling;
    \item IQ: the coefficient $\lambda$ of distillation loss term is set in accordance with the exploration proportion $\epsilon$ of the agent in all tasks: $\lambda=1-\epsilon$, the number $k$ of context to be divided is set to 3 for all classic control tasks and 4 for all Atari games, and the output dimension of random encoder $d$ is set to 50 for all Atari games.
\end{itemize}

\section{Additional Experimental Results}

\begin{figure*}[t]
    \centering
    \setlength{\abovecaptionskip}{3pt}
    \subfigure[$N=50,000$]{
        \begin{minipage}[b]{0.98\textwidth}
			\includegraphics[width=0.24\textwidth]{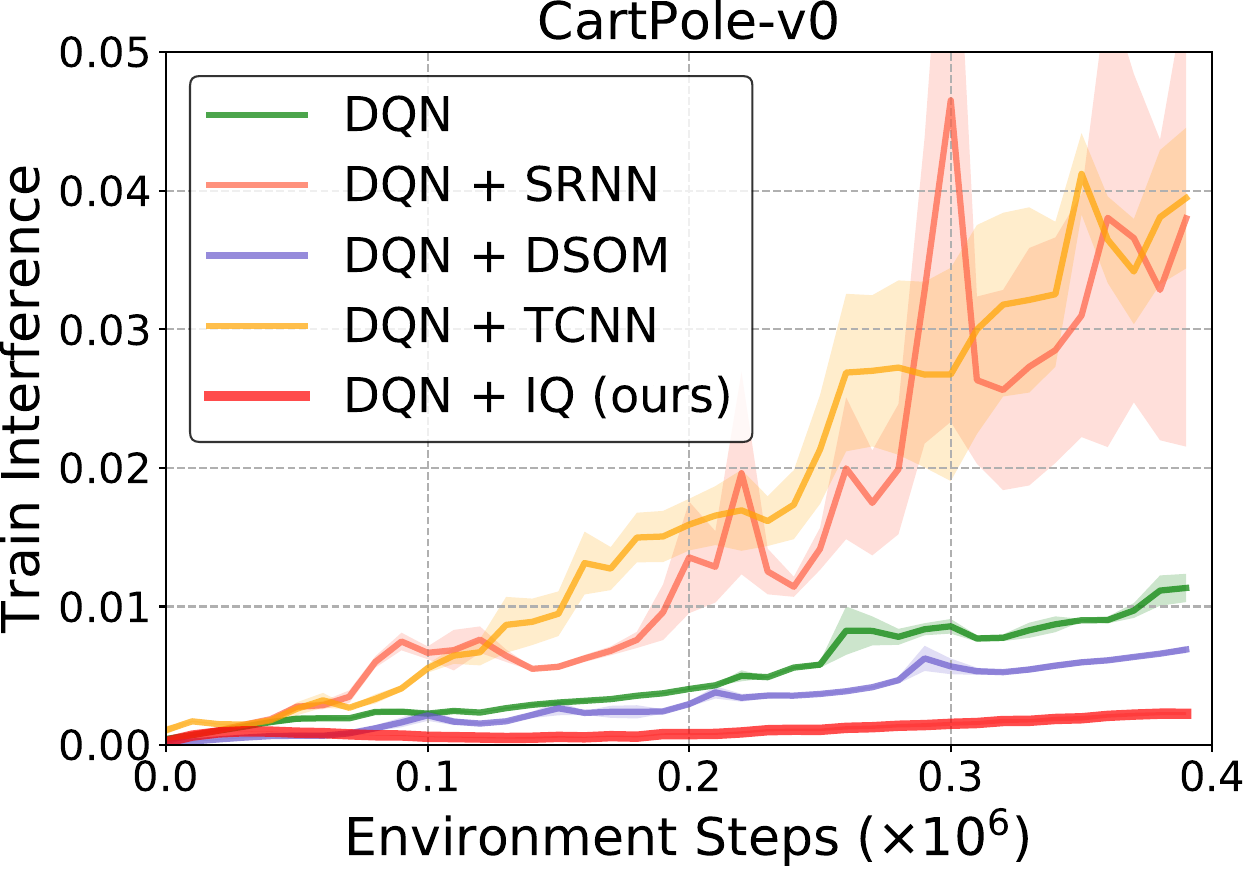}
			\includegraphics[width=0.252\textwidth]{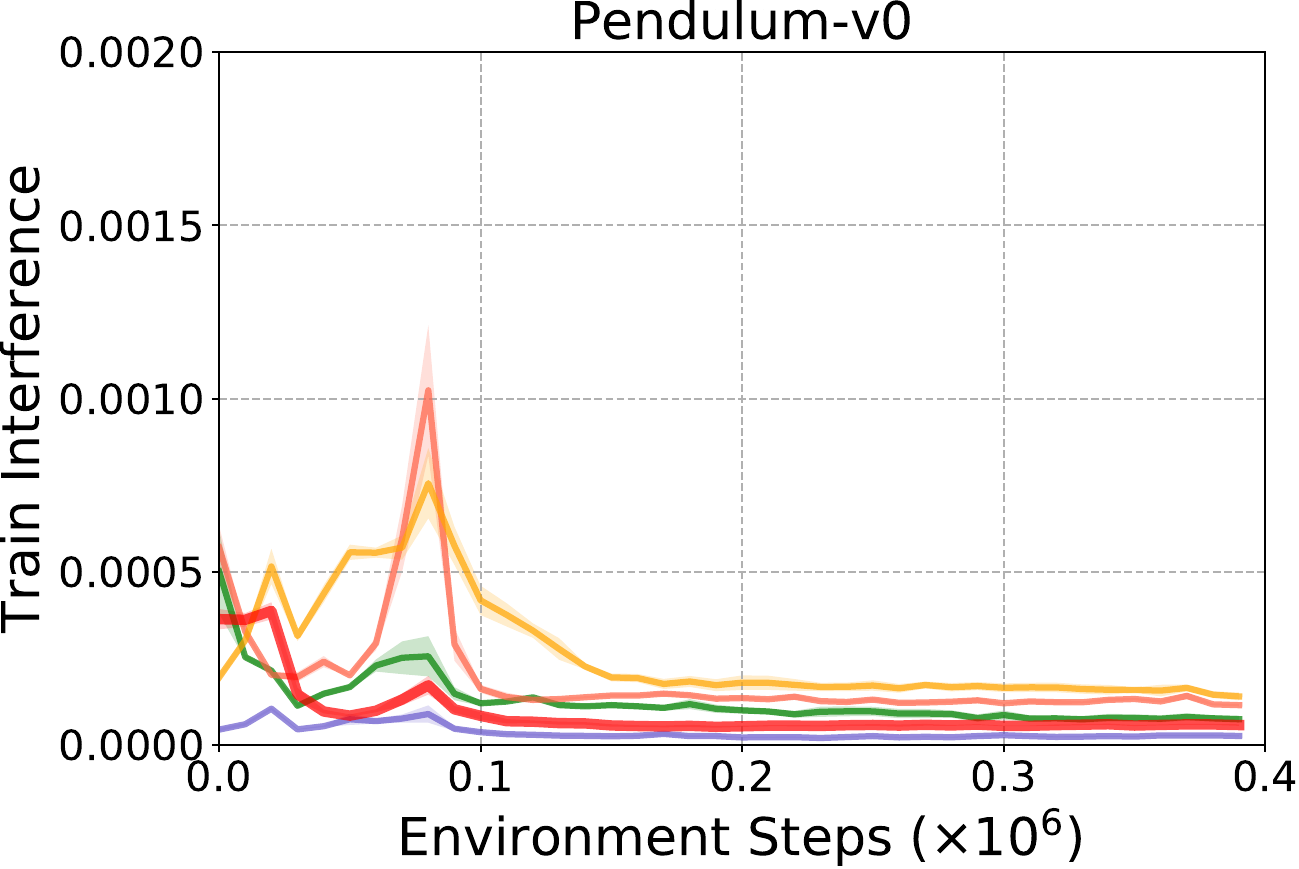}
			\includegraphics[width=0.24\textwidth]{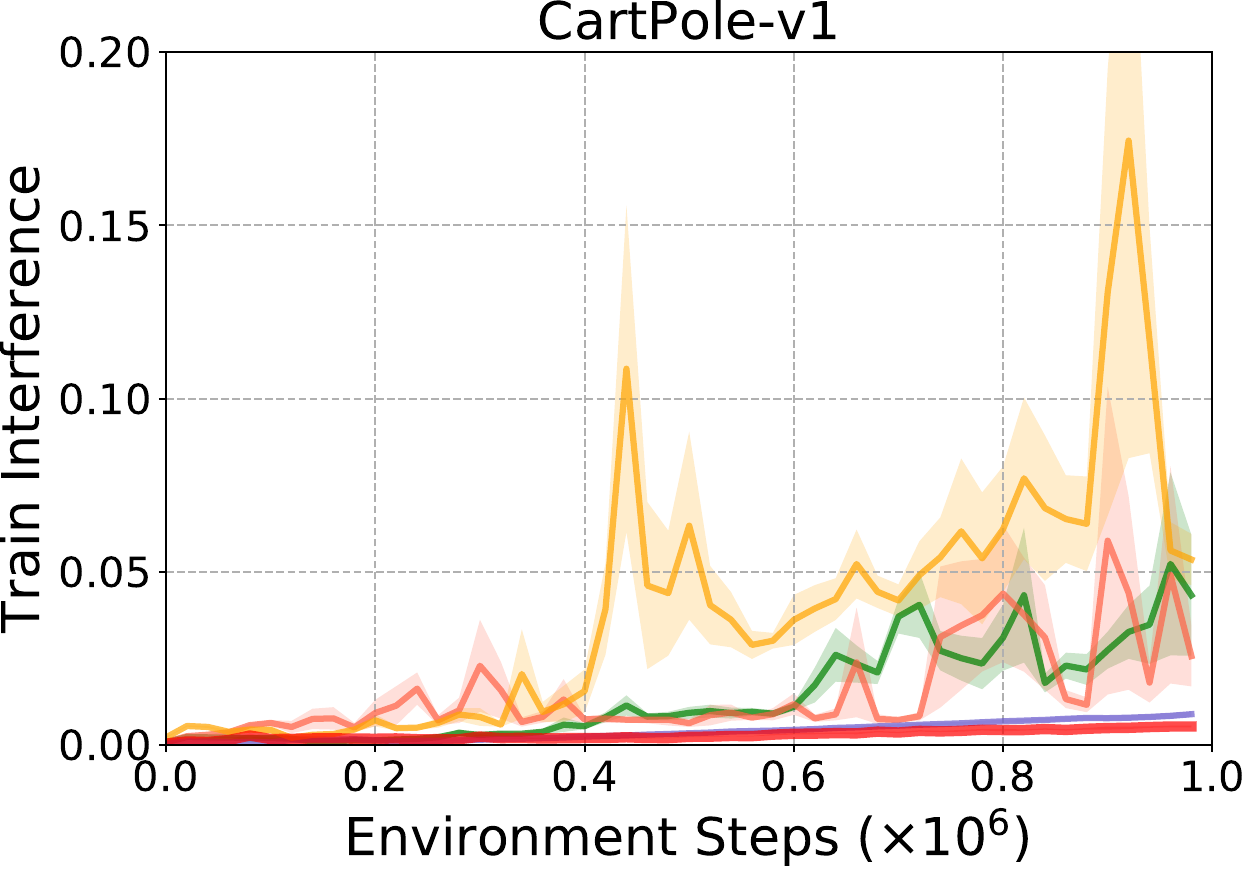}
			\includegraphics[width=0.247\textwidth]{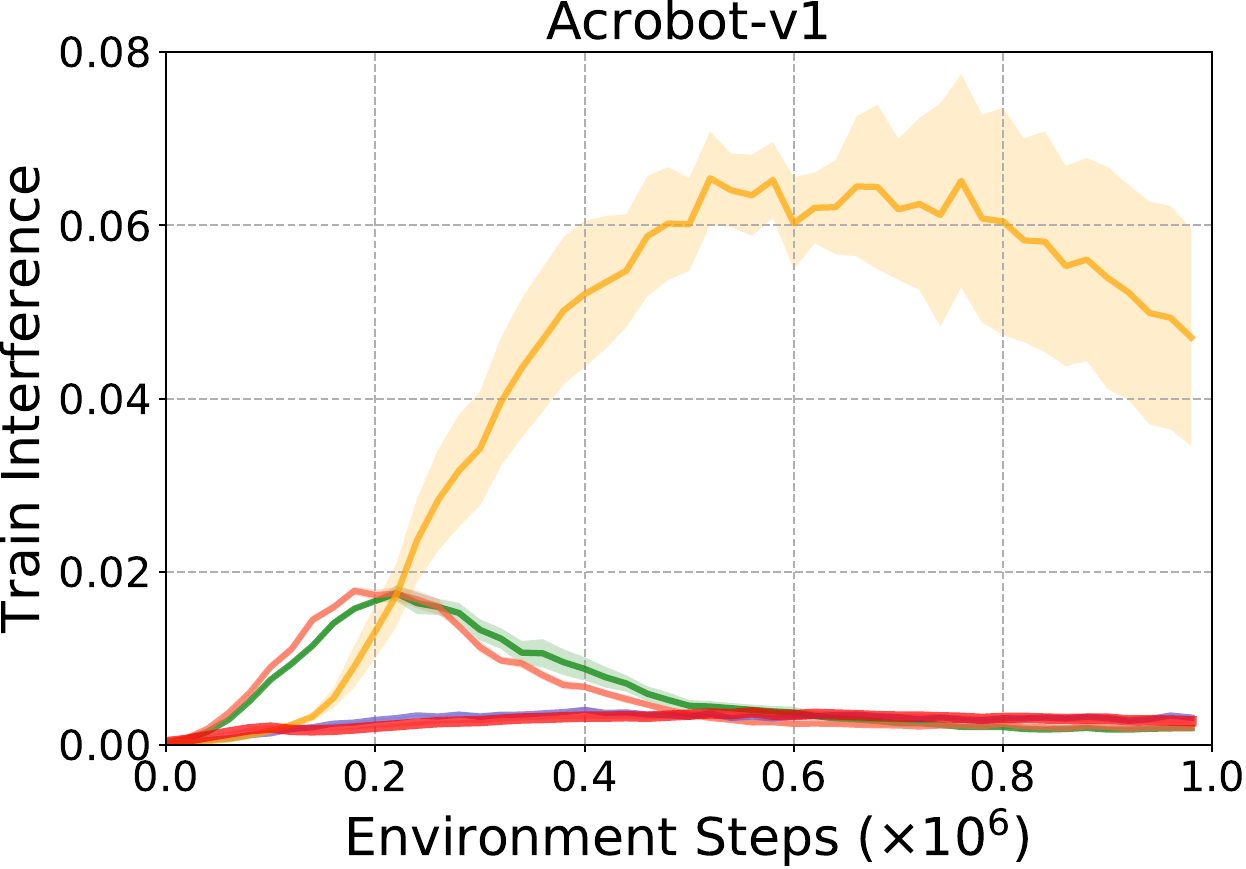}
		\end{minipage}
    }\\
    \subfigure[$N=100$]{
        \begin{minipage}[b]{0.98\textwidth}
			\includegraphics[width=0.24\textwidth]{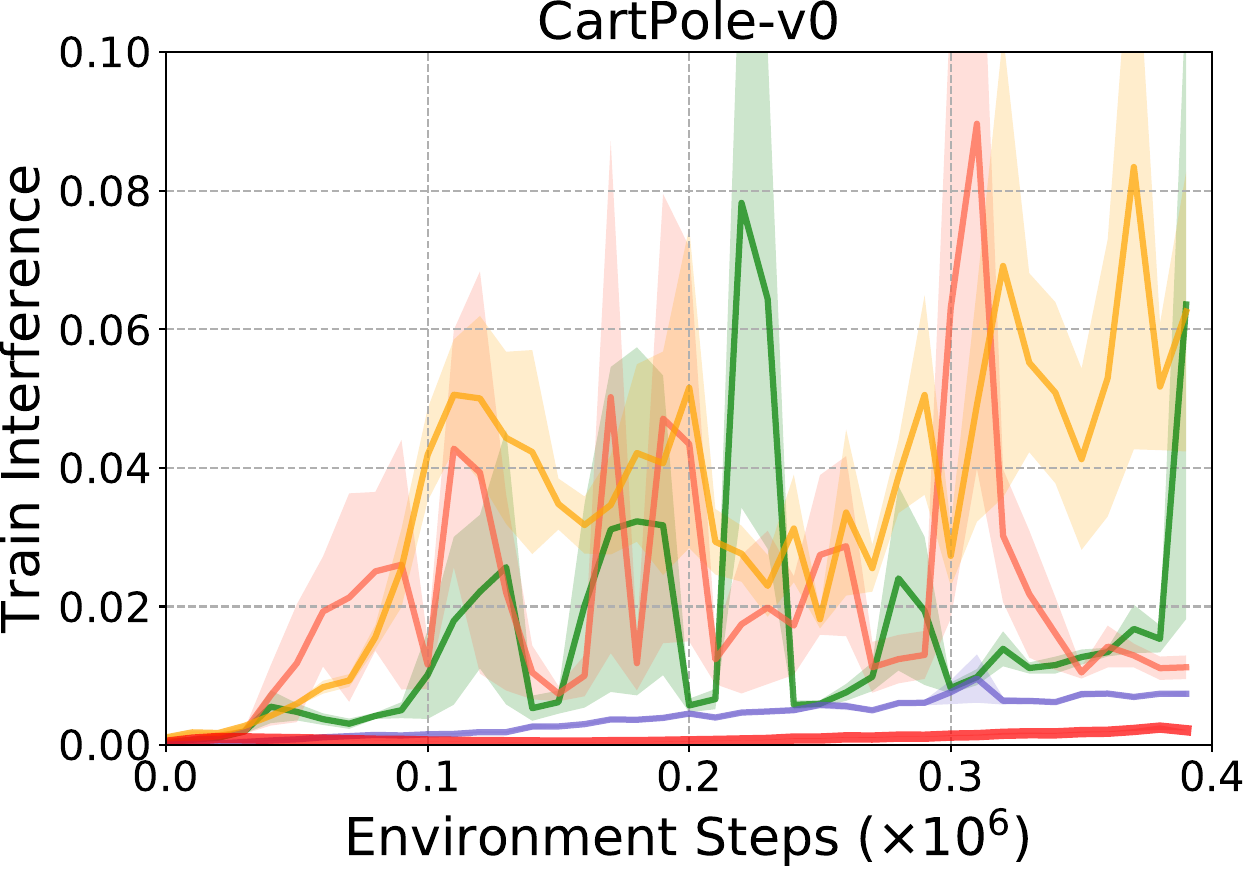}
			\includegraphics[width=0.252\textwidth]{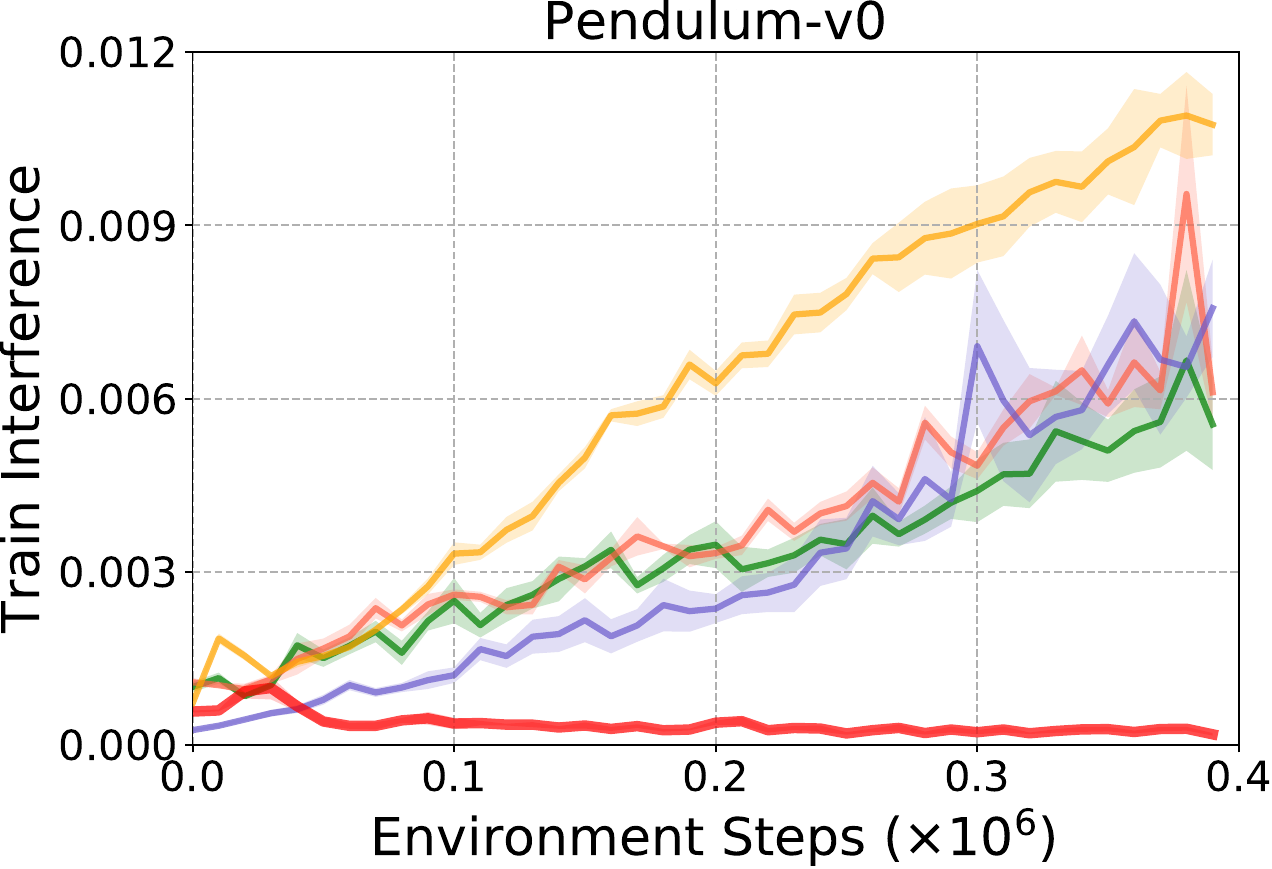}
			\includegraphics[width=0.24\textwidth]{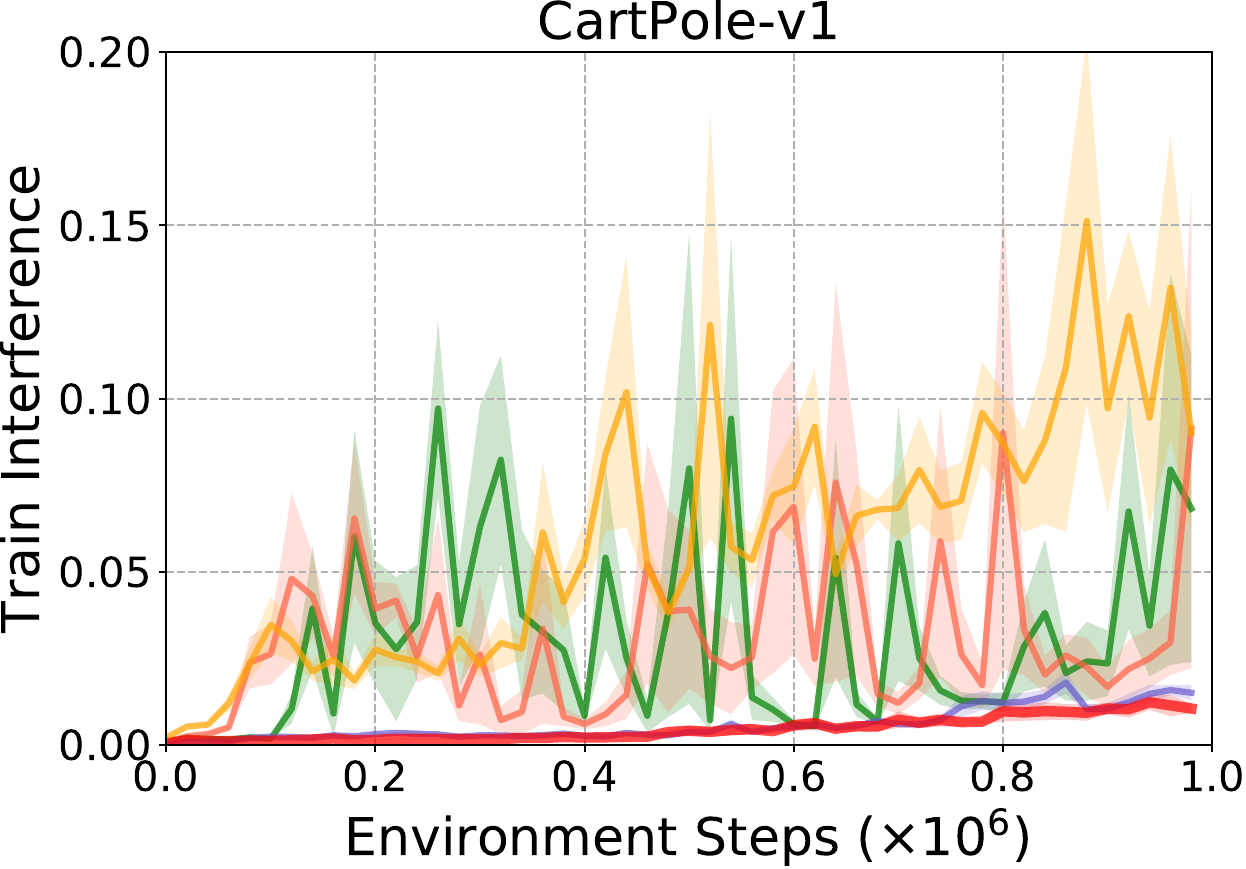}
			\includegraphics[width=0.247\textwidth]{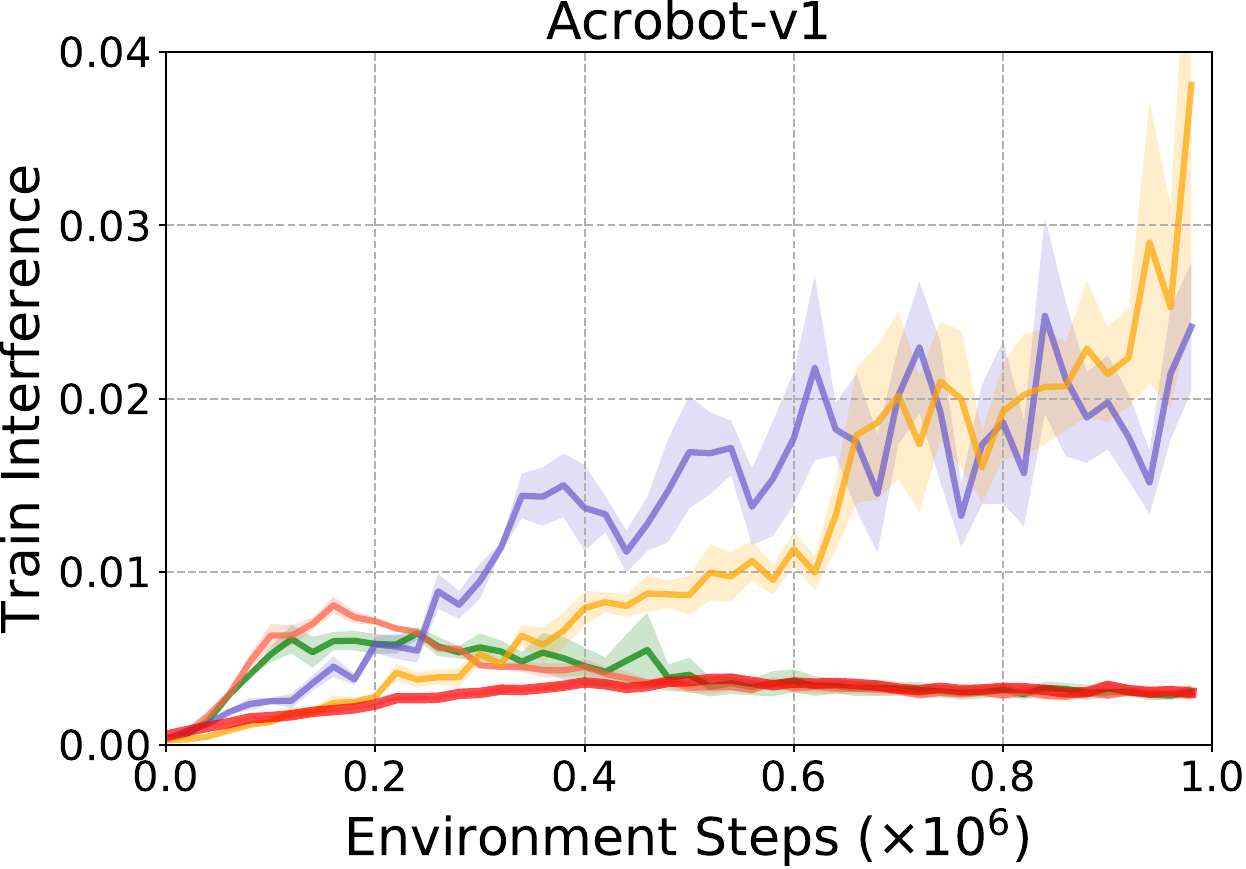}
		\end{minipage}
    }\\
    \subfigure[$N=1$]{
        \begin{minipage}[b]{0.98\textwidth}
			\includegraphics[width=0.24\textwidth]{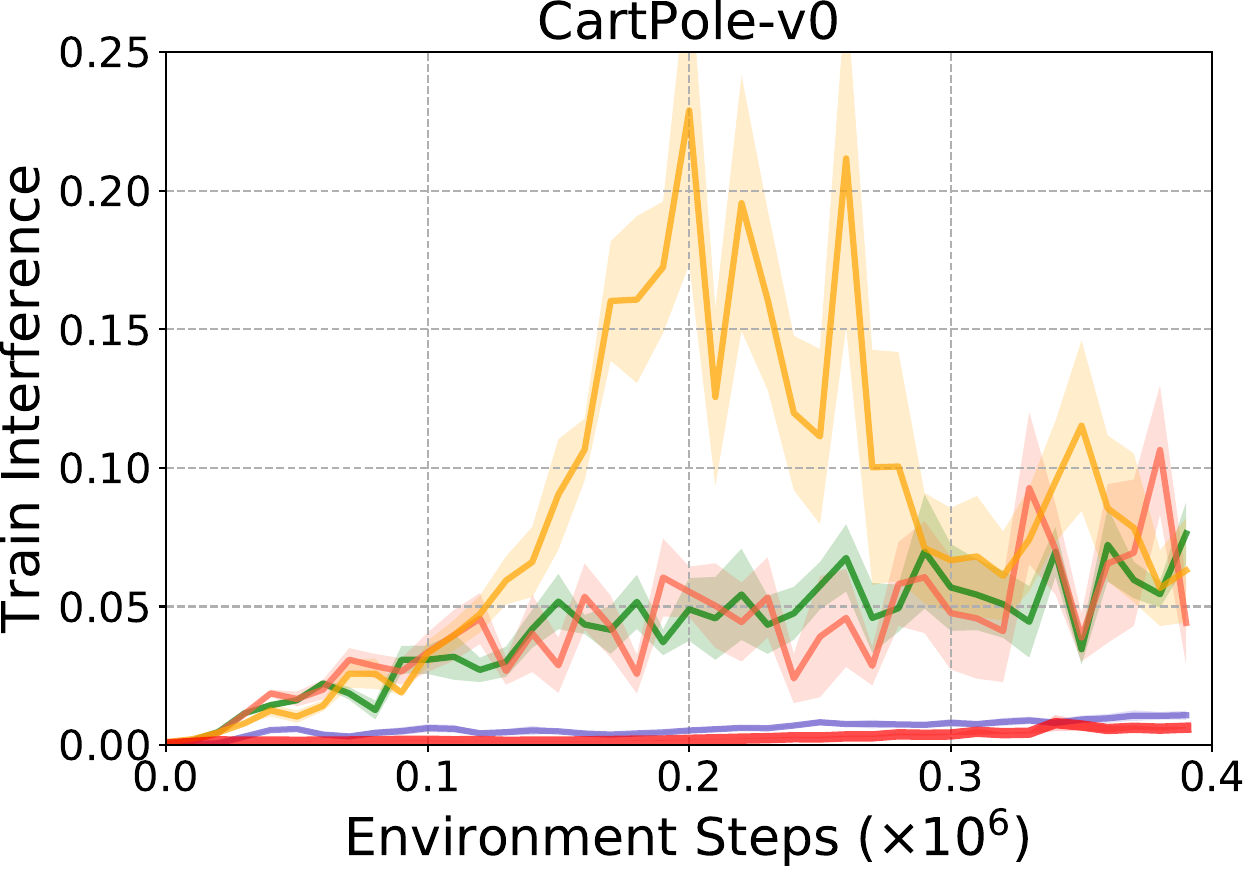}
			\includegraphics[width=0.252\textwidth]{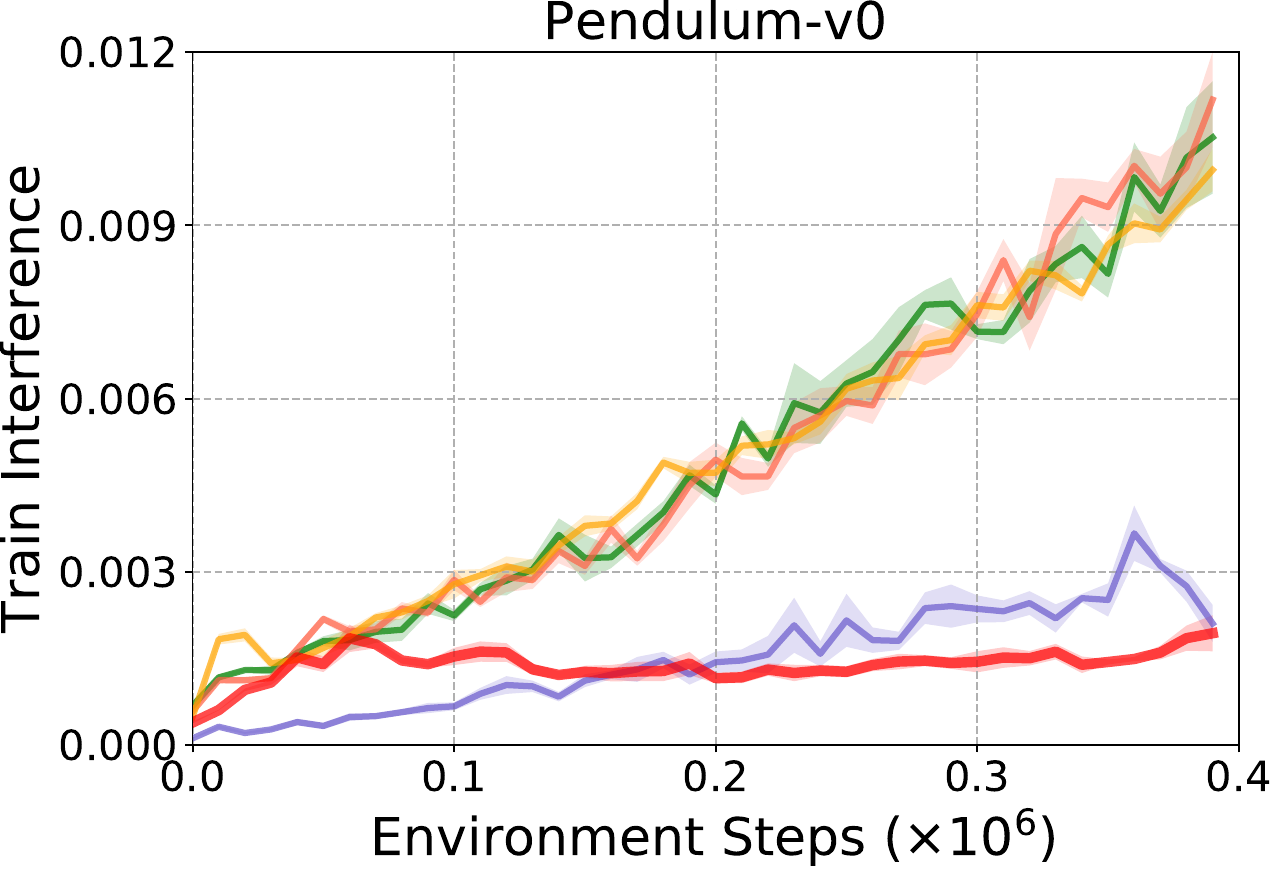}
			\includegraphics[width=0.24\textwidth]{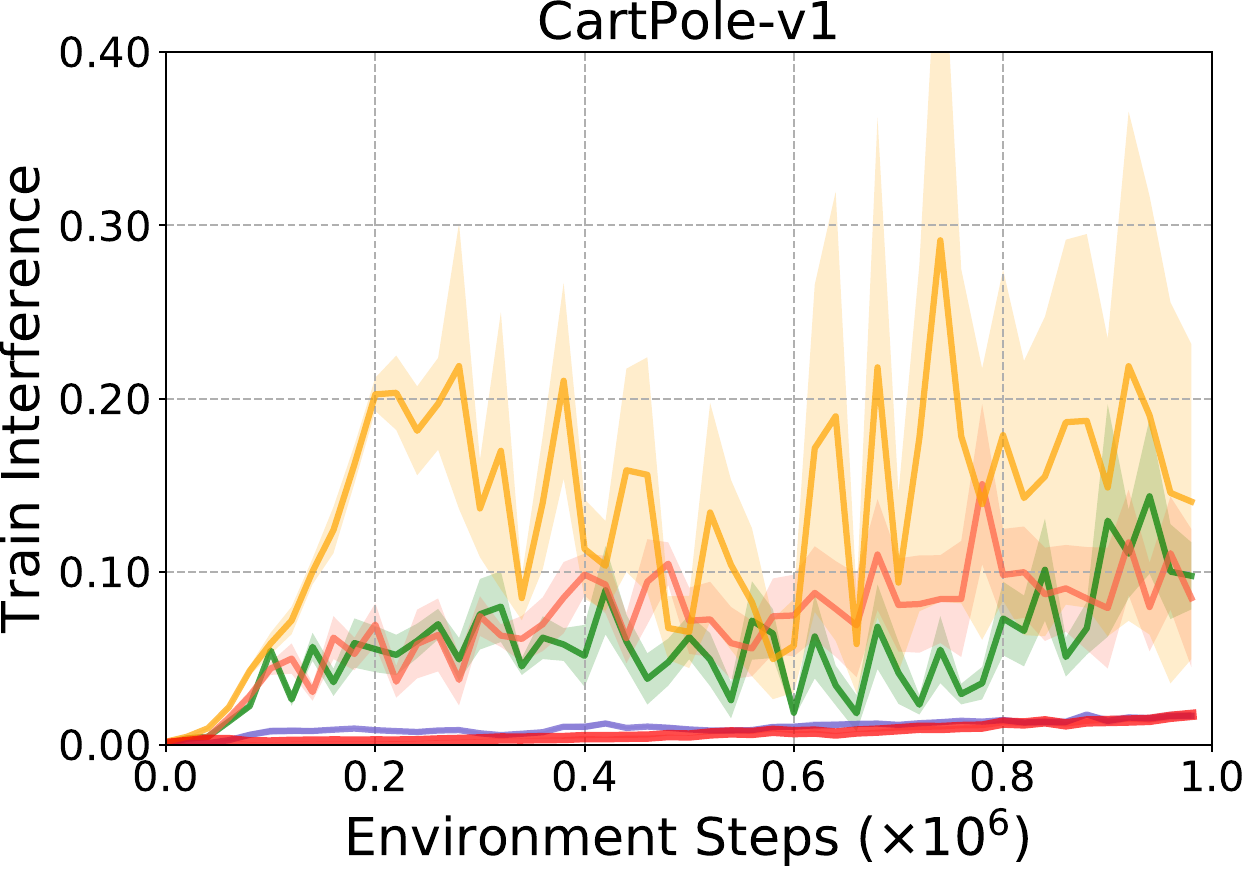}
			\includegraphics[width=0.247\textwidth]{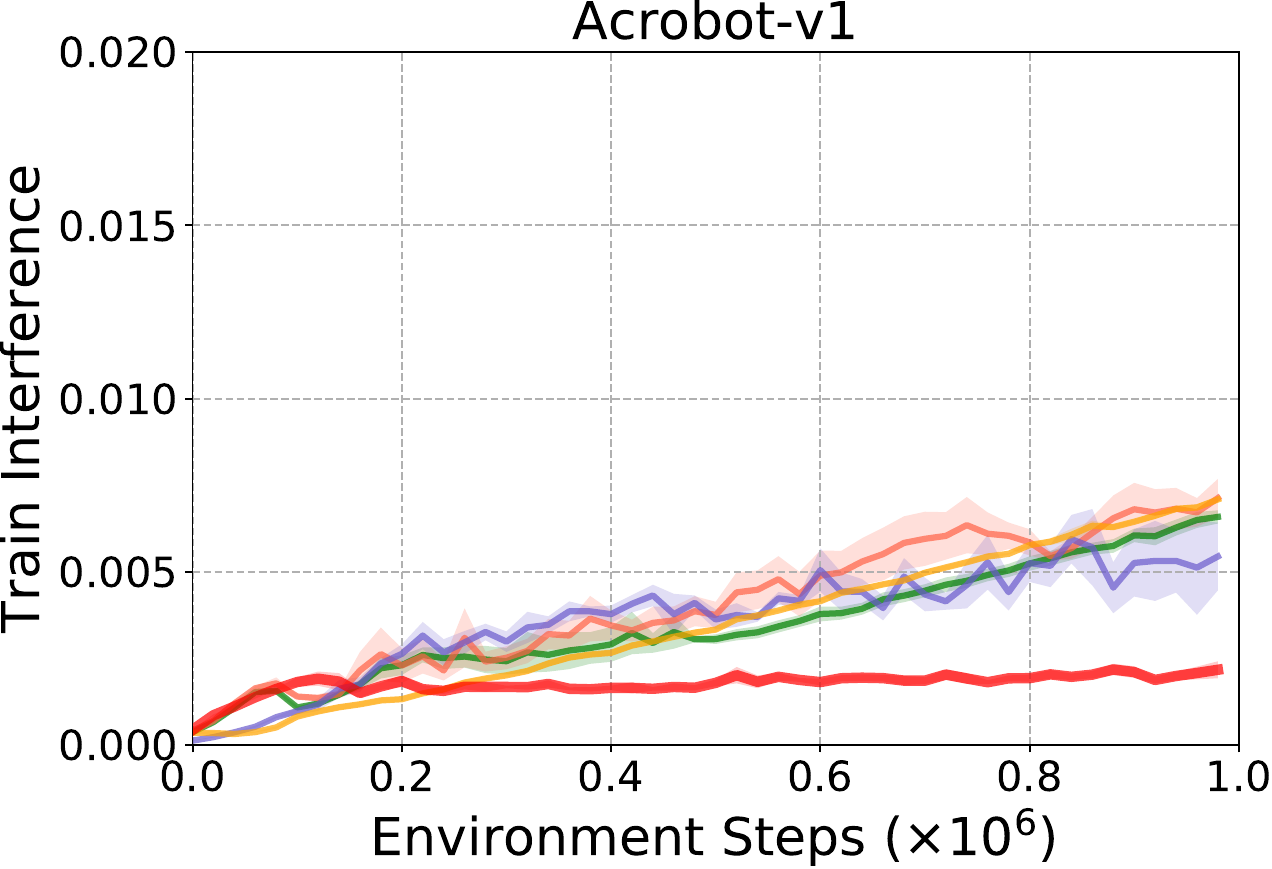}
		\end{minipage}
    }
    \caption{Interference analysis on classic control tasks with different replay buffer capacities $N$ (See Fig. \ref{fig:Classic_Control} for corresponding learning performance curves). In principle, IQ does indeed significantly alleviates the negative interference encountered by the base RL agents during learning progress.}
    \label{fig:interference_measuring}
\end{figure*}

\subsection{Measuring Interference}
\label{Measuring Interference}
We show a comparison in terms of {\em Approximate Expected Interference} (AEI, referring to Eq. \eqref{AEI}) in this section. Fig. \ref{fig:interference_measuring} records the negative interference generated after each iteration during the training process of each agent, where all curves correspond to the learning curves in Fig. \ref{fig:Classic_Control}, respectively. In our experiments, we approximate $\hat{d}$ in Eq. \eqref{AEI} with a buffer containing recent transitions of capacity 10K, keeping the settings being same as that in Fig. \ref{fig:Interference_Performance}. It should be noted that, to demonstrate the interference caused by each model iteration more clearly and intuitively, we only show the negative interference in each update step, that is, all update steps in which the losses on recent 10K transitions increase after updating the model.

From Fig. \ref{fig:interference_measuring}, we can see that the interference curves corresponding to IQ (red curves) are are below those corresponding to all other baselines in almost all task settings. Therefore, we can further confirm that our proposed scheme IQ does indeed dramatically reduce the catastrophic interference commonly encountered by the existing basic RL methods, and outperform to other baselines for interference mitigation during learning in the single RL task settings.

\subsection{Additional Experimental Results on MountainCar}
\label{mountaincar}
\begin{figure}[!t]
    \centering
    \setlength{\abovecaptionskip}{5pt}
    \subfigure[$N=50,000$]
        {\includegraphics[width=0.49\linewidth]{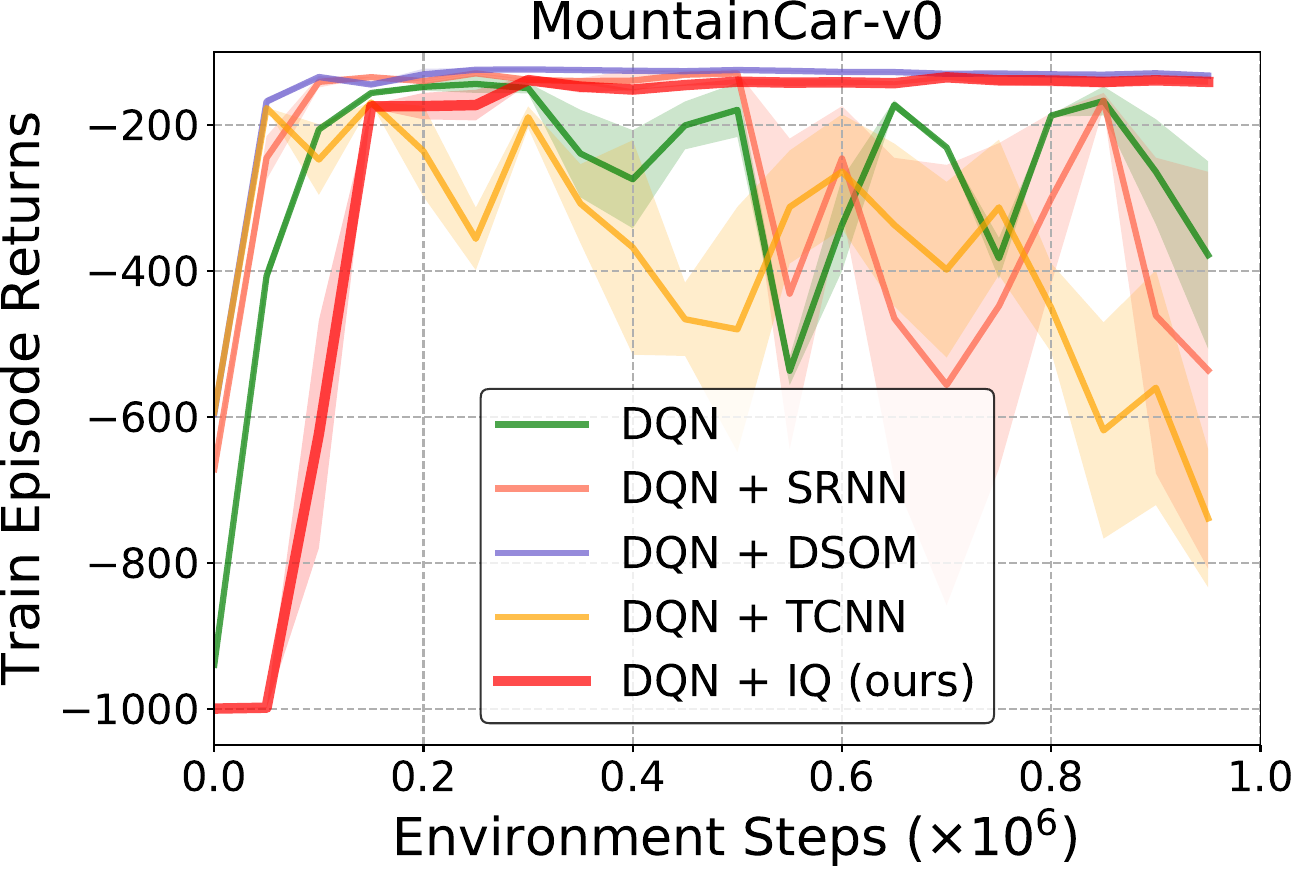}}
    \subfigure[$N=100$]
        {\includegraphics[width=0.49\linewidth]{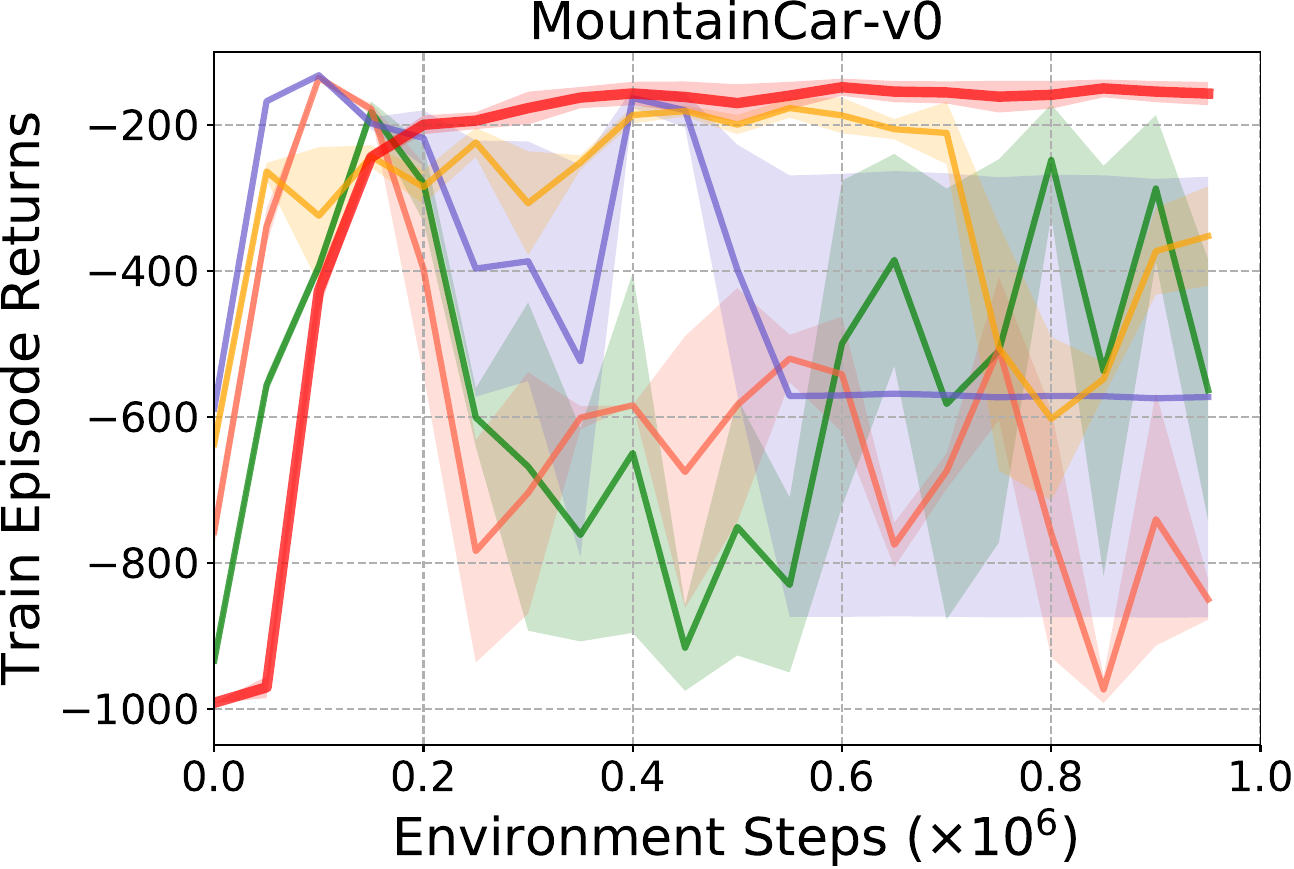}}
    \caption{Learning curves on {\em MountainCar} with different replay buffer sizes.}
    \label{fig:mountaincar}
\end{figure}

We conduct additional performance evaluation experiments on {\em MountainCar} with sparse reward. The reward threshold used to determine the success of {\em MountainCar} is $-110.0$. In our experiments, the episode terminates if it takes more than 1000 steps. All implementation details, including network architecture and parameter settings, are exactly the same as those in {\em CartPole-v1} and {\em Acrobot-v1}, except additional reward shaping. We reshape the reward for each time step based on potentials by referring to the solution\footnote{https://github.com/Pechckin/MountainCar/blob/master/MountainCar-v0.py.} on the task leadboard\footnote{https://github.com/openai/gym/wiki/Leaderboard.}. In addition, we add an extra bonus of 100 when the car reaches the target position. The results in Fig. \ref{fig:mountaincar} demonstrate the consistent superiority of IQ over the baseline methods as shown in Fig. \ref{fig:Classic_Control}.

\begin{table*}[t]
\centering
\setlength{\tabcolsep}{0.8mm}
\caption{$P$-values of the AUC of the learning curves in Fig. \ref{fig:Atari_games} (The settings where the  difference between\\ the AUC of the learning curves is significant are marked in boldface, which means\\ that the significance test rejects the null hypothesis that the area\\ under the learning curves come from normal distribution\\ with equal means at the $1\%$ level)}
\label{table:ks-test}
\begin{tabular}{c|cccc|cccc}
\toprule
Base RL                      & \multicolumn{4}{c|}{DQN}                                                                                                           & \multicolumn{4}{c}{Rainbow}  \\ \hline
Method                       & \multicolumn{2}{c}{IQ-RE vs DQN}                               & \multicolumn{2}{c|}{IQ-RE vs SRNN}                                & \multicolumn{2}{c}{IQ-RE vs Rainbow}                                & \multicolumn{2}{c}{IQ-RE vs SRNN}      \\ \hline
N                            & 1,000,000                   & 10,000                           & 1,000,000                     & 10,000                            & 1,000,000                      & 10,000                             & 1,000,000                     & 10,000              \\ \hline
\textit{Pong}                & \bm{$1.62\times10^{-06}$}   & \bm{$1.32\times10^{-07}$}        & \bm{$4.09\times10^{-04}$}     & \bm{$1.91\times10^{-04}$}         & $8.69\times10^{-02}$           & \bm{$5.90\times10^{-04}$}          & \bm{$5.75\times10^{-05}$}     & \bm{$1.62\times10^{-04}$}     \\
\textit{Breakout}            & $1.21\times10^{-01}$        & \bm{$1.37\times10^{-12}$}        & $2.19\times10^{-01}$          & \bm{$6.03\times10^{-04}$}         & \bm{$6.61\times10^{-18}$}      & \bm{$3.96\times10^{-05}$}          & \bm{$5.19\times10^{-08}$}     & \bm{$4.87\times10^{-05}$}     \\
\textit{Carnival}            & $2.00\times10^{-01}$        & \bm{$1.62\times10^{-06}$}        & $2.56\times10^{-02}$          & \bm{$8.45\times10^{-04}$}         & \bm{$1.56\times10^{-20}$}      & \bm{$2.22\times10^{-07}$}          & \bm{$2.25\times10^{-19}$}     & \bm{$7.96\times10^{-28}$}     \\
\textit{Freeway}             & \bm{$2.60\times10^{-05}$}   & $1.00\times10^{-00}$             & \bm{$1.02\times10^{-11}$}     & $1.00\times10^{-00}$              & $2.00\times10^{-01}$           & \bm{$2.60\times10^{-06}$}          & $1.34\times10^{-01}$          & \bm{$\ast1.20\times10^{-03}$}      \\
\textit{Tennis}              & $2.00\times10^{-01}$        & \bm{$2.48\times10^{-05}$}        & $7.60\times10^{-01}$          & \bm{$\ast7.14\times10^{-11}$}     & $2.00\times10^{-01}$           & \bm{$2.03\times10^{-03}$}          & $4.01\times10^{-01}$          & \bm{$2.80\times10^{-06}$}     \\
\textit{FishingDerby}        & $4.67\times10^{-01}$        & $8.69\times10^{-02}$             & $6.91\times10^{-02}$          & $2.88\times10^{-01}$              & $9.69\times10^{-01}$           & \bm{$8.30\times10^{-03}$}          & \bm{$7.96\times10^{-14}$}     & \bm{$6.09\times10^{-07}$}     \\
\bottomrule
\end{tabular}
\end{table*}

\begin{table*}[!t]
\centering
\setlength{\tabcolsep}{1.5mm}
\caption{Statistics of the maximum deterioration ratios suffered during training on\\ Atari games (based on the performance of five runs in Fig. \ref{fig:Atari_games}, and\\ the best performance is marked in boldface.)}
\label{table:maximum_deterioration_ratio}
\begin{tabular}{c|cccccc|cccccc}
\toprule
\specialrule{0em}{1pt}{1pt}
Method                & \multicolumn{2}{c}{DQN}            & \multicolumn{2}{c}{DQN + SRNN}      & \multicolumn{2}{c|}{DQN + IQ-RE (ours)}     & \multicolumn{2}{c}{Rainbow}        & \multicolumn{2}{c}{Rainbow + SRNN}  & \multicolumn{2}{c}{Rainbow + IQ-RE (ours)}     \\ \hline
N                     & 10,000           & 1,000,000       & 10,000           & 1,000,000        & 10,000          & 1,000,000                 & 10,000          & 1,000,000        & 10,000        & 1,000,000           & 10,000           & 1,000,000                   \\ \hline
\textit{Pong}         & $0.041$          & $0.017$         & $0.029$          & $0.022$          & $0.016$         & $0.016$                   & $0.026$         & $0.005$          & $0.014$       & $0.005$             & $0.013$          & $0.005$              \\ 
\textit{Breakout}     & $0.009$          & $0.084$         & $0.008$          & $0.111$          & $0.035$         & $0.094$                   & $0.071$         & $0.256$          & $0.057$       & $0.298$             & $0.071$          & $0.250$              \\ 
\textit{Carnival}     & $0.012$          & $0.007$         & $0.021$          & $0.004$          & $0.009$         & $0.013$                   & $0.041$         & $0.080$          & $0.019$       & $0.019$             & $0.045$          & $0.076$              \\ 
\textit{Freeway}      & $1.000$          & $0.176$         & $1.000$          & $0.095$          & $1.000$         & $0.046$                   & $0.010$         & $0.002$          & $0.524$       & $0.140$             & $0.067$          & $0.003$              \\ 
\textit{Tennis}       & $0.586$          & $0.118$         & $0.400$          & $0.106$          & $0.370$         & $0.181$                   & $0.493$         & $0.176$          & $0.527$       & $0.136$             & $0.325$          & $0.188$              \\ 
\textit{FishingDerby} & $0.096$          & $0.107$         & $0.152$          & $0.100$          & $0.113$         & $0.082$                   & $0.042$         & $0.019$          & $0.061$       & $0.018$             & $0.030$          & $0.014$              \\ 
\hline
Average               & $0.291$          & $0.085$         & $0.268$          & $0.073$          & \bm{$0.257$}    & \bm{$0.072$}      & $0.114$          & $0.090$          & $0.200$       & $0.103$             & \bm{$0.092$}     & \bm{$0.089$}         \\ 
\specialrule{0em}{1pt}{1pt}
\bottomrule
\end{tabular}
\end{table*}

\begin{figure*}[t]
    \centering
    \setlength{\abovecaptionskip}{7pt}
        \begin{minipage}[b]{0.98\textwidth}
            \vspace{2mm}
			\includegraphics[width=0.24\textwidth]{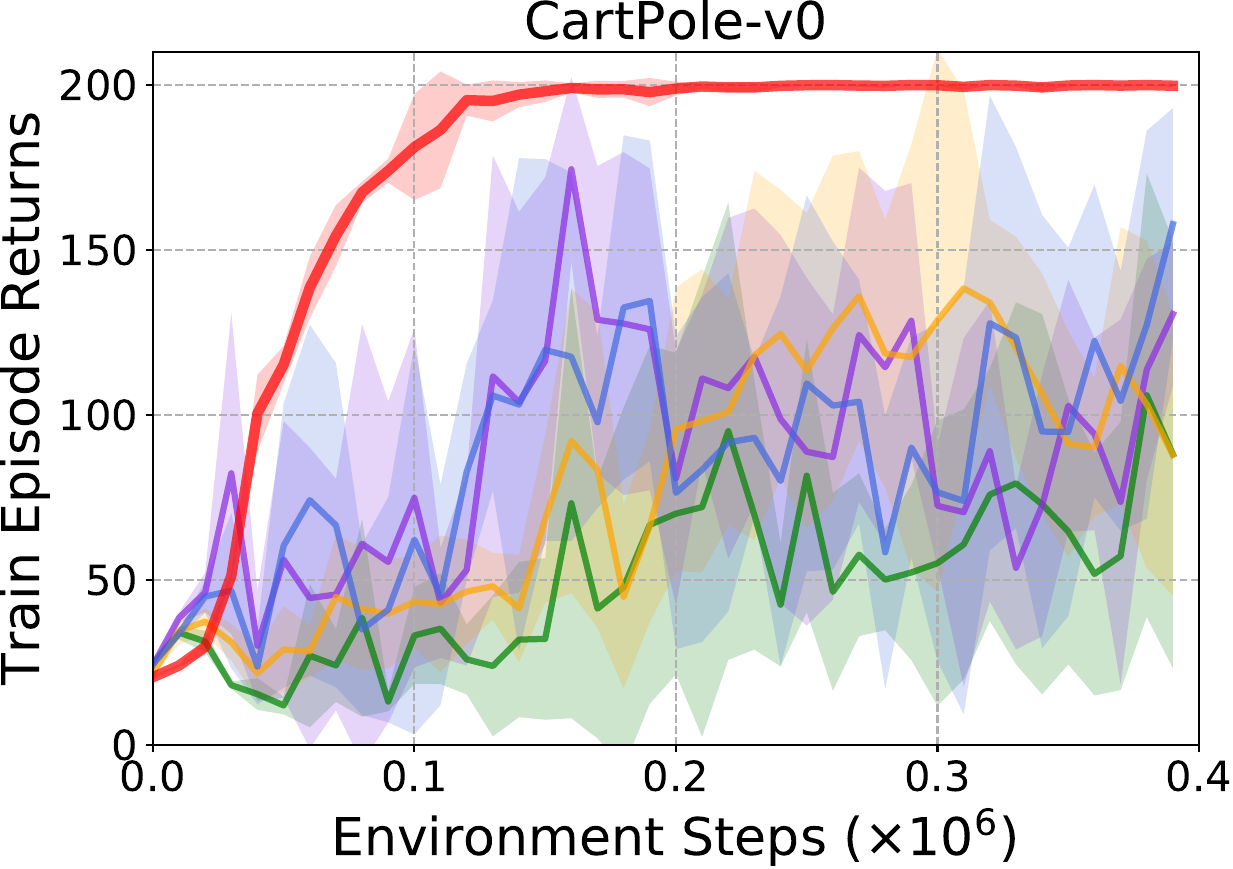}
			\includegraphics[width=0.252\textwidth]{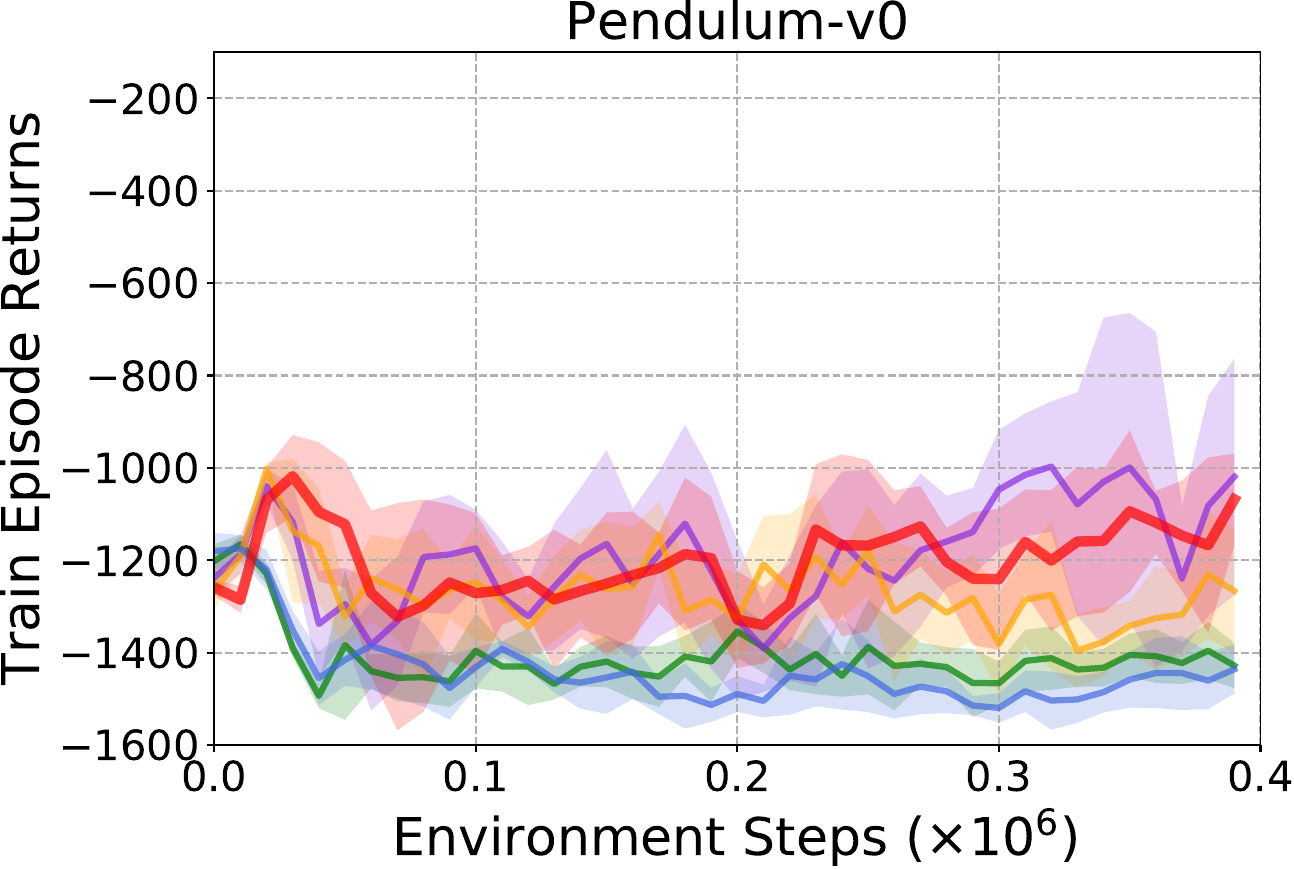}
			\includegraphics[width=0.24\textwidth]{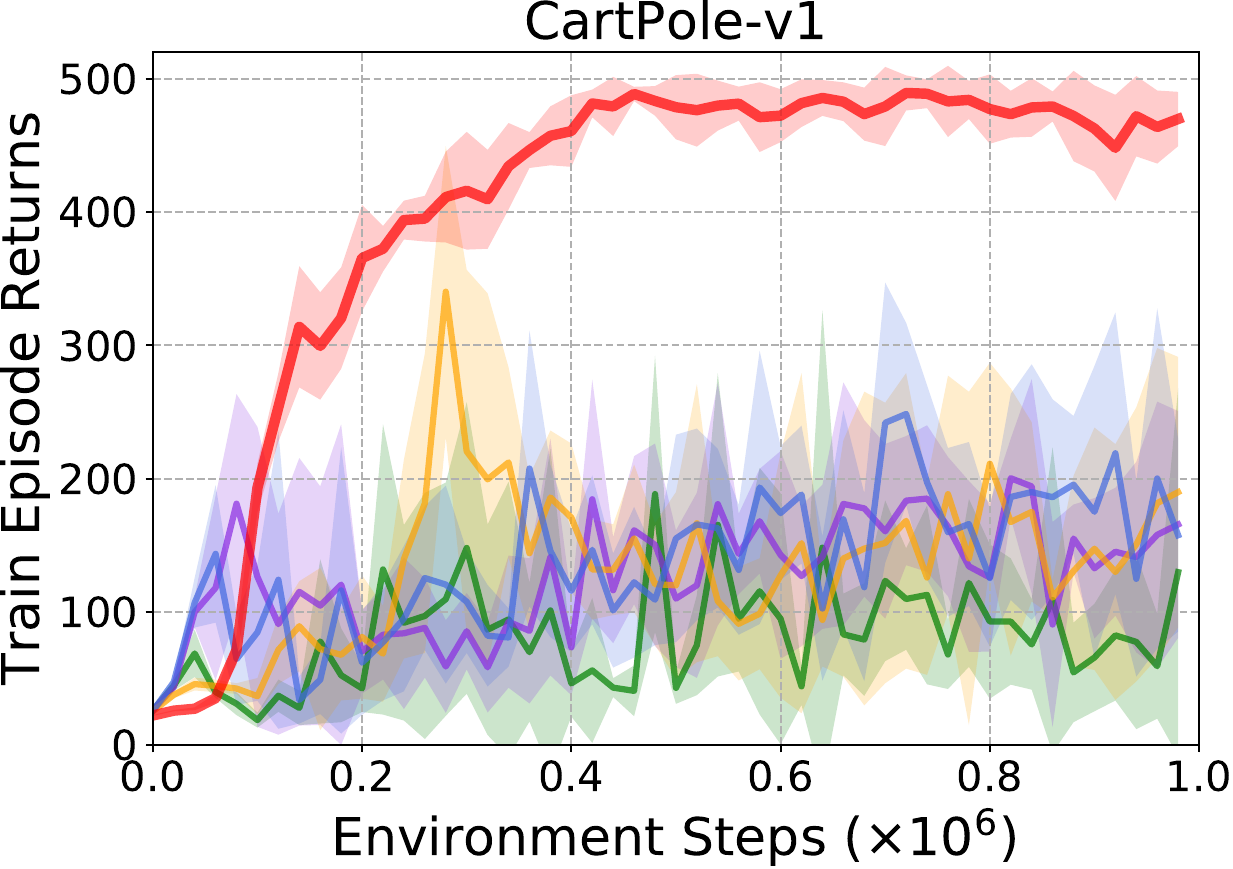}
			\includegraphics[width=0.247\textwidth]{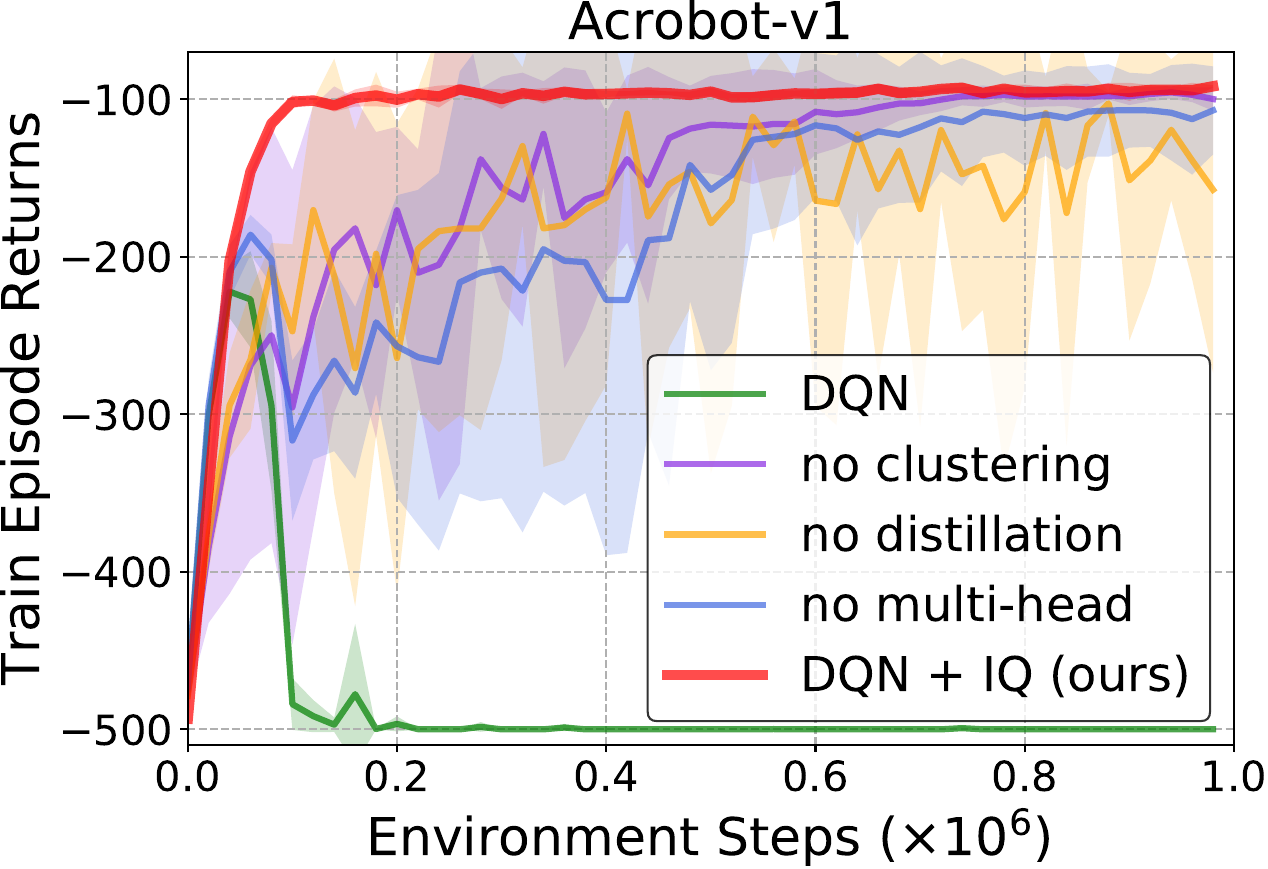}
		\end{minipage}
    \caption{Additional experimental results of IQ incorporated with DQN and three different ablations, on each individual task ($N=1$).}
    \label{fig:add_ablation_study}
\end{figure*}

\begin{figure*}[t]
    \centering
    \setlength{\abovecaptionskip}{7pt}
    \subfigure[$N=50,000$]{
        \begin{minipage}[b]{0.98\textwidth}
			\includegraphics[width=0.24\textwidth]{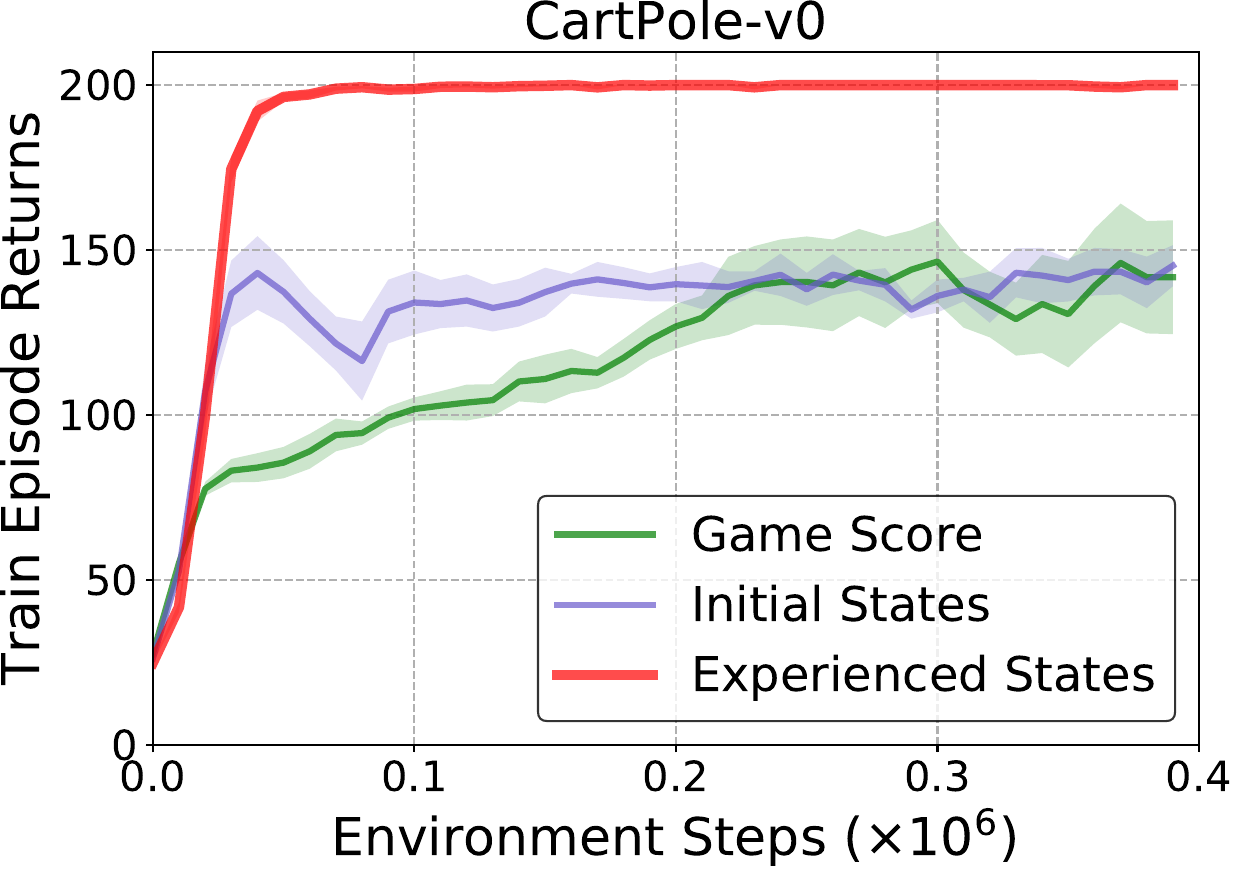}
			\includegraphics[width=0.252\textwidth]{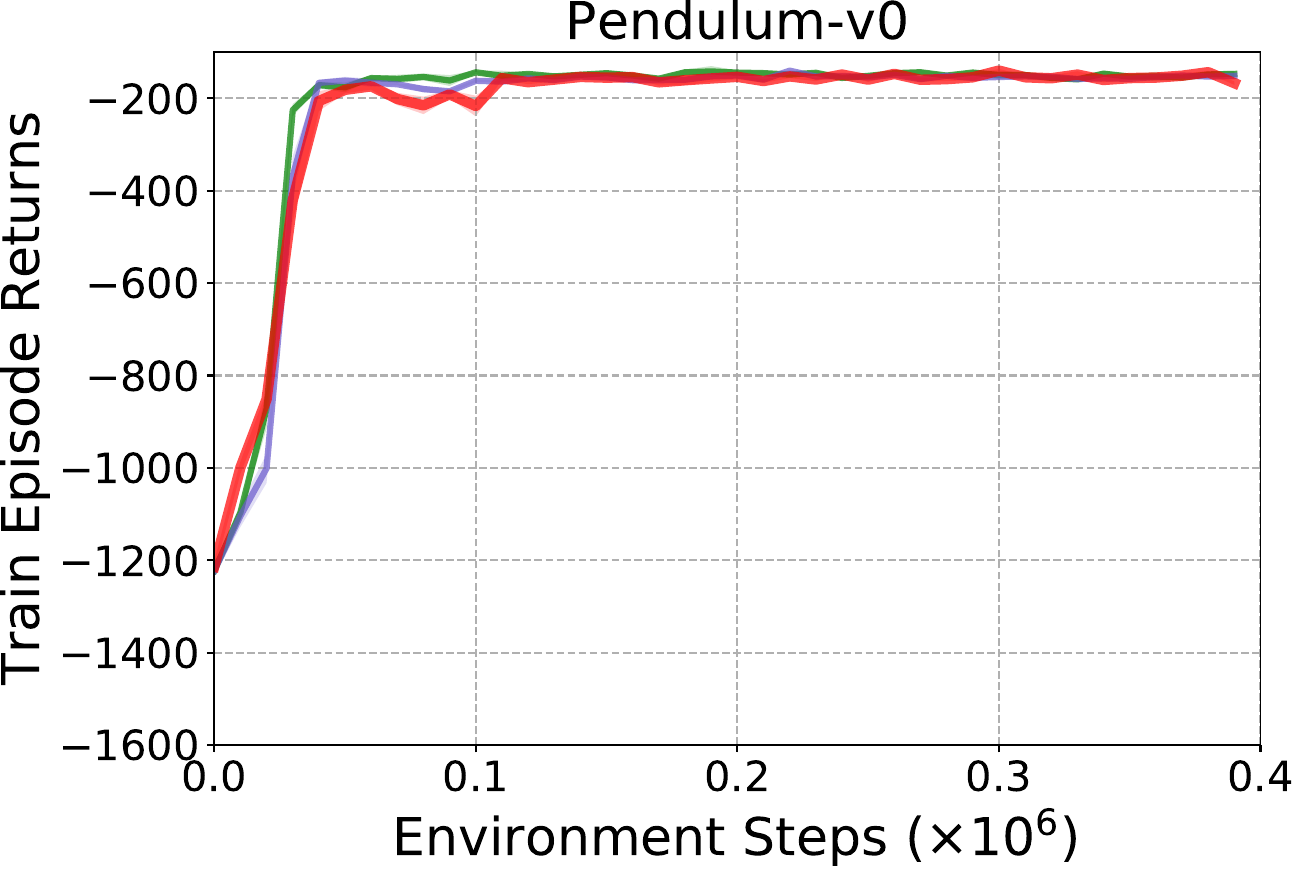}
			\includegraphics[width=0.24\textwidth]{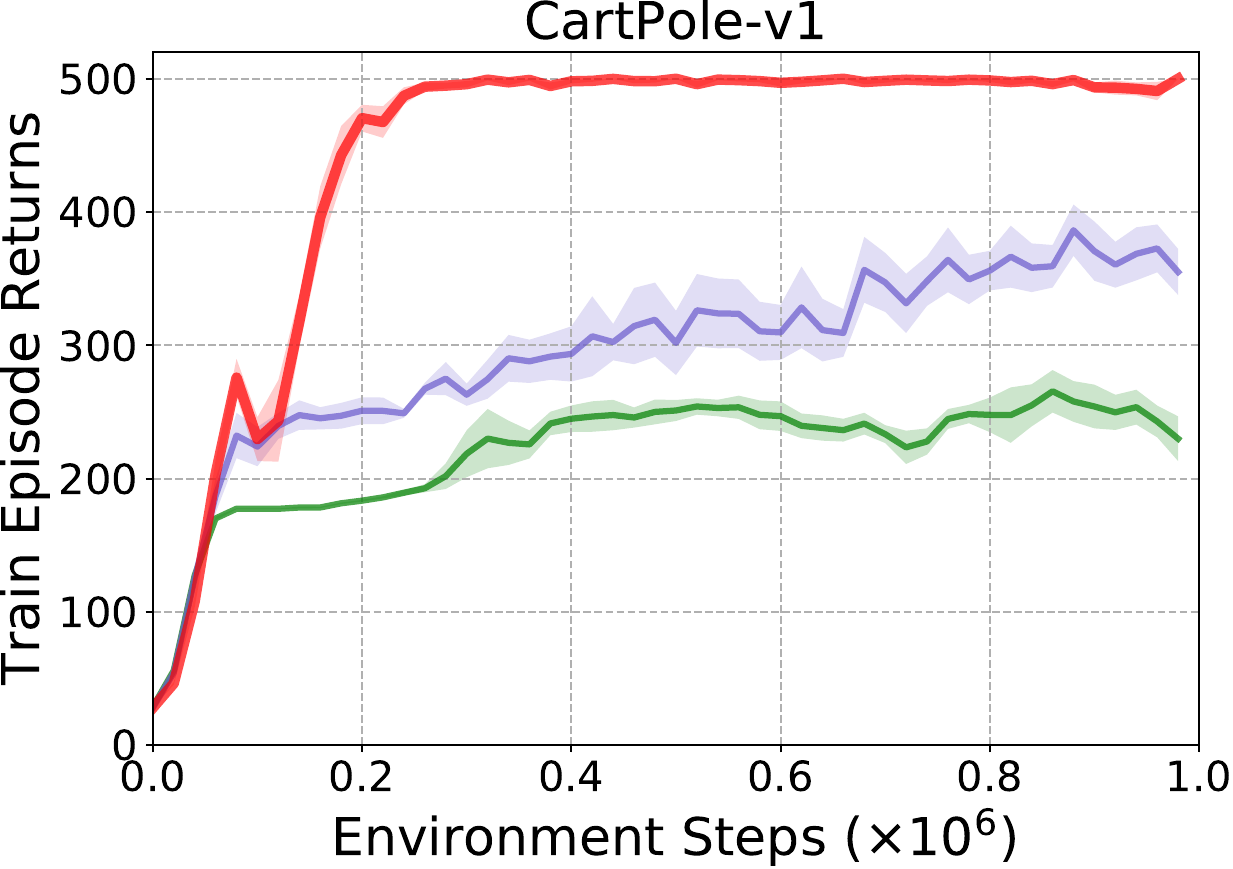}
			\includegraphics[width=0.247\textwidth]{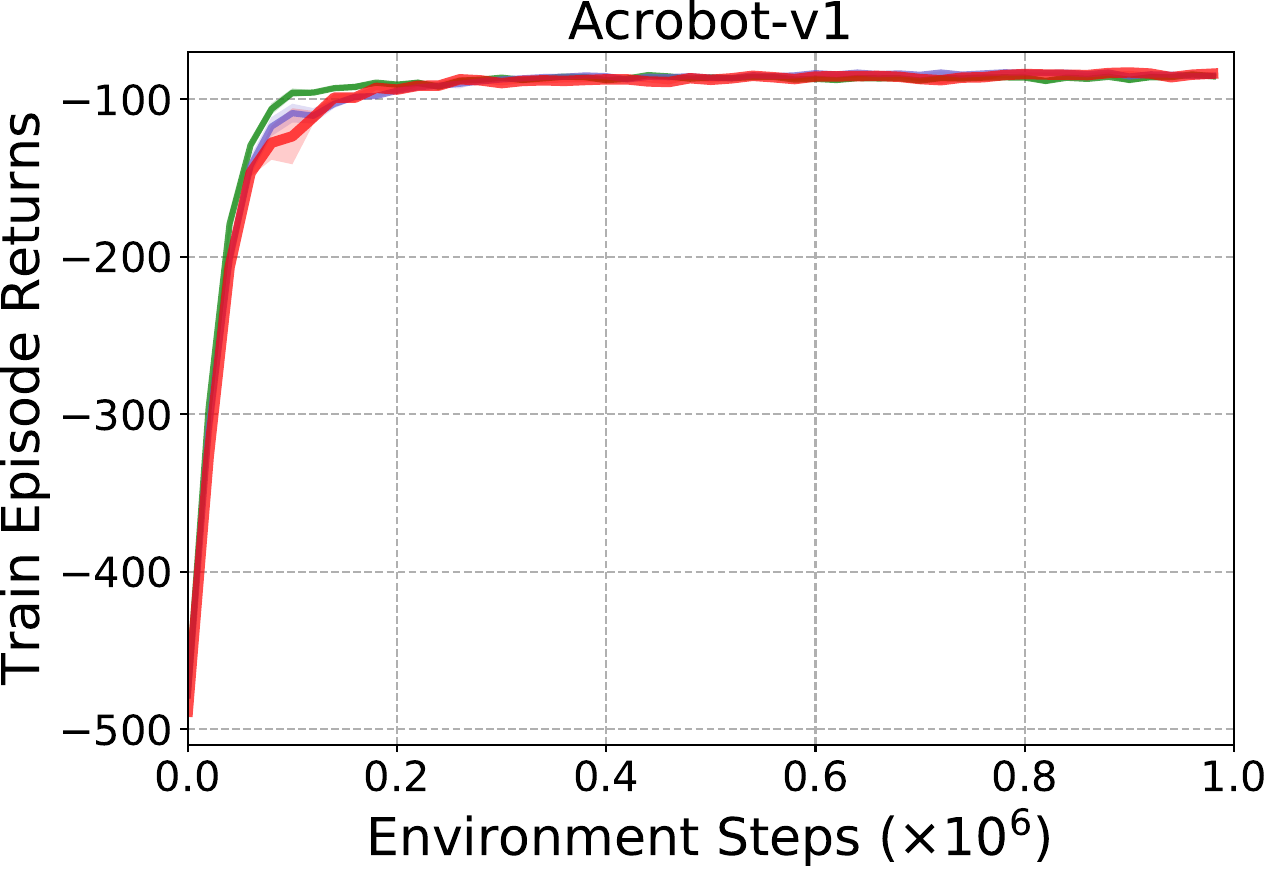}
		\end{minipage}
    }\\
    \subfigure[$N=1$]{
        \begin{minipage}[b]{0.98\textwidth}
			\includegraphics[width=0.24\textwidth]{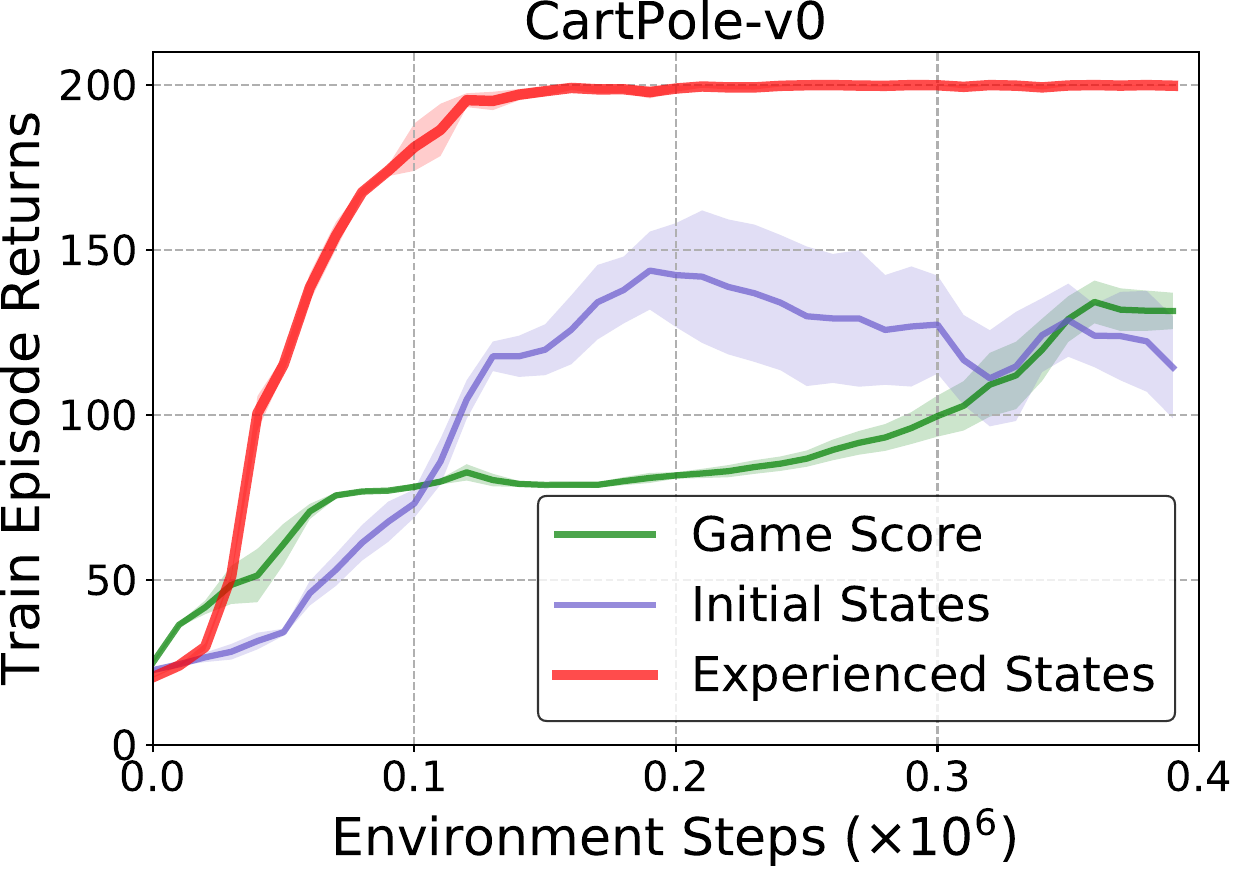}
			\includegraphics[width=0.252\textwidth]{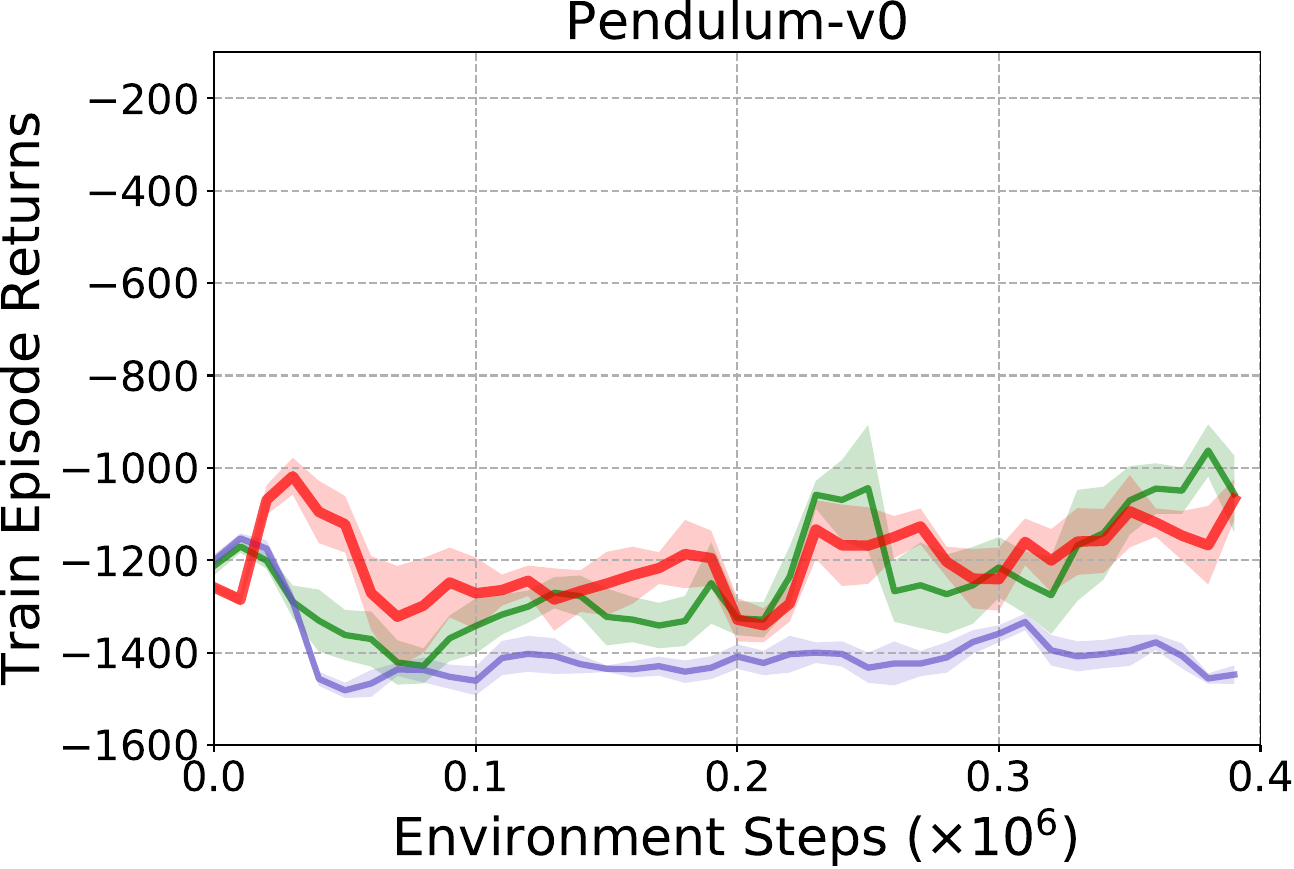}
			\includegraphics[width=0.24\textwidth]{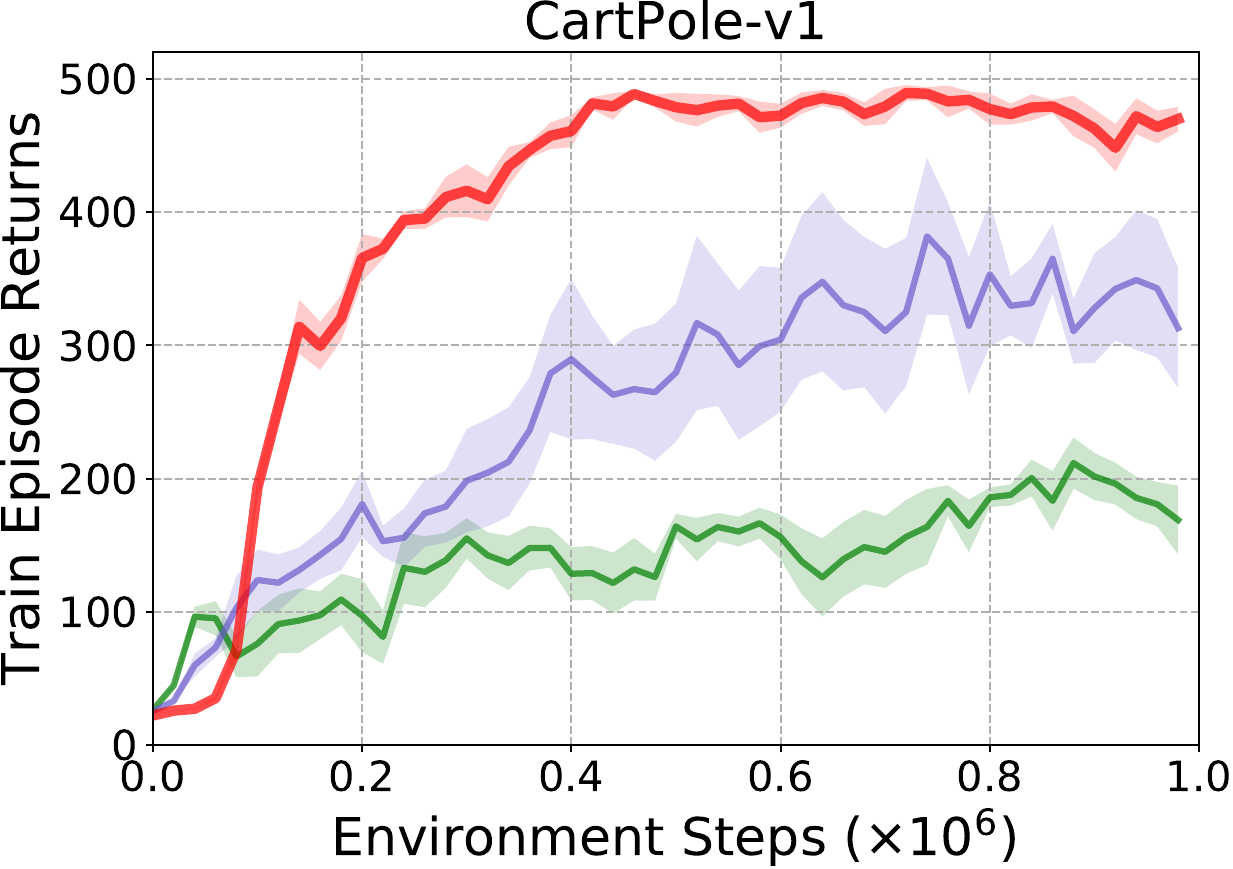}
			\includegraphics[width=0.247\textwidth]{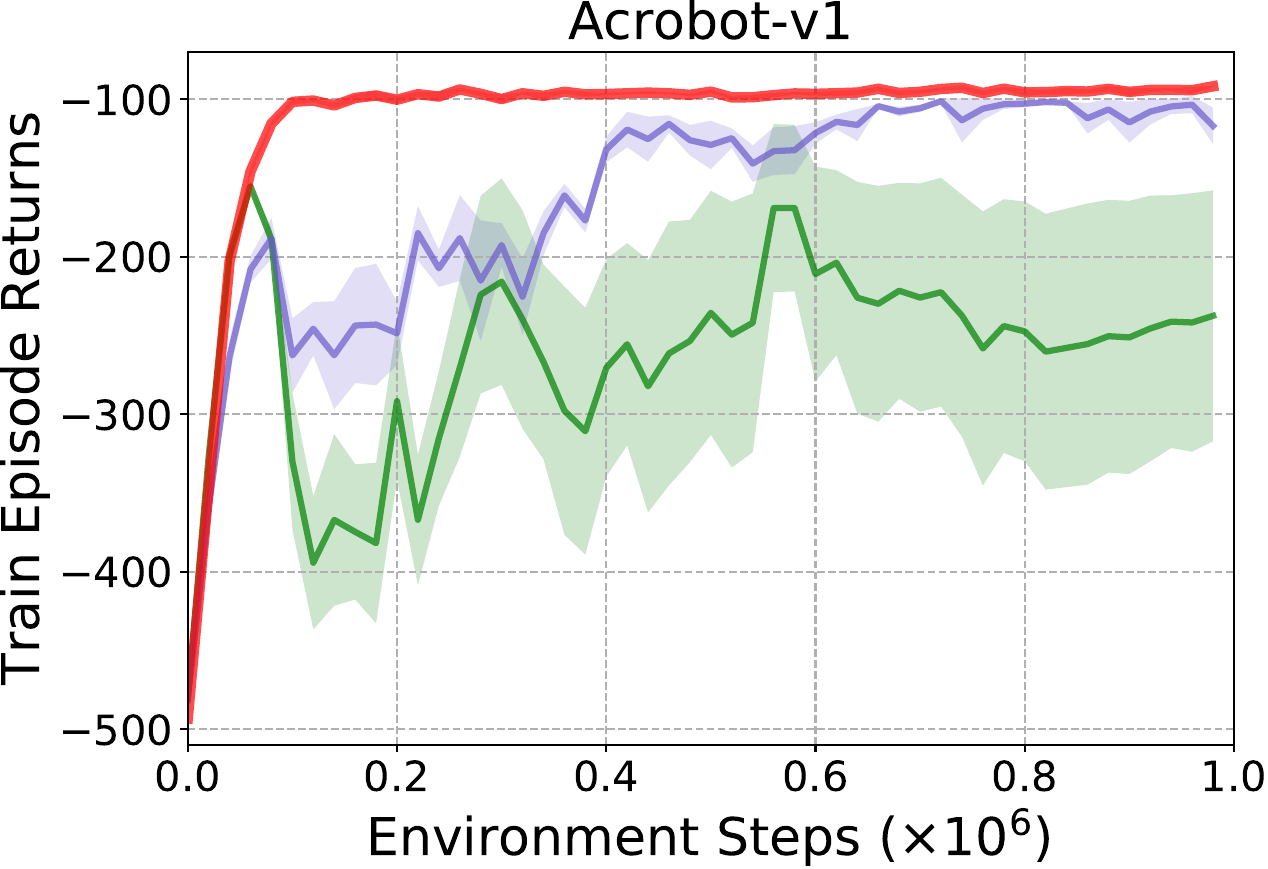}
		\end{minipage}
    }
    \caption{Additional experimental results of comparisons of IQ (red, performs context division on all experienced states) with other two context division strategies (other colors) with $N=50,000$ and $N=1$.}
    \label{fig:Add_cd_strategy}
\end{figure*}

\subsection{Two-Sample Kolmogorov-Smirnov Test Results for Fig. \ref{fig:Atari_games}}
\label{ks_test}
We performed a two-sample Kolmogorov-Smirnov test on the area under the learning curves reported in Fig. \ref{fig:Atari_games}. Table \ref{table:ks-test} summarizes the $p$-values of the test, where each cell in the table reports the $p$-value from the two-sample Kolmogorov-Smirnov test comparing the 5 AUC (one for each run) values of the corresponding learning curve in Fig. \ref{fig:Atari_games}. At the $1\%$ significant level, we mark the values with significant difference in boldface in which the two values with asterisk indicate that IQ-RE is worse than SRNN, while in all other cases IQ-RE produces statistically significant improvement over corresponding baselines. The values not in boldface represent the cases where the performance of IQ-RE is comparable to baselines.

\subsection{Statistics of the Maximum Deterioration Ratios}
In order to further investigate the maximum degradation degree of the performance during model training, the maximum deterioration ratios compared to the previous maximal episode returns are given in Table \ref{table:maximum_deterioration_ratio}. From which, we can see that IQ-RE achieves less performance degradation in most tasks compared to its counterparts. Under a small buffer setting ($N=10,000$), DQN and Rainbow incorporated with IQ-RE surpass the original RL methods by $3.4\%$ and $2.2\%$, respectively, and exceed that combined with SRNN by $1.1\%$ and $10.8\%$, respectively. Note that, even with a large memory ($N=1,000,000$), IQ-RE still shows certain advantages over the baselines.

\subsection{Additional Results of Ablation Experiments}
\label{Additional results of ablation experiments}
We provide additional experimental results of IQ incorporated with DQN and three different ablations on four classic control tasks with $N=1$ in Fig. \ref{fig:add_ablation_study}. Overall, all three ablations do improve the performance of DQN, but are still not comparable to IQ that combines all components. Those results are consistent with the analysis about ablation study in Section \ref{Analysis}.

\subsection{Additional Results of Context Division Strategy}
\label{Additional Results of Context Division Strategy}
We show additional experimental results of the comparison to other two context division strategies ({\em i.e.}, context division based on game score or initial state space) under $N=50,000$ and $N=1$ settings in Fig. \ref{fig:Add_cd_strategy}, which further confirm the effectiveness of our proposed context division strategy based on all experienced states during learning.

\section{Calculation of model complexity}    
\textbf{Computational Complexity of Context Division}. At each time step, {\em SKM} only needs to calculate the distances between the current state and $k$ context centroids. Given a $d$-dimensional state space and $T$-step environment interactions, the time complexity of context division is $\mathcal{O}(Tkd)$. At the same time, since only $k$ additional context centroids need to be stored for clustering, the space complexity is $\mathcal{O}(kd)$.

\textbf{Calculation of Floating Point Operations}. We obtain the number of operations per forward pass for all layers in the encoder (denoted by $E$) and the number of operations per forward pass for all MLP layers in each output head (denoted by $M$), as in \hyperlink{https://openai.com/blog/ai-and-compute/}{https://openai.com/blog/ai-and-compute/}. Therefore, the number of FLOPs of IQ-RE is:
\begin{equation*}
\resizebox{.91\linewidth}{!}{$
    \displaystyle
    2bI(E+kM)+2bI(E+kM)+T(E+kM)+TE$}
\end{equation*}
where $b$ is the batch size; $T$ is the number of environment steps; $I$ is the number of training updates. The first two terms are for the forward and backward passes required in training updates, respectively. The latter two terms are for the forward passes required to compute the policy action and obtain the low-dimensional representation from the random encoder, respectively. In our experiments: $b=32$, $T=1e7$, $I=0.25e7$, $k=4$, $E=28.582$ MFLOPs, $M=3.420$ MFLOPs for Rainbow and $3.215$ MFLOPs for DQN.

\end{document}